\documentclass[opre,nonblindrev]{informs3} 

\OneAndAHalfSpacedXI 

\usepackage{endnotes}
\let\footnote=\endnote
\let\enotesize=\normalsize
\def\notesname{Endnotes}%
\def\makeenmark{$^{\theenmark}$}
\def\enoteformat{\rightskip0pt\leftskip0pt\parindent=1.75em
	\leavevmode\llap{\theenmark.\enskip}}
\makeatletter
\newcommand{\leqnomode}{\tagsleft@true\let\veqno\@@leqno}
\newcommand{\reqnomode}{\tagsleft@false\let\veqno\@@eqno}
\makeatother

\usepackage{natbib}
\bibpunct[, ]{(}{)}{,}{a}{}{,}%
\def\bibfont{\small}%
\usepackage{booktabs}
\usepackage{multirow}
\usepackage{graphicx}
\usepackage[caption=false, font=normalsize, labelfont=sf, textfont=sf]{subfig}
\usepackage{algorithm,algpseudocode}
\usepackage{epstopdf}
\usepackage{pgfplots}
\pgfplotsset{compat=newest,compat/show suggested version=false}
\usepackage{lscape}
\usepackage{pgfplots}
\usepackage{pgf}
\usepackage{mcr}
\usepackage{pifont}
\usepackage{mathrsfs}
\usepackage{makecell}
\usepackage{hyperref}
\TheoremsNumberedThrough     
\ECRepeatTheorems

\EquationsNumberedThrough    

\begin{document}
	
	
	\RUNAUTHOR{Chaosheng and Bo}
	
	\RUNTITLE{Inferring Parameters Through Inverse Multiobjective Optimization}
	\TITLE{Inferring Parameters Through Inverse Multiobjective Optimization}
	
	\ARTICLEAUTHORS{%
		\AUTHOR{Chaosheng Dong, Bo Zeng}
		\AFF{Department of Industrial Engineering, University of Pittsburgh, PA 15261, United States, \EMAIL{chaosheng@pitt.edu},
			\EMAIL{bzeng@pitt.edu}.} 
		
	} 
	
	\ABSTRACT{Given a set of human's decisions that are observed, inverse optimization has been developed and utilized to infer the underlying decision making problem. The majority of existing studies assumes that the decision making problem is with a single objective function, and attributes data divergence to noises, errors or bounded rationality, which, however, could lead to a corrupted inference when decisions are tradeoffs among multiple criteria. In this paper, we take a data-driven approach and design a more sophisticated inverse optimization formulation to explicitly infer parameters of a multiobjective decision making problem from noisy observations. This framework, together with our mathematical analyses and advanced algorithm developments, demonstrates a strong capacity in estimating critical parameters, decoupling  ``interpretable'' components from noises or errors, deriving the denoised \emph{optimal} decisions, and ensuring statistical significance. In particular, for the whole decision maker population, if suitable conditions hold, we will be able to understand the overall diversity and the distribution of their preferences over multiple criteria, which is important when a precise inference on every single decision maker is practically unnecessary or infeasible. Numerical results on a large number of experiments are reported to confirm the effectiveness of our unique inverse optimization model and the computational efficacy of the developed algorithms.
	}%
	
	
	\KEYWORDS{utility estimization; inverse optimization; statistical inference; ADMM; clustering}
	
	\maketitle
	
	%
	
	
	\section{Introduction}
	In business and management practice, a fundamental issue is to interpret the
	observed individuals' behaviors and decisions, and then to develop a sound
	understanding (or inference) on their underlying desires, utility functions, restrictions and overall
	decision making schemes.
	Such information or knowledge, if derived appropriately, should be of a great value
	to enterprises and organizations in promoting better interactions with their
	stakeholders and achieving a stronger performance.
	For example, many studies have been done to help the system planner to infer,
	based on the observed traffic counts on every road link,
	the traffic volume for every origin and destination pair in a road network.
	Such inferred information will be used to support new link constructions or
	capacity expansions with better traffic performance \citep{yang1992estimation}.
	Actually, as digital devices are widely deployed and intensively utilized
	in various business and operations generating abundant data, this issue
	has become more critical and the associated opportunities are actively
	explored among many emerging businesses and practices, such as designing a
	demand response program in a smart grid   and inventory management
	of e-commercial companies.
	
	A common assumption made in the literature is
	that people are rational, i.e., they acquire and carry out optimal
	decisions in their decision making problems.  Then, the inference problem with observed data (i.e., decisions)
	is often formulated as an inverse optimization problem (IOP)
	\citep{troutt2006behavioral,keshavarz2011imputing,bertsimas2015data,aswani2016inverse,esfahani2017data}
	to  estimate parameters of the underlying decision making problem (DMP), e.g., those in the utility function
	or in constraints. The basic idea is that with the estimated parameters,
	which consist of the solution of an IOP, the DMP's expected outcome
	should closely match the observations. Conventionally, the inference on parameters through inverse optimization is based on a single observed decision \citep{ahuja2002combinatorial,heuberger2004inverse,deaconu2008inverse,guler2010capacity}. As more and more data become observable and available, the majority of recent papers adopt the data-driven strategy that directly handles many original observational data with little subjective presumptions. In particular, note that a large amount of observational data unavoidably contain errors, variances or noises. Recent formulations relax the aforementioned assumption and explicitly consider the noisy data issue under different names, e.g., the issue of data inconsistency, imperfect information, suboptimal or approximation solutions \citep{dempe2006inverse,troutt2006behavioral,keshavarz2011imputing,bertsimas2015data,aswani2016inverse,esfahani2017data}.
	
	Specifically, as in \cite{dempe2006inverse,keshavarz2011imputing,bertsimas2015data,aswani2016inverse,esfahani2017data}, considering  a situation where a decision $y_i$ (with respect to an input signal $u_i$) for each $i\in[N]$ is observed and recorded, the IOP model can be formulated to minimize an empirical loss as in the following:
	\begin{eqnarray}
	\label{eq_emp_loss}
	\min_{\theta\in \Theta}\frac{1}{N}\sum_{i=1}^N \textsl{l}_\theta(u_i,y_i),
	\end{eqnarray}
	where $\theta$ denotes the parameters to be estimated and $\Theta$ is the associated domain, and   $\textsl{l}_\theta(u_i,y_i)$ is a loss function that captures the discrepancy between the model inferred from data and the actual model. Among a few loss functions, a typical one is the quadratic loss function, i.e.,
	\begin{eqnarray}
	\textsl{l}_\theta(u_i,y_i) =  && \min_{x_i} \parallel x_i-y_i\parallel^2_2 \label{eq_IOPdist}\\
	&& x_i\in \mfS_{\theta}(u_i)=\arg\min\{f_\theta(x):x\in \mathbf{X}(\theta,u_i)\}, \label{eq_dmp}
	\end{eqnarray}
	where $\mfS_\theta(u_i)$ is the optimal solution set of DMP for given $\theta$ and $u_i$ defined in the right-hand-side of \eqref{eq_dmp}.
	Similar to the situation in regression or design of experiments, this loss function is  to minimize the distance between an optimal solution and the observed decision, and demonstrates a strong statistical performance \citep{dempe2006inverse,aswani2016inverse}.
	
	Recent studies show that such an inverse optimization scheme could be effective in handling data with noise or errors in parameter estimation \citep{keshavarz2011imputing,bertsimas2015data,aswani2016inverse,esfahani2017data}. Nevertheless, simply using noises, errors, variances or even suboptimality to interpret data divergence probably is not appropriate and does not reveal the actual case, especially when data are collected from many decision makers. Note that it has been often observed that decisions are made as a result of trade-off among multiple criteria and different people could have different preferences. For example, investment decisions are basically made to achieve a risk-return balance, which is customized to reflect individual investors' attitudes on these two measures.  Under such a situation, ignoring the impact of their varying preferences over multiple criteria on decision making and simply assuming the same DMP for all decision makers will unlikely produce reasonable inferences. As illustrated in the next example, inverse optimization built upon that simplification could lead to a serious misunderstanding on decision makers' intentions.
	
\begin{figure}[ht]
		\centering
		\includegraphics[width=0.4\linewidth]{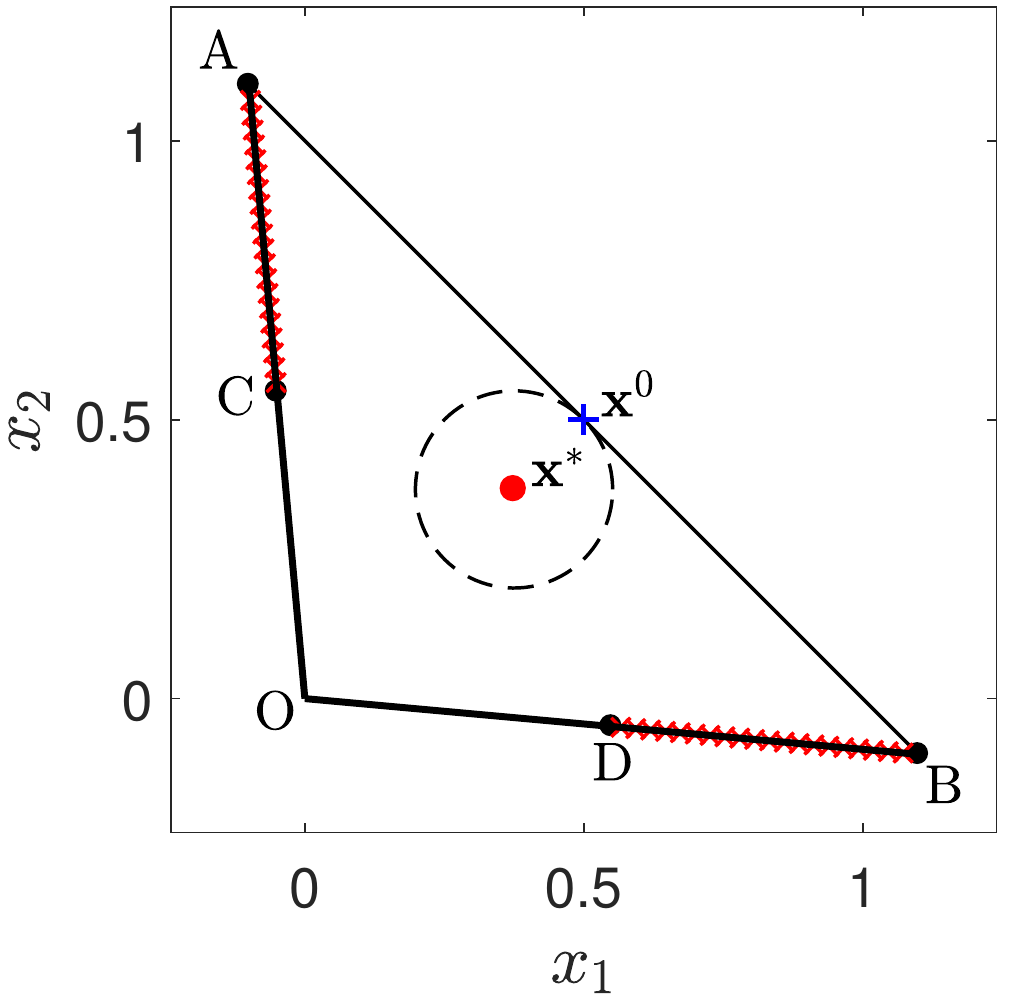}
		\caption{$ O(0,0) $, $ A(-0.2,1.2) $, and $ B(1.2,-0.2) $ are the vertices of the feasible region. $ C(-0.1,0.6) $, $ D(0.6,-0.1) $ and $ \mfx^{0}(0.5,0.5) $ are the midpoints of $ OA $, $ OB $, and $ AB $, respectively. The red dot $ \mfx^*(0.375,0.375) $ is the geometric mean of all the points in segments $ AC $ and $ BD $. The bold segments $ OA $ and $ OB $ are the efficient  (solution) set for the bi-objective linear programming problem.}
		\label{fig:iopvsIOP}
	\end{figure}

	Consider a scenario where decision makers are subject to same restrictions but need to make their individualized \emph{optimal} decisions considering two objective functions, as in the following bi-objective linear programming problem  with $ a > b > 0$ and $c > 0$.  Figure \ref{fig:iopvsIOP} displays the feasible region of an instance with $ a=6, b=1, c=1$, i.e., the triangle $AOB$.
	
	\begin{subequations}
		\begin{align}
		\min \;&  x_1 \\
		\min \;&  x_2 \\
		s.t. \;& a x_{1} + b x_{2} \geq  0,  \label{eq_exp_con1}\\
		&  b x_{1} + a x_{2} \geq 0, \label{eq_exp_con2}\\
		&  x_{1} + x_{2} \leq  c. \label{eq_exp_con3}
		\end{align}
	\end{subequations}
	
	With multiple objectives, rational decision makers seek efficient solutions, which are
	those that cannot be improved without sacrificing performances in one or more criteria (see Section \ref{subsection-mop}). In Figure \ref{fig:iopvsIOP}, it is straightforward to see that points on edges OA and OB are efficient solutions that could be selected by rational decision makers. Assume that many observed decisions evenly occur in segments $AC$ and $BD$. If they are treated as noisy observations of a pristine solution to $\min\{\mfc^{T}\mfx: \eqref{eq_exp_con1}-\eqref{eq_exp_con3}\}$, we can infer the coefficient $\mathbf c$ and obtain a denoised solution $\mathbf{x}^*$ through computing the IOP model with the quadratic loss function, i.e., (\ref{eq_emp_loss}-\ref{eq_dmp}). Actually, noting that optimal $\mathbf{x}^*$  minimizes the averaged distance to those observations, we can derive its analytical characterization.
	
	Specifically, the sum of squares of the Euclidean distance between $\mathbf{x}^*$ and evenly distributed observations on $AC$ and $BD$ can be represented as the following integral:
	\begin{align*}
	& \int_{\frac{bc}{2(a-b)}}^{\frac{bc}{a-b}}\norm{\icolp{x^*_{1} \\ x^*_{2}} - \icolp{-v \\ \frac{a}{b}v}}^{2}\, d\,v + \int_{\frac{bc}{2(a-b)}}^{\frac{bc}{a-b}}\norm{\icolp{x^*_{1} \\ x^*_{2}} - \icolp{\frac{a}{b}v \\ -v}}^{2}\, d\,v \\
	& = \frac{bc}{a-b}(x^*_{1} - \frac{3}{8}c)^{2} + \frac{bc}{a - b}(x^*_{2}- \frac{3}{8}c)^{2} + \Delta,
	\end{align*}
	where $\Delta$ depends on $a, b$ and $c$ only. Thus, $\mfx^{*} = (\frac{3}{8}c,\frac{3}{8}c) $, the arithmetic mean of observations, minimizes this integration. As $\mathbf x^*$ is an interior point, the only $ \mfc $ that renders $\mathbf x^*$ optimal is the trivial one, i.e., $ (c_1, c_2)= (0,0)$, which does not have any relevance to the actual objective functions.
	
	Indeed, we still cannot obtain reasonable explanation of the data, even if taking an additional consideration by restricting $ \mfx^{*}$  to be on the boundary of the feasible region, which helps to avoid the previous trivial estimation. Note from Figure \ref{fig:iopvsIOP} $\mathbf x^0$, i.e., the projection of $\mathbf x^*$ on $AB$, is the optimal boundary point to that integration.  Because $\mathbf x^0$ is in the interior of $AB$, the unique $\mfc$ that renders $\mathbf{x}^0$ optimal is $(c_1, c_2) = (-1,-1)$. This inference basically reflects opposite information regarding decision makers' intentions or desires.
	
	Through this example, it  can be seen that the implicit assumption in the most existing IOP studies could be quite restrictive, especially when observed decisions exhibit a rather diverse pattern across decision makers. Hence, in this paper, we design and study a more general and flexible framework of inverse optimization that is able to explicitly infer multiple objective functions from noisy observations. Such a framework, together with our computationally algorithms and mathematical analysis, demonstrates a strong capacity in estimating critical
	parameters, decoupling  ``interpretable'' components from noises or errors, deriving the denoised \emph{optimal} decisions, and ensuring statistical significance. In particular, for the whole decision maker population, if suitable conditions hold, we will be able to understand the overall diversity and the distribution of their preferences over multiple criteria.  This result could be more important for a manufacturer or service provider, noting that having a precise estimation on every single customer's DMP is practically unnecessary or infeasible when the customer population is large.
	
	We note a couple of IOP studies have also investigated multiple objective function optimization \citep{roland2013inverse,chan2014generalized}. As pointed out in the following literature reviews, our research differs from them in model construction, computational
	methods, and statistical analysis and significance.

	\subsection{Literature Review}
	\label{sect_review}
	Up to now, many studies on parameter estimations through inverse optimization have been designed and developed, where almost all of them assume that the underlying DMP is of a single objective function.  According to the model development and the treated observations, they can be classified into four groups, i.e., inverse optimization with $(i)$ a single observation without noise, $(ii)$  a single observation subject to noise, $(iii)$  multiple observations without noise, and $(iv)$  multiple observations subject to noises.
	
	In the first group, structured inverse network and combinatorial optimization problems are probably the first set of IOP  studies in the literature, where costs of individual arcs are estimated to render the given solution (e.g., network flows, paths, spanning trees) optimal \citep{burton1992instance,zhang1996calculating,ahuja2000faster,ahuja2002combinatorial,heuberger2004inverse,deaconu2008inverse,guler2010capacity}. General linear programming IOP with  a single observation is investigated in the seminal paper by \citet{ahuja2001inverse}, where the distance between the estimated objective function, to which the observation is an optimal solution, and a nominal objective function serves as the loss function. This paper shows that its IOP using $L_{1}$ or $ L_{\infty} $ norm is also a linear program. This research is then further extended to study IOPs of more general decision making schemes, including inverse conic problems \citep{iyengar2005inverse}, inverse optimization for linearly constrained convex separable programming problems \citep{zhang2010inverse}, constrained inverse quadratic programming problems \citep{zhang2010augmented}, and inverse integer programming problems \citep{Schaefer2009,wang2009cutting}. In addition, \citet{ng2000algorithms} considered the problem of inverse reinforcement learning that seeks to extract a reward function given optimal behavior in a Markov decision process.
	
	Different from studies in group one that assume the observation is an optimal decision, which is rather restrictive in practice, IOP studies in the second group allow the observation to be noisy. To the best of our knowledge, \citet{dempe2006inverse} probably produce the first general study considering noisy observation.  They adopt the bilevel optimization to construct an IOP, where the lower level problem receives the utility function estimation and generates an optimal solution of the underlying DMP, and the upper level problem is to determine a utility function that minimizes the distance between that optimal solution and the noisy observation.  \citet{timothy2015inverse} analyze a similar linear programming IOP for a noisy observation, where closed form solutions for several special cases are derived with clear geometric intuitions. Actually, we point out that, although implicitly, the popular O-D matrix estimation problem that in fact is an IOP, has also been treated traditionally as a bilevel model, e.g.,  \citet{yang1992estimation}. Hence, similar to the argument made in \cite{aswani2016inverse}, we believe that bilevel optimization scheme probably provides the most appropriate modeling tool to connect the inference intention and the underlying DMP.
	
	Studies of IOP in the third group extend to consider multiple optimal observations, which can been found in the research on model predictive control (MPC) \citep{baes2008every,nguyen2014inverse,hempel2015inverse,nguyen2015any,Nguyen2017}. In this context, a control law, which might be a piecewise function with each piece representing an optimal solution over a region in a polyhedral partition of the parameter space, will be used to recover parameters of the underlying DMP.  Note that multiple pieces of that function, which are treated as multiple optimal observations, should be considered simultaneously in the associated IOP mode \citep{nguyen2014inverse,hempel2015inverse,nguyen2015any,Nguyen2017}.
	
	The research of IOP in the fourth group, which takes the data-driven approach to directly consider multiple noisy observations, recently has received a substantial attention   \citep{troutt2006behavioral,keshavarz2011imputing,bertsimas2015data,aswani2016inverse,esfahani2017data}.  In \citet{troutt2006behavioral}, an IOP formulation that minimizes the \emph{decisional regret}, which is the value differences between observed decisions and expected solutions associated with the cost estimation, is developed and then is illustrated for cost estimation in production planning. \citet{keshavarz2011imputing} present an IOP framework to impute a convex objective function by minimizing the residuals of Karush-Kuhn-Tucker (KKT) conditions incurred by noisy data. Similarly, an inverse variational inequalities problem, which is a more general scheme, is introduced in \citet{bertsimas2015data}, noting that solutions of an optimization problem can be represented as solutions to a set of variational inequalities. Then, parameter estimation is derived to minimize the slackness needed to render observations to (approximately) satisfy those variational inequalities. We mention that in \citet{aswani2016inverse} a bilevel optimization based IOP that minimizes the differences between observations and expected optimal solutions is introduced, whose, for the first time, statistical consistency properties with respect to noisy observations are systematically analyzed and established. In the most recent paper \citep{esfahani2017data}, the authors propose to adopt the suboptimality loss in IOP, which has a clear advantage in the computational tractability over that in \citet{aswani2016inverse}, and formulate a distributionally robust IOP model to achieve some out-of-sample guarantees.

	The inverse optimization research for DMP with multiobjective functions is rather new and much less investigated. \cite{roland2013inverse} consider  an IOP for a binary integer DMP given a set of linear objective functions, and develops branch-and-bound and cutting plane algorithms, which are not numerically evaluated yet, to find minimal adjustment of the objective functions such that a given set of feasible solutions becomes efficient. Research in \cite{chan2014generalized} addresses another situation where preferences or weights of several known (linear) criteria in the decision making problem will be inferred based on a single noisy observation. A demonstration on cancer therapy shows that their inversely optimized weights of medical metrics leads to clinically acceptable treatments. Different from those studies, our study follows the data-driven approach to build an IOP framework that directly considers many noisy observations to infer multiple objective functions or constraints of a convex DMP with a solid statistical significance.
	
	\subsection{Contributions}

	We summarize our main contributions in the following.\\
	$(i)$ \textit{A new inverse optimization model with a stronger inferring capability:} We develop  a new inverse multiobjective optimization problem (IMOP) that is able to infer multiple criteria (or constraints) over which the trade-off decisions are made. Comparing to most existing studies that are primarily different in loss functions, it has a more sophisticated structure and a stronger capacity in decoupling  ``interpretable'' components from noises or errors and revealing parameters of the actual objectives adopted by  decision makers.\\
	$(ii)$ \textit{A solid theoretical analysis on inference's significance:} We provide a solid analysis to ensure the statistical significance of the inference results from our IMOP model. In particular, a completely new type of consistency is investigated such that we are able to asymptotically recover the  underlying diversity and the distribution of decision makers' preferences over multiple criteria, which is of a critical value when inference of a single decision maker is practically unnecessary or infeasible. Also, the concept of identifiability is defined in the context of multiple objectives with the first procedure to verify whether a DMP is identifiable. \\
	$(iii)$ \textit{A couple of effective and generally applicable algorithms:} To handle the challenge of a large number of observations, we consider in the first algorithm the use of ADMM as heuristic for solving the learning problem. Moreover, with a deep insight on its structure, we reveal a hidden connection between our IMOP and the popular K-means clustering problem, and leverage the latter one in designing another powerful algorithm to handle many noisy data. Numerical results on a large number of experiments confirm that the proposed algorithms can solve IMOP with a great accuracy while drastically improve the computational efficacy.
	
	\subsection{Organization}
	The remainder of the paper is organized as follows. In Section 2, we first present preliminaries for the decision making problem with multiple objectives. Then, we propose the inverse optimization models to infer parameter of a multiobjective decision making problem. In Section 3, we show the risk consistency of the estimators constructed by solving the inverse optimization models. Section 4 introduces the concept of identifiability for a decision making problem, and discuss its relationship with the estimation consistency of the parameter and preference. Section 5 derives two algorithms for solving the inverse optimization model we propose. Numerical results are reported in Section 6. We conclude the paper in Section 7. The omitted proofs for lemmas, mathematical reformulations and data are included in Appendix.

	\subsection{Notation}
	Throughout this paper we use $ \one_{n} $ and $ \zero_{n} $ to denote the vector of ones and all zeros in $ \bR^{n} $, respectively. We let $ \bI $ denote the identity matrix. For any $ n \geq 1 $, the set of integers $ \{1,\ldots,n\} $ is denoted by $ [n] $. We let $ \bR^{p}_{+} = \{\mfx \in \bR^{p}: \mfx_{i} \geq 0, \forall i \in[p] \} $, and $ \bR^{p}_{++} = \{\mfx \in \bR^{p}: \mfx_{i} > 0, \forall i \in[p] \} $.

	\section{Inference through Inverse Multiobjective Optimization}
	
	\subsection{Decision Making Problem with Multiple Objectives} \label{subsection-mop}
	Consider the following decision making problem with $p$ $(\geq 2)$ objective functions parameterized by $\theta$:
	\leqnomode\begin{align}
	\label{mop}
	\tag*{DMP}
	\begin{array}{llll}
	\min\limits_{\mfx \in \bR^{n}} & \{f_{1}(\mfx,\theta),f_{2}(\mfx,\theta),\ldots,f_{p}(\mfx,\theta)\} \\
	\;s.t. & \mfx \in X(\theta).
	\end{array}
	\end{align}\reqnomode
	For easy  exposition, we use $ \mathbf{f}(\mfx,\theta)$ to denote the vector of objective functions  $(f_{1}(\mfx,\theta),f_{2}(\mfx,\theta),\ldots,f_{p}(\mfx,\theta))^{T} $. Also, the set $X(\theta)$ is characterized as $ X(\theta) = \{\mfx \in \bR^{n}: \mathbf{g}(\mfx,\theta) \leq \zero\} $, where $ \mathbf{g}(\mfx,\theta)=(g_{1}(\mfx,\theta),\ldots,g_{q}(\mfx,\theta))^{T} $ is another vector-valued function. Following the current mainstream of inverse optimization study \citep{keshavarz2011imputing,bertsimas2015data,aswani2016inverse,esfahani2017data}, we  restrict our focus to a convex DMP defined next.
	
	\begin{definition}
		\ref{mop} is said to be convex if $ X(\theta) $ is a convex set and $ \mathbf{f(\mfx,\theta)}$ is continuous and convex on $X(\theta)$, i.e.,  $f_{l}(\mfx) $ is continuous and convex on $X(\theta)$ for all $l \in [p] $.
	\end{definition}
	
	Noting that these objective functions reflect human decision makers' multiple desires, it would be ideal to derive a decision that would be optimal for all of them simultaneously. Nevertheless, no such optimal decision may exist due to their incompatibility. Hence, we must treat them in a comprehensive way and derive one or more \emph{strong} decisions that capture the trade-off between those objective functions. Next, for a DMP with fixed $\theta$, we introduce formal definitions of those strong trade-off decisions, and present relevant structural properties.
	
	\begin{definition}[efficiency]
		A decision vector $\mfx^{*} \in X(\theta)$ is said to be \textit{efficient} (or \textit{Pareto optimal}, or \textit{non-dominated}) if there exists no other decision vector $\mfx \in X(\theta)$ such that $f_{i}(\mfx,\theta) \leq f_{i}(\mfx^{*},\theta)$ for all $i \in [p]$, and $f_{k}(\mfx,\theta) < f_{k}(\mfx^{*},\theta)$ for some  $k \in [p]$.
	\end{definition}
	The set of all efficient solutions is denoted by $ X_{E}(\theta) $, which is then called the \textit{efficient set}. Based on the definition, an efficient solution, which is evaluated according to multiple criteria, is one that cannot be further improved without sacrificing performance in some criterion. It can be seen as an analogy in the context of multiple objective functions to an optimal solution to optimization with a single objective function. Certainly, by varying our preferences over those evaluation criteria, different efficient solutions are likely to be derived. A natural and common strategy to derive an efficient solution is to compute an  optimization problem with a single objective function constructed by a weighted sum of original functions, i.e., to solve the weighting problem (WP) \citep{gass1955computational} defined in the following:
	\begin{align}
	\label{weighting problem}
	\tag*{WP}
	\begin{array}{llll}
	\min & w^{T}\mathbf{f}(\mfx,\theta) \vspace{1mm}\\
	\;s.t. & \mfx \in X(\theta),
	\end{array}
	\end{align}
	where $ w = (w^{1},\ldots,w^{p})^{T}$ is a nonnegative weight vector. Indeed, without loss of generality, any realistic weight vector can be equivalently represented by a vector in set $\mathscr{W}_{p}  \equiv \{ w\in \bR^{p}_{+} : \; \one^{T}w = 1 \}$. When all weight components are required to be positive, such set is denoted by $\mathscr{W}^+_{p}$.

	Denote $S(w,\theta)$ the set of optimal solutions for \ref{weighting problem} with a particular $w$, i.e., $S(w,\theta) = \argmin_{\mfx}\left\{ w^{T}\mathbf{f}(\mfx,\theta): \mfx \in X(\theta) \right\}$. Then, we have a couple of theoretical results regarding \ref{weighting problem} that directly follow Theorems 3.1.1 - 3.1.3 of \citet{miettinen2012nonlinear}.
	\begin{proposition}\label{prop:unique-weakly}
		Let $ \mfx \in S(w,\theta) $ be an optimal solution of \ref{weighting problem}. The following statements hold.
		\begin{description}
			\item[(a)] If $ w \in \mathscr{W}_{p}^{+} $, then $ \mfx \in X_{E}(\theta) $.
			\item[(b)] If $ \mfx $ is the unique optimal solution of \ref{weighting problem}, then $ \mfx \in X_{E}(\theta) $.
		\end{description}
	\end{proposition}
	
	According to Proposition 3.10 of \citet{ehrgott2005mutiobjective} and Theorem 3.1.4 of \citet{miettinen2012nonlinear}, all efficient solutions of a convex \ref{mop} can be found by solving \ref{weighting problem}.
	\begin{proposition}\label{weight_convex}
		Given that \ref{mop} is convex and $ \mfx\in X_{E}(\theta)$, there exists a weight vector $ w \in \mathscr{W}_{p} $ such that $ \mfx $ is an optimal solution to \ref{weighting problem}, i.e., $\mfx\in S(w,\theta) $.
	\end{proposition}
	
	Based on Propositions \ref{prop:unique-weakly} and \ref{weight_convex}, the following inclusive relationships can be derived.
	\begin{corollary}\label{coro:inclusion}
		For a convex \ref{mop},
		\begin{align*}
		\bigcup_{w\in \mathscr{W}_{p}^{+} }S(w,\theta) \subseteq X_{E}(\theta) \subseteq \bigcup_{w\in \mathscr{W}_{p}}S(w,\theta).
		\end{align*}
	\end{corollary}
	\textbf{Remark:}  $(i)$ Results in Corollary \ref{coro:inclusion} provides us a theoretical basis to make use of the weighted sum method to derive all efficient solutions. Actually, when \ref{mop} is convex and the objective functions are strictly convex, we have $ X_{E}(\theta) = \bigcup_{w\in \mathscr{W}_{p}}S(w,\theta)$.
	$(ii)$ When \ref{mop} is convex and $X(\theta)$ is compact, one important property of $X_{E}(\theta)$ is that it is a connected set, which, however, might not be convex as stated in \citet{warburton1983quasiconcave,ehrgott2005mutiobjective}. We note that it is very different from the situation of a convex single objective optimization problem, whose optimal solution set is convex.
	
	\subsection{Inverse Multiobjective Optimization with Noisy Observations}
	In this section, we present the development of our inverse optimization models for parameter learning. Specifically, given a set of observations that are noisy efficient solutions collected from the decision maker population under study, we construct an inverse optimization model to infer parameter $\theta$ of the multiobjective decision making problem defined in \ref{mop}. In addition to its more sophisticated structure, it is worth pointing out that we must handle a new challenge that does not occur in any inverse optimization with a single objective function. Different from the single objective case that typically employs observations consisting of clear signal-response pairs \citep{keshavarz2011imputing,bertsimas2015data,aswani2016inverse,esfahani2017data}, decision makers' decisions are often observed without any information on their trade-off among objective functions. Under such a situation, as demonstrated in this section, a non-traditional inverse optimization framework shall be developed to address this challenge.
	
	\subsubsection{Loss Function and Its Sampling Based Variants}\label{sec:loss function}
	We consider a set of observations that are noisy efficient solutions collected with  possible measurement errors or decision makers' bounded rationality. Let $ \mfy $ denote one such observation that is distributed according to an unknown distribution $ \bP_{\mfy} $ and supported on $\cY $. As noted in \cite{aswani2016inverse,esfahani2017data}, noise might come from measurement error, and thus $ \mfy $ does not necessarily belong to $ X(\theta) $. Next, we describe the construction of our loss function with respect to a hypothesis $\theta$. When weights over objective functions, i.e., the weight vector $w$, are known, the conventional loss function in \eqref{eq_IOPdist}-\eqref{eq_dmp} can be directly applied with respect to $\mfy$ and $S(w,\theta)$. Nevertheless, as previously mentioned, $w$ is often missing and the efficient set should be adopted instead as in the following.
	\begin{align}\label{loss function}
	\tag*{loss function}
	l(\mfy,\theta) = \min_{\mfx \in X_{E}(\theta)} \norm{\mfy - \mfx}^{2},
	\end{align}
	where  $X_{E}(\theta)$ is the efficient set of \ref{mop} for a given $\theta$.
	
	One challenge is that there is no general approach to comprehensively and explicitly characterize the efficient set $X_{E}(\theta)$. One way is to introduce weight variable representing the appropriate weight and convert the \ref{loss function} into $$\min_{w\in \mathscr{W}_{p}, \mfx \in S(w,\theta)} \norm{\mfy - \mfx}^{2}.$$
	
	However, this approach might not be suitable for a data-driven study, since it results in a drastically complicated model, where every single observation requires one weight variable and the nonlinear term between it and $\theta$ is heavily involved. On the contrary, according to Corollary \ref{coro:inclusion} and its following remarks, we adopt a sampling approach to generate $w_{k}\in \mathscr{W}_{p}$ for each $k\in[K]$ and approximate $X_{E}(\theta)$ as the union of their $S(w_{k},\theta)$s. Then, by utilizing binary variables that select an appropriate efficient solution from this union, the loss function is converted into the following
	\textit{sampling based loss problem}.
	\begin{align}
	\begin{array}{llll}
	l_{K}(\mfy,\theta) &= & \min_{\mfx_k, z_k\in \{0,1\}} \ \norm{\mfy - \sum_{k\in[K]}z_{k}\mfx_{k}}^{2} \vspace{1mm}\\
	& \mbox{s.t.} & \sum\limits_{k\in[K]}z_{k} = 1, \  \mfx_{k} \in S(w_{k},\theta).
	\end{array}
	\end{align}
	
	\begin{remark}
		\begin{description}
			\item[(i)]
			Constraint $\sum_{k\in[K]}z_{k} = 1$ ensures that exactly one of efficient solutions will be chosen to measure the distance to $\mfy$. Hence, solving this optimization problem identifies some $w_k$ with $k\in [K]$ such that the corresponding efficient solution $S(w_{k},\theta)$ is closest to $\mfy$.
			\item[(ii)] As shown in Corollary \ref{coro:inclusion}, it is guaranteed that no efficient solution will be excluded if all weight vectors in $\mathscr{W}_{p}$ are enumerated.
			As it is practically infeasible, we can control the number of sampled weights to achieve a desired tradeoff between the approximation accuracy and computational efficacy. Certainly, if the computational power is strong, we would suggest to draw a large number of weights evenly in $\mathscr{W}_{p}$ to avoid any bias. Although a set of binary variables is needed for each observation, the number of sampled weights is independent from the number of observations.
			\item[(iii)] Indeed, as shown in Section \ref{section:identifiability}, the large number of weight samples help recover the distribution of weights among decision makers under suitable conditions. As discussed earlier, such information should be very critical to manufacturers or service providers when dealing with many customers.
		\end{description}
	\end{remark}
	
	\subsubsection{Models for IMOP} \label{section: estimators}
	Using the \ref{loss function}, our inverse optimization problem can be formulated as follows
	\leqnomode\begin{align*}
	\label{general-inverse-model}
	\tag*{IMOP}
	\begin{array}{llll}
	\min\limits_{\theta \in \Theta} & M(\theta) \equiv \bE\bigg( l(\mfy,\theta) \bigg),
	\end{array}
	\end{align*}\reqnomode
	where $ \Theta  $ is the feasible set of $ \theta $. Similar to existing statistical studies \citep{aswani2016inverse,esfahani2017data}, function $M(\theta)$ is also called the risk of the loss function $l(\mfy,\theta)$. As in most inverse optimization studies \citep{keshavarz2011imputing,bertsimas2015data,aswani2016inverse,esfahani2017data}, we make the next assumption in the remainder of this paper.
	\begin{assumption}
		\label{convex_setting}
		$ \Theta$ is a convex set. For each $ \theta \in \Theta $, $ \mathbf{f}(\mfx,\theta) $ and $ \mathbf{g}(\mfx,\theta) $ are convex in $ \mfx $.
	\end{assumption}
	
	Practically, $ \theta $ can not be learned by directly solving \ref{general-inverse-model} as $ \bP_{\mfy} $ is not known a priori. Given available observations $ \{\mfy_{i}\}_{i \in [N]} $, it is often the case that $\theta$ will be inferred through solving the following empirical risk minimizing problem
	\leqnomode\begin{align*}
	\label{saa-inverse-model}
	\tag*{IMOP-EMP}
	\begin{array}{llll}
	\min\limits_{\theta \in \Theta} & M^{N}(\theta)\equiv\frac{1}{N}\sum\limits_{i \in [N]}l(\mfy_{i},\theta). \\
	\end{array}
	\end{align*}\reqnomode
	
	As previously mentioned, we indeed do not have the explicit representation of $X_{E}(\theta)$. Through the sampling approach described in the last subsection, variants of \ref{general-inverse-model} using sampled weights can be easily defined. The following one is to reformulate \ref{general-inverse-model} with weight samples, which helps us in performing theoretical analysis of the reformulation of \ref{saa-inverse-model}.
	\leqnomode\begin{align*}
	\label{ws-general-inverse-model}
	\tag*{IMOP-WS}
	\begin{array}{llll}
	\min\limits_{\theta \in \Theta } & M_{K}(\theta) \equiv \bE\bigg( l_{K}(\mfy,\theta) \bigg).
	\end{array}
	\end{align*}\reqnomode

	Next, we provide the reformulation of \ref{saa-inverse-model} with weight samples. As it serves as the primary model for analysis and computation, we present its comprehensive form to facilitate our discussion and understanding.

	\begin{align*}
	\label{saa-general-inverse-model}
	\tag*{IMOP-EMP-WS}
	\begin{array}{llll}
	\min\limits_{\theta \in \Theta } & M^{N}_{K}(\theta) \equiv \frac{1}{N}\sum\limits_{i \in [N]}\lVert \mfy_{i} - \sum\limits_{k \in [K]}z_{ik}\mfx_{k}\rVert_{2}^{2} \vspace{1mm}\\
	\;\text{s.t.} & \mfx_{k} \in S(w_{k},\theta), & \forall k \in [K], \vspace{1mm}\\
	& \sum\limits_{k \in [K]}z_{ik}  = 1, & \forall i \in [N], \vspace{1mm} \\
	& z_{ik} \in \{0,1\}, & \forall i \in [N],\; k \in [K].
	\end{array}
	\end{align*}
	
	\begin{remark}
		By making use of optimality conditions to represent $S(w_{k},\theta)$, \ref{saa-general-inverse-model}  can be solved numerically to derive an estimation of $\theta$. According to \cite{aswani2016inverse,esfahani2017data}, existing data-driven inverse optimization models primarily differ from each other by using different loss functions. Our \ref{saa-general-inverse-model} model clearly has a more sophisticated structure with many new variables and constraints, which probably are necessary due to the learning context and task. To handle the incurred computational challenge, advanced algorithm developments are presented in Section \ref{solution-approach}, which support our real applications with a greatly improved efficiency.
	\end{remark}	
	
	It occurs that partial information on some parameters of objective functions or constraints are available, or some decisions are observed with  knowledge on the range of weights over those incomplete objective functions. For example, some decision makers are risk-averse, indicating that their decisions are with large weights over the function representing risk. Under such a situation, our model can be easily extended to handle observations that have some weight-decision information. Specifically, the following constraints can be used to replace the second set of constraints in \ref{saa-general-inverse-model}.
	\begin{align}
	\label{saa-general-inverse-model-modified}
	\begin{array}{llll}
	& \sum\limits_{k \in \widetilde K_{i}}z_{ik}  = 1 & \forall i \in [N'], \vspace{1mm} \\
	& \sum\limits_{k \in [K]}z_{ik}  = 1 & \forall i \in [N]\setminus [N'],
	\end{array}
	\end{align}
	where the first $ N' $ observations are with some information on weights captured in subset $\widetilde{K}_{i}\subseteq [K]$ for each $i\in [N']$. If we would like to emphasize the contribution of the observations in learning, the objective function of \ref{saa-general-inverse-model} can be modified as follows:
	$$\min\limits_{\theta \in \Theta}  \frac{1}{N}\sum\limits_{i \in [N]\setminus[N']}\lVert \mfy_{i} - \sum\limits_{k \in [K]}z_{ik}\mfx_{k}\rVert_{2}^{2} + \frac{\lambda}{N}\sum\limits_{i \in [N']}\lVert \mfy_{i} - \sum\limits_{k \in \widetilde K_{i}}z_{ik}\mfx_{k}\rVert_{2}^{2},$$
	where coefficient $ \lambda \geq 1$ reflects the value of such more specific information.

	Before proceeding to next section, we summarize the proposed models for IMOP in Table \ref{table:four models}, where \textbf{Empirical} and \textbf{Obj} mean that we use empirical risk and the specific objective function, respectively. Here, $N$ is the number of observations, and $K$ denotes the number of weight samples.
	
	\newcommand{\xmark}{\ding{55}}%
	\begin{table}[ht]
		\centering
		\caption{Summary of Four IMOP Models}
		\label{table:four models}
		\begin{tabular}{|c|c|c|c|c|c|}
			\hline
			Model                             & Risk/Empirical & Loss function        & Obj                 & Estimator              & Computable   \\ \hline
			{\scriptsize \ref{general-inverse-model} }    & Risk           & $l(\mfy,\theta)$     & $M(\theta)$         & $\theta^{*}$           & \xmark         \\ \hline
			{\scriptsize \ref{saa-inverse-model} }        & Empirical      & $l(\mfy,\theta)$     & $M^{N}(\theta)$     & $\hat{\theta}^{N}$     & \xmark         \\ \hline
			{\scriptsize \ref{ws-general-inverse-model}}  & Risk           & $l_{K}(\mfy,\theta)$ & $M_{K}(\theta)$     & $\hat{\theta}_{K}$     & \xmark          \\ \hline
			{\scriptsize \ref{saa-general-inverse-model}} & Empirical      & $l_{K}(\mfy,\theta)$ & $M^{N}_{K}(\theta)$ & $\hat{\theta}^{N}_{K}$ & \checkmark \\ \hline
		\end{tabular}
	\end{table}

	\section{Estimators' Risk Consistency and Generalization Bound}
	\label{sect:stat_I}
	In this section, we perform theoretical studies on a statistical property, i.e., risk consistency, of estimators constructed in Section \ref{section: estimators}. More specifically, we show that these estimators asymptotically predict as well as the best possible result this type of inverse optimization model can achieve. In addition, we provide a generalization bound for the estimator constructed in \ref{saa-general-inverse-model}.

	\subsection{Uniform Convergence of the Empirical Risks}
	\begin{figure}
		\centering
		\includegraphics[width=0.5\linewidth]{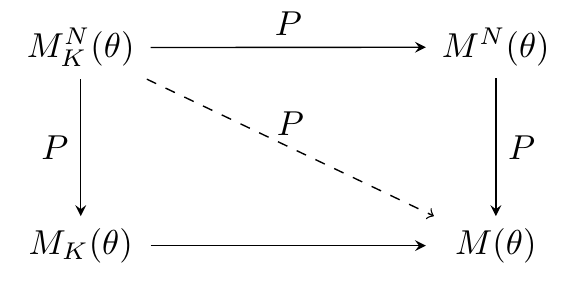}
		\caption{Uniform convergence diagram for empirical risks. $ \overset{P}{\longrightarrow} $  means convergence in probability. $ \longrightarrow $ indicates the convergence of a sequence of numbers. $ \overset{P}{\dashrightarrow} $ means convergence in probability for double-index random variable.}
		\label{fig:diagram}
	\end{figure}
	
	Before proving the risk consistency of the estimators, we first need to prove the uniform convergence of the empirical risks as shown in Figure \ref{fig:diagram}. Different from conventional learning tasks that consider convergence only in data size $ N $, we need to show that the empirical risk $ M^{N}_{K}(\theta) $ uniformly converges to the risk $ M(\theta) $ in two directions, that is, in $ N $ and $ K $ simultaneously. We now introduce a few assumptions typically adopted in the literature to define a friendly structure of our \ref{mop}.
	
	\begin{assumption}
		\label{set-continuous-assumption}
		\begin{description}
			\item[(i)] The parameter set $ \Theta $ is compact.
			\item[(ii)] For each $ \theta \in \Theta $,  $ X(\theta) $ is compact, and has a nonempty relatively interior. Also, $ X(\theta) $ is uniformly bounded. Namely, there exists $ B >0 $ such that $ \norm{\mfx} \leq B $ for all $ \mfx \in X(\theta) $ and $ \theta \in \Theta $.
			\item[(iii)] Functions $ \mathbf{f}(\mfx,\theta) $ and $ \mathbf{g}(\mfx,\theta) $ are continuous on $ \bR^{n} \times \Theta $.
			\item[(iv)] $ \bE[\mfy^{T}\mfy] < +\infty $.
		\end{description}
	\end{assumption}
	These assumptions are practically mild and widely adopted in existing inverse optimization studies, e.g., \cite{aswani2016inverse}. Assumptions $(ii)$ and $(iii)$ are important for the continuity of $X_E(\theta)$. Also,  Assumption $(iv)$, which is ensured once variance of the noise is finite, is fundamental to applying  \textit{the uniform law of large numbers} (ULLN) \citet{jennrich1969asymptotic},  one of the most used tools in performing consistency analysis.
	\begin{lemma}\label{feasible-set-continuous}
		Suppose Assumptions \ref{convex_setting} - \ref{set-continuous-assumption} hold. $ X(\theta) $ is continuous on $ \Theta $.
	\end{lemma}
	The continuity of $ X(\theta) $ follows from its lower semicontinuity (l.s.c.) and upper semicontinuity (u.s.c.), both of which can be derived by using \cite{hogan1973point} under our assumptions.
	
	\begin{lemma}\label{lemma:efficient-set-continuous}
		Suppose Assumptions \ref{convex_setting} - \ref{set-continuous-assumption} hold. If $ \mathbf{f}(\mfx,\theta) $ is strictly convex in $ \mfx $ for each $ \theta \in \Theta $, then $ X_{E}(\theta) $ is continuous on $ \Theta $.
	\end{lemma}

	\begin{remark}
		Several things need to be emphasized when applying Theorem 7.1 of \citet{tanino1980stability} to prove Lemma \ref{lemma:efficient-set-continuous}. $ (i) $ This theorem employs the condition that $ X(\theta) $ is uniformly compact near $ \theta $, which guarantees that a sequence $ \{\mfx_{k}\} $, generated from $ X(\theta_{k}) $, contains a convergent subsequence. In Euclidean spaces, the uniform boundedness of $ X(\theta) $, as stated in Assumption \ref{set-continuous-assumption}, is also adequate in the proof.
		$ (ii) $ This theorem gives the sufficient conditions for the l.s.c. of $ X_{E}(\theta) $. All of these conditions are naturally satisfied under Assumptions \ref{convex_setting} - \ref{set-continuous-assumption} except the one that requires $ \mathbf{f}(\mfx,\theta) $ to be one-to-one, i.e., injective in $ \mfx $. In fact, we can safely replace the one-to-one condition by the strict quasi-convexity of $ \mathbf{f}(\mfx,\theta) $ in $ \mfx $ without affecting the result. Since strict convexity implies strict quasi-convexity, the lower semicontinuity naturally follows.
	\end{remark}
	
	\begin{proposition}[ULLN for $ M^{N}(\theta) $ in $ N $]\label{prop:M^{N} - M}
		Under the same conditions of Lemma \ref{lemma:efficient-set-continuous}, $ M^{N}(\theta) $ uniformly converges to $ M(\theta) $ in $ N $. That is,
		\begin{align*}
		\sup\limits_{\theta \in \Theta}|M^{N}(\theta)- M(\theta)| \overset{p}{\longrightarrow} 0 .
		\end{align*}
	\end{proposition}
	\proof{Proof. }	
	We apply Theorem 2 of \citet{jennrich1969asymptotic} in our proof. We start by checking that the three conditions for using this theorem are satisfied. First, by Lemma \ref{lemma:efficient-set-continuous}, $ X_{E}(\theta) $ is continuous. Then,  applying Berge Maximum Theorem \citep{berge1963topological} to \ref{saa-inverse-model} implies that the empirical risk $ M^{N}(\theta) $ is continuous. Second, by Assumption \ref{set-continuous-assumption}, $ \Theta $ is a compact set. Third, $\forall \mfy \in \mathcal{Y}, \min_{\mfx \in X_{E}(\theta)}\lVert \mfy - \mfx\rVert_{2}^{2} \leq \norm{\mfy}^{2} + B^{2} + 2B\norm{\mfy} $ and the right-hand side is integrable with respect to $ \mfy $ under Assumption \ref{set-continuous-assumption}. Consequently, all three conditions are satisfied and the proof is concluded.
	\Halmos\endproof
	
	\begin{proposition}[ULLN for $ M_{K}^{N}(\theta) $ in $ N $]\label{prop:M_{K}^{N} - M_{K} }
		Under the same conditions of Lemma \ref{lemma:efficient-set-continuous}, $ M_{K}^{N}(\theta) $ uniformly converges to $ M_{K}(\theta) $ in $ N $. That is, $ \forall K $,
		\begin{align*}
		\sup\limits_{\theta \in \Theta}|M_{K}^{N}(\theta)- M_{K}(\theta)| \overset{p}{\longrightarrow} 0.
		\end{align*}
	\end{proposition}
	\proof{Proof. }	
	Similar to Proposition \ref{prop:M^{N} - M}, the key step is to show the continuity of $ M_{K}^{N}(\theta) $ in $ \theta $ for each $ K $. It suffices to show that $ \bigcup_{k\in [K]}  S(w_{k},\theta) $ is continuous in $ \theta$ for all $ K $. First, let us establish the continuity of $ S(w_{k},\theta) $ in $ \theta $ for each $ k \in [K] $. Note that the feasible region $ X(\theta) $ is irrelevant to $ w $. Thus, applying the Berge Maximum Theorem \citep{berge1963topological} to \eqref{weighting problem} implies that $ S(w_{k},\theta) $ is upper semicontinuous in $ \theta $. Hence, $ S(w_{k},\theta) $ is continuous in $ \theta $ as it is a single-valued set. Second, let us show the continuity of $ \bigcup_{k\in [K]}  S(w_{k},\theta) $ in $ \theta$. By Propositions 2 and 4 of \citet{hogan1973point}, we know that a finite union of continuous sets, i.e., $ \bigcup_{k\in [K]}  S(w_{k},\theta) $, is continuous in $ \theta $. Finally, applying Theorem 2 of \citet{jennrich1969asymptotic} yields the uniform convergence of $ M_{K}^{N}(\theta) $ to $ M_{K}(\theta) $ in $ N $.
	\Halmos\endproof
	
	
	Throughout the paper, we use $ K_{2} \geq K_{1} $ to denote the set of weights $ \{w_{k}\}_{k\in[K_{1}]} \subseteq \{w_{k}\}_{k\in[K_{2}]} $, and $ K_{2} > K_{1} $ to denote the set of weights $ \{w_{k}\}_{k\in[K_{1}]} \subsetneq \{w_{k}\}_{k\in[K_{2}]} $. Then, we have the following two lemmas depicting the monotonicity of $ \{M_{K}(\theta)\} $ and $ \{M_{K}^{N}(\theta)\} $ in $ K $ for each $ \theta \in \Theta $.
	
	\begin{lemma}[Monotonicity of $ \{M_{K}(\theta)\} $ and $ \{M_{K}^{N}(\theta)\} $ in $ K $]\label{lemma:M_J monotone} We have the following:
		\begin{description}
			\item[\textbf{(a)}] The sequence $ \{M_{K}(\theta)\} $ is monotone decreasing in $ K $ for all $ \theta \in \Theta $. Moreover, $ \{M_{K}(\hat{\theta}_{K})\} $ is monotone decreasing in $ K $. Specially, $ M_{K}(\hat{\theta}_{K}) \geq M(\theta^{*}) $.
			\item[\textbf{(b)}] Given any $ \{\mfy_{i}\}_{i \in [N]} $, the sequence $ \{M_{K}^{N}(\theta)\} $ is monotone decreasing in $ K $ for all $ \theta \in \Theta $. Moreover, $ \{M_{K}^{N}(\hat{\theta}_{K}^{N})\} $ is monotone decreasing in $ K $. Specially, $  M_{K}^{N}(\hat{\theta}^{N}_{K}) \geq M^{N}(\hat{\theta}^{N}) $.
		\end{description}
	\end{lemma}

	\begin{lemma}\label{lemma:lpshitz of S(w,theta)}
		Suppose Assumptions \ref{convex_setting} - \ref{set-continuous-assumption} hold. Suppose also that $ \mathbf{f}(\mfx,\theta) $ is strongly convex in $ \mfx $ for each $ \theta \in \Theta $, that is, $\forall l \in [p] $, $ \exists \lambda_{l} > 0 $, $ \forall \mfx, \mfy \in \bR^{n} $,
		\begin{align*}
		f_{l}(\mfy,\theta) \geq f_{l}(\mfx,\theta) + \nabla f_{l}(\mfx,\theta)^{T}(\mfy - \mfx) + \frac{\lambda_{l}}{2}\norm{\mfy - \mfx}^{2}.
		\end{align*}
		
		Then, $ \forall \theta \in \Theta $, $ \forall w, w_{0} \in \mathscr{W}_{p} $,
		\begin{align*}
		\lVert S(w,\theta) - S(w_{0},\theta) \rVert_{2} \leq \frac{2L}{\lambda}\norm{w - w_{0}},
		\end{align*}
		where $ L = \sqrt{p}\cdot\max_{l \in [p], \theta \in \Theta, \mfx \in X(\theta)}|f_{l}(\mfx,\theta)|  $ is a finite number, and $ \lambda = \min_{l \in [p]}\{\lambda_{l}\} $.
	\end{lemma}
	%
	%
	%
	%
	
	
	\begin{proposition}[Uniform convergence of $ M_{K}(\theta) $ in $ K $]\label{prop:M_J - M}
		Under the same conditions of Lemma \ref{lemma:lpshitz of S(w,theta)}, $ M_{K}(\theta) $ uniformly converges to $  M(\theta) $ in $ K $ for $ \theta \in \Theta $. That is, $ \sup\limits_{\theta \in \Theta}|M_{K}(\theta)- M(\theta)| \longrightarrow 0 $.
	\end{proposition}
	\proof{Proof. }
	Note that $ \forall \theta \in \Theta $, $ S(w,\theta) $ is single-valued due to the fact that $ \mff $ is strongly convex. $ \forall \mfy \in \cY $, let $ \mfx_{\mfy} \in X_{E}(\theta) $ be the nearest point to $ \mfy $. By Proposition \ref{weight_convex}, there exists a $ w_{\mfy} \in \mathscr{W}_{p} $ such that $ \mfx_{\mfy} = S(w_{\mfy},\theta) $. Let $ w^{NK}_{\mfy} $ be the nearest one to $ w_{\mfy} $ among the weight samples $ \{w_{k}\}_{k \in [K]} $. Then,
	\begin{align}\label{M_J uniform convergence1}
	\begin{array}{llll}
	M_{K}(\theta) & = \bE\bigg( l_{K}(\mfy,\theta) \bigg) \vspace{1mm} \\
	& \leq \bE\bigg( \lVert \mfy - S(w^{NK}_{\mfy},\theta) \rVert_{2}^{2}\bigg) \vspace{1mm} \\
	& = \bE\bigg( \lVert \mfy - S(w_{\mfy},\theta) \rVert_{2}^{2}\bigg) + \bE\bigg( \lVert S(w_{\mfy},\theta) - S(w^{NK}_{\mfy},\theta) \rVert_{2}^{2}\bigg) \vspace{1mm} \\
	& \;\;\;\;+ 2\bE\bigg( \big\langle \mfy - S(w_{\mfy},\theta), S(w_{\mfy},\theta) - S(w^{NK}_{\mfy},\theta) \big\rangle \bigg)  \vspace{1mm}\\
	& \leq \bE\bigg( \lVert \mfy - S(w_{\mfy},\theta) \rVert_{2}^{2}\bigg) + \bE\bigg( \lVert S(w_{\mfy},\theta) - S(w^{NK}_{\mfy},\theta) \rVert_{2}^{2}\bigg) \vspace{1mm} \\
	& \;\;\;\;+ 2\bE\bigg( \lVert \mfy - S(w_{\mfy},\theta) \rVert_{2}\lVert S(w_{\mfy},\theta) - S(w^{NK}_{\mfy},\theta) \rVert_{2} \bigg)  \;\;\;\; \text{(Cauchy Schwarz inequality)}  \vspace{1mm}\\
	& = M(\theta) + \bE\bigg( \lVert S(w_{\mfy},\theta) - S(w^{NK}_{\mfy},\theta) \rVert_{2}^{2}\bigg) \vspace{1mm} \\
	& \;\;\;\;+ 2\bE\bigg( \lVert \mfy - S(w_{\mfy},\theta) \rVert_{2}\lVert S(w_{\mfy},\theta) - S(w^{NK}_{\mfy},\theta) \rVert_{2} \bigg),
	\end{array}
	\end{align}
	where the first inequality is due to the fact that $ l_{K}(\mfy,\theta) = \min_{k \in [K]}\{\norm{\mfy - \mfx_{k}}^{2}: \mfx_{k} = S(w_{k},\theta)\} \leq \lVert \mfy - S(w^{NK}_{\mfy},\theta) \rVert_{2}^{2} $.
	
	Let $ A_{K} :=  \sup_{\mfy \in \cY, \theta \in \Theta}\lVert S(w_{\mfy},\theta) - S(w^{NK}_{\mfy},\theta) \rVert_{2}$. Then,
	\begin{align}\label{M_J uniform convergence2}
	\bE\bigg( \lVert S(w_{\mfy},\theta) - S(w^{NK}_{\mfy},\theta) \rVert_{2}^{2}\bigg) \leq A_{K}^{2}.
	\end{align}
	
	Moreover,
	\begin{align}\label{M_J uniform convergence3}
	\begin{array}{llll}
	\bE\bigg( \lVert \mfy - S(w_{\mfy},\theta) \rVert_{2}\lVert S(w_{\mfy},\theta) - S(w^{NK}_{\mfy},\theta) \rVert_{2} \bigg) & \leq A_{K}\bE\bigg( \lVert \mfy - S(w_{\mfy},\theta) \rVert_{2} \bigg) \vspace{1mm} \\
	& \leq A_{K}\bE\bigg( \lVert \mfy \rVert_{2} + \norm{S(w_{\mfy},\theta) }  \bigg) \vspace{1mm} \\
	& \leq A_{K}\bE\bigg( \lVert \mfy \rVert_{2} + B  \bigg).\\
	\end{array}
	\end{align}
	
	Note that $ \bE\bigg( \lVert \mfy \rVert_{2} + B  \bigg) $ in \eqref{M_J uniform convergence3} is a finite number under our assumptions. Putting \eqref{M_J uniform convergence2} and \eqref{M_J uniform convergence3} into \eqref{M_J uniform convergence1}, and further noticing that $ M_{K}(\theta) \geq  M(\theta) $ by part (a) of Lemma \ref{lemma:M_J monotone}, we have
	\begin{align}\label{M_J uniform convergence4}
	0 \leq M_{K}(\theta) - M(\theta) \leq A_{K}\bigg(A_{K} + 2B + 2\bE\big( \lVert \mfy \rVert_{2} \big)\bigg).
	\end{align}
	
	By \eqref{M_J uniform convergence4}, we will conclude the proof if we can show $ A_{K} \longrightarrow 0  $ in $ K $. By Lemma \ref{lemma:lpshitz of S(w,theta)},
	\begin{align}\label{M_J uniform convergence5}
	A_{K} \leq \frac{2L}{\lambda}\sup_{\mfy \in \cY}\norm{w_{\mfy} - w^{NK}_{\mfy}}.
	\end{align}
	
	\eqref{M_J uniform convergence5} implies that we only need to show $ \lVert w_{\mfy} -w^{NK}_{\mfy} \rVert_{2}^{2}  \longrightarrow 0  $ in $ K $ for any $ \mfy \in \cY $. It suffices to show that given any $ w \in \mathscr{W}_{p} $, the nearest $ w_{k} $ to $ w $ among $ \{w_{k}\}_{k \in [K]} $ can be arbitrarily small as $ K \rightarrow \infty $. This is readily satisfied since we evenly sample $ \{w_{k}\}_{k \in [K]} $ from $ \mathscr{W}_{p} $.
	\Halmos\endproof
	
	Next, we present a very mild assumption to bound random observations.
	\begin{assumption}\label{bounded-data}
		The support $ \cY $ of the distribution $ \mfy $ is contained within a ball of radius $ R $ almost surely, where $ R < \infty $. That is, $ \bP(\norm{\mfy} \leq R) = 1	$.
	\end{assumption}
	
	\begin{proposition}[Uniform convergence of $ M_{K}^{N}(\theta) $ in  $ K $]\label{prop:M_J^N to M^N}
		Suppose Assumptions \ref{convex_setting} - \ref{bounded-data} hold. If $ \mathbf{f}(\mfx,\theta) $ is strongly convex in $ \mfx $ for each $ \theta \in \Theta $, then $ M_{K}^{N}(\theta) $ uniformly converges to $  M(\theta) $ in $ K $ for $ \theta \in \Theta $ and $ N $. That is, $ \forall N $, $ \sup\limits_{\theta \in \Theta}|M_{K}^{N}(\theta)- M^{N}(\theta)| \overset{p}{\longrightarrow} 0 $.
	\end{proposition}
	\proof{Proof. }
	We use notations here similar to those in Proposition \ref{prop:M_J - M}. We have
	\begin{align}\label{lemma:M_J^N to M^I1}
	\begin{array}{llll}
	M_{K}^{N}(\theta) & = \frac{1}{N}\sum\limits_{i \in [N]} \min\limits_{k \in [K]}\lVert \mfy_{i} - \mfx_{k}\rVert_{2}^{2} \vspace{1mm} \\
	& \leq  \frac{1}{N}\sum\limits_{i \in [N]}\lVert \mfy_{i} - S(w^{NK}_{\mfy_{i}},\theta) \rVert_{2}^{2} \vspace{1mm} \\
	& =  \frac{1}{N}\sum\limits_{i \in [N]}\lVert \mfy_{i} - S(w_{\mfy_{i}},\theta) \rVert_{2}^{2} + \frac{1}{N}\sum\limits_{i \in [N]} \lVert S(w_{\mfy_{i}},\theta) - S(w^{NK}_{\mfy_{i}},\theta) \rVert_{2}^{2} \vspace{1mm} \\
	& \;\;\;\;+ \frac{2}{N}\sum\limits_{i \in [N]} \big\langle \mfy_{i} - S(w_{\mfy_{i}},\theta), S(w_{\mfy_{i}},\theta) - S(w^{NK}_{\mfy_{i}},\theta) \big\rangle  \vspace{1mm}\\
	& \leq \frac{1}{N}\sum\limits_{i \in [N]} \lVert \mfy_{i} - S(w_{\mfy_{i}},\theta) \rVert_{2}^{2} + \frac{1}{N}\sum\limits_{i \in [N]} \lVert S(w_{\mfy_{i}},\theta) - S(w^{NK}_{\mfy_{i}},\theta) \rVert_{2}^{2} \vspace{1mm} \\
	& \;\;\;\;+ \frac{2}{N}\sum\limits_{i \in [N]} \lVert \mfy_{i} - S(w_{\mfy_{i}},\theta) \rVert_{2}\lVert S(w_{\mfy_{i}},\theta) - S(w^{NK}_{\mfy_{i}},\theta) \rVert_{2}   \;\;\;\; \text{(Cauchy Schwarz inequality)}.
	\end{array}
	\end{align}
	
	Moreover, by part (b) of Lemma \ref{lemma:M_J monotone}, we have $ M_{K}^{N}(\theta) - M^{N}(\theta) \geq 0$. To this end, through a similar argument as in the proof of Proposition \ref{prop:M_J - M}, we have
	\begin{align}\label{lemma:M_J^N to M^I4}
	0 \leq M_{K}^{N}(\theta) - M^{N}(\theta) \leq A_{K}\bigg(A_{K} + 2B + 2R\bigg),
	\end{align}
	where the last inequality follows from the fact that $ \max_{i \in [N],\theta \in \Theta}\lVert S(w_{\mfy_{i}},\theta) - S(w^{NK}_{\mfy_{i}},\theta) \rVert_{2} \leq A_{K} $.
	
	The remaining proof is exactly the same as that of Proposition \ref{prop:M_J - M}.
	\Halmos\endproof
	
	We would like to point out that previous four convergence results are provided merely for theoretical understanding as neither the distribution of $\mfy$ or the efficient set $X_E(\theta)$ is available in practice. Nevertheless, they serve as the bridge to prove the uniform convergence of the numerically computable one of $ M_{K}^{N}(\theta) $ to the abstract concept of $M(\theta)$. Before establishing the formal proof, we introduce one definition to support our convergence analysis with respect to both $N$ and $K$.
	
	\begin{definition}[Double-index convergence]
		Let $ \{X_{mn}\} $ be an array of double-index random variables. Let $ X $ be a random variable. If $ \forall \delta > 0, \forall \epsilon >0 $, $ \exists N $, s.t. $ \forall m, n \geq N $, $ \bP(|X_{mn} - X| > \epsilon) < \delta $.
		Then $ X_{mn} $ is said to converge in probability to $ X $ (denoted by $ X_{mn} \overset{P}{\dashrightarrow} X $).
	\end{definition}
	
	\begin{proposition}[Uniform convergence of $ M_{K}^{N}(\theta) $ in $ N $ and $ K $]\label{prop:M_J^N - M}
		Under the same conditions of Proposition \ref{prop:M_J^N to M^N}, $ M_{K}^{N}(\theta) $ uniformly converges to $  M(\theta) $ in $ N $ and $ K $ for all $ \theta \in \Theta $. That is,
		\begin{align*}
		\sup\limits_{\theta \in \Theta}|M_{K}^{N}(\theta)- M(\theta)| \overset{P}{\dashrightarrow} 0.
		\end{align*}
	\end{proposition}
	\proof{Proof. }
	$ \forall \theta \in \Theta $, $ |M_{K}^{N}(\theta)- M(\theta)| \overset{P}{\dashrightarrow} 0  $ if and only if $ \forall \delta > 0, \forall \epsilon >0 $, $ \exists J $, s.t. $ \forall N,K \geq J $,
	\begin{align}\label{prop:M_J^N uniform convergence1}
	\bP(|M_{K}^{N}(\theta)- M(\theta)| > \epsilon) < \delta.
	\end{align}
	
	To prove the above statement, we first note that
	\begin{align}\label{prop:M_J^N uniform convergence2}
	\begin{array}{llll}
	\bP(|M_{K}^{N}(\theta)- M(\theta)| > \epsilon) & = \bP(|M_{K}^{N}(\theta)- M^{N}(\theta) + M^{N}(\theta) - M(\theta)| > \epsilon) \vspace{1mm} \\
	& \leq \bP(|M_{K}^{N}(\theta)- M^{N}(\theta) | + |M^{N}(\theta) - M(\theta)| > \epsilon) \vspace{1mm} \\
	& \leq \bP(|M_{K}^{N}(\theta)- M^{N}(\theta)| > \epsilon/2) + \bP(|M^{N}(\theta)- M(\theta)| > \epsilon/2). \\
	\end{array}
	\end{align}
	
	For the first term on the last line of \eqref{prop:M_J^N uniform convergence2}, by Proposition \ref{prop:M_J^N to M^N}, $ \exists K_{1} $, s.t. $ \forall K \geq K_{1} $, $ \forall N $,
	\begin{align}\label{prop:M_J^N uniform convergence3}
	\bP(|M_{K}^{N}(\theta)- M^{N}(\theta)| > \epsilon/2) < \delta/2.
	\end{align}
	
	For the second term on the last line of \eqref{prop:M_J^N uniform convergence2}, by Proposition \ref{prop:M^{N} - M}, $ \exists N_{1} $, s.t. $ \forall N \geq N_{1} $,
	\begin{align}\label{prop:M_J^N uniform convergence4}
	\bP(|M^{N}(\theta)- M(\theta)| > \epsilon/2) < \delta/2.
	\end{align}
	
	Now, let $ J = \max\{N_{1},K_{1}\} $. Putting \eqref{prop:M_J^N uniform convergence3} and \eqref{prop:M_J^N uniform convergence4} in \eqref{prop:M_J^N uniform convergence2}, we have $ \forall N,K \geq J $,
	\begin{align}
	\bP(|M_{K}^{N}(\theta)- M(\theta)| > \epsilon) < \delta.
	\end{align}
	
	Hence, we complete the proof.
	\Halmos\endproof
	
	\subsection{Risk Consistency of the Estimators}
	
	We denote $ \Theta^{*} $ the set of parameters that minimizes the risk and refer to it as the optimal set. Namely, $ \Theta^{*} = \{\theta^{*} \in \Theta: M(\theta^{*}) = \min_{\theta \in \Theta} M(\theta)\} $.
	To this end, we can prove risk consistency.
	
	\begin{theorem}[Consistency of \ref{saa-inverse-model}]\label{prediction-consistency}
		Suppose Assumptions \ref{convex_setting} - \ref{set-continuous-assumption} hold. If $ \mathbf{f}(\mfx,\theta) $ is strictly convex in $ \mfx $ for each $ \theta \in \Theta $, then $ M(\hat{\theta}^{N}) \overset{p}{\longrightarrow} M(\theta^{*}) $.
	\end{theorem}
	\proof{Proof. }	
	Let $ \theta^{*} \in \Theta^{*} $, and $\hat{\theta}^{N} \in \arg\min\{M^{N}(\theta):\theta \in \Theta\} $. Then, $ M(\hat{\theta}^{N}) - M(\theta^{*}) \geq 0 $. Also,
	\begin{align*}
	M(\hat{\theta}^{N}) - M(\theta^{*}) &= M(\hat{\theta}^{N}) - M^{N}(\hat{\theta}^{N}) + M^{N}(\hat{\theta}^{N}) - M(\theta^{*}) \\
	& \leq M(\hat{\theta}^{N}) - M^{N}(\hat{\theta}^{N}) +  M^{N}(\theta^{*})- M(\theta^{*}) \\
	& \leq 2\sup\limits_{\theta \in \Theta}|M^{N}(\theta)- M(\theta)|,
	\end{align*}
	where the first inequality follows the fact that $ M^{N}(\hat{\theta}^{N}) \leq M^{N}(\theta^{*}) $.
	
	Hence, applying Proposition \ref{prop:M^{N} - M} yields that $ M(\hat{\theta}^{N}) - M(\theta^{*}) \overset{p}{\longrightarrow} 0 $.
	\Halmos\endproof
	
	Theorem \ref{prediction-consistency} states that $ \hat{\theta}^{N} $ converges in probability to one point in the optimal set $ \Theta^{*} $.

	\begin{theorem}[Consistency of \ref{ws-general-inverse-model}]\label{theorem: M(theta_J) - M}
		Suppose Assumptions \ref{convex_setting} - \ref{set-continuous-assumption} hold. If $ \mathbf{f}(\mfx,\theta) $ is strongly convex in $ \mfx $ for each $ \theta \in \Theta $, then $ M(\hat{\theta}_{K}) \overset{P}{\longrightarrow} M(\theta^{*}) $.
	\end{theorem}
	Proof of Theorem \ref{theorem: M(theta_J) - M} is essentially the same to that of Theorem \ref{prediction-consistency}, and is omitted.
	
	Theorem \ref{theorem: M(theta_J) - M} indicates that $ \hat{\theta}_{K} $ also converges in probability to one point in the optimal set $ \Theta^{*} $.

	Recall that \ref{saa-general-inverse-model} is the only one we can and will solve to infer the unknown parameters of a decision making problem among the four models listed in Table \ref{table:four models}. Thus, the following theorem is the most important one from the perspective of computation.
	\begin{theorem}[Consistency of \ref{saa-general-inverse-model}]\label{theorem: M(theta_J^N) - M}
		Suppose Assumptions \ref{convex_setting} - \ref{bounded-data} hold. If $ \mathbf{f}(\mfx,\theta) $ is strongly convex in $ \mfx $ for each $ \theta \in \Theta $, then $ M(\hat{\theta}_{K}^{N}) \overset{P}{\dashrightarrow} M(\theta^{*}) $.
	\end{theorem}
	Proof of Theorem \ref{theorem: M(theta_J^N) - M} is essentially the same to those of Theorems \ref{prediction-consistency} and \ref{theorem: M(theta_J^N) - M}, and is omitted.
	
	Similar to Theorems \ref{prediction-consistency} - \ref{theorem: M(theta_J) - M}, Theorem \ref{theorem: M(theta_J^N) - M} indicates that $ \hat{\theta}^{N}_{K} $ converges in probability to one point in the optimal set $ \Theta^{*} $. Actually, as we will see in EXAMPLE \ref{ex:counter-ex1} and \ref{ex:counter-ex2}, if no information about decision makers' preference or partial understanding on $\theta$ is imposed, the optimal set $ \Theta^{*} $ is often not a singleton even when the objective functions are strongly convex. This indicates one challenge of parameter inference through inverse multiobjective optimization. With such an observation, the risk consistency, or persistence in \citet{greenshtein2004persistence}, is a more realistic standard for the estimator when learning parameters through solving IMOP.

	\subsection{Generalization Bound of \ref{saa-general-inverse-model}} \label{ws-consistency}
	
	For fixed weight samples $ \{w_{k}\}_{k \in [K]} $, we want to estimate the risk $ M_{K}(\hat{\theta}^{N}_{K})$ as it quantifies how well the performance of our estimator $ \hat{\theta}^{N}_{K} $ generalizes to the unseen data. However, this quantity cannot be obtained since the distribution $ \bP_{\mfy} $ is unknown, and thus is a random variable (since it depends on the data). Hence, one way to make a statement about this quantity is to say how it relates to an estimate such as the empirical risk $ M_{K}^{N}(\hat{\theta}^{N}_{K}) $. Before providing the main theorem, we first introduce some important definitions and lemmas.
	
	\begin{definition}[Rademacher random variables]
		Random variables $ \sigma_{1},\ldots,\sigma_{N} $ are called \textit{Rademacher random variables} if they are independent, identically distributed and $ \bP(\sigma_{i} = 1) = \bP(\sigma_{i} = -1) = 1/2 $ for $ i \in [N] $.
	\end{definition}
	Let $ \mathcal{F} $ be a class of functions mapping from $ Z $ to $ [a,b] $, and $ Z_{1},\ldots,Z_{N} $ be independent and identically distributed (i.i.d.) random variables on $ Z $.
	\begin{definition}
		The Rademacher complexity of $ \mathcal{F}  $ is
		\begin{align*}
		Rad_{N}(\mathcal{F} ) =  \frac{1}{N}\bE\left[\sup_{f \in \mathcal{F} }\sum_{i \in [N]}\sigma_{i}f(Z_{i})\right],
		\end{align*}
		where the expectation is taken over $ \sigma $ and $ Z_{1},\ldots,Z_{N}$.
	\end{definition}
	
	Intuitively, $ Rad_{N}(\mathcal{F} ) $ is large if one can find function $ f \in \mathcal{F} $ that look like random noise, that is, these functions are highly correlated with Racemacher random variables $ \sigma_{1},\ldots,\sigma_{N} $.
	
	\begin{lemma}
		\label{guarantee: risk-bounds}
		Let $ \mathcal{F}  $ be a class of functions mapping from $ Z $ to $ [a,b] $. Let $ Z_{1},\cdots,Z_{N} $ be i.i.d. random variables on $ Z $. Then, for any $ 0 < \delta <1 $, with probability at least $ 1 - \delta $, every $ f \in \mathcal{F} $ satisfies
		\begin{align*}
		\bE[f(Z)] \leq \frac{1}{N}\sum_{i \in [N]}f(Z_{i}) + 2Rad_{N}(\mathcal{F} ) + (b - a)\sqrt{\frac{log(1/\delta)}{2N}}.
		\end{align*}
	\end{lemma}
	
	
	\begin{remark}
		The last term of the inequality in Lemma \ref{guarantee: risk-bounds} might not be tight. We are able to obtain tighter bounds using more complex methods such as the one in \citet{bartlett2002rademacher}. We refer the reader to \citet{vapnik2013nature,bousquet2004introduction} for detailed introductions on how to characterize the generalization bound that the estimators may have in given situations.
	\end{remark}
	
	Given $ K $ and $ \theta $, we define a function $ f(\cdot,\theta) $ by $ f(\mfy,\theta) = \min\limits_{k \in [K]}\lVert \mfy - \mfx_{k}\rVert_{2}^{2} $, where $ \mfx_{k} \in S(w_{k},\theta) $ for all $ k \in [K] $. Now consider the class of functions $ \mathcal{F} = \{f(\cdot,\theta) : \theta \in \Theta\} $. To bound the risk $ \bE[f(\mfy,\theta)] $ using Lemma \ref{guarantee: risk-bounds}, we need to either compute the vaule of $ Rad_{N}(\mathcal{F} ) $ or find an upper bound of it. Note that the computation of $ Rad_{N}(\mathcal{F} ) $ involves solving a difficult optimization problem over $ \mathcal{F}  $. In contrast, obtaining a bound of $ Rad_{N}(\mathcal{F} ) $ is relatively easier. Therefore, we seek to bound $ Rad_{N}(\mathcal{F} ) $ in the following lemma.
	
	\begin{lemma}
		\label{guarantee: rademacher-complexity}
		The Rademacher complexity of $ \mathcal{F}  $ is bounded by a function of sample size $ N $,
		\begin{align*}
		Rad_{N}\big(\mathcal{F}\big) \leq \frac{K}{\sqrt{N}}\bigg(B^{2} + 2BR\bigg).
		\end{align*}
	\end{lemma}

	We are now ready to state the main result in this section.
	
	\begin{theorem}[Generalization bound]\label{theorem:risk-bound}
		Suppose Assumptions \ref{convex_setting} - \ref{bounded-data} hold. For any $ 0 < \delta < 1 $, with probability at least $ 1 - \delta $ with respect to the observations,
		\begin{align*}
		M_{K}(\hat{\theta}^{N}_{K}) \leq M_{K}^{N}(\hat{\theta}^{N}_{K}) + \frac{1}{\sqrt{N}}\bigg( 2K(B^{2}+2BR) + (B+R)^{2}\sqrt{\log(1/\delta)/2} \bigg)\;\; \text{for each}\;\; K.
		\end{align*}
	\end{theorem}
	\proof{Proof. }
	We specialize Lemmas \ref{guarantee: risk-bounds} and \ref{guarantee: rademacher-complexity} to prove the theorem. Note that
	\begin{align*}
	0 \leq f(\mfy,\theta) = \min\limits_{k \in [K]}\lVert \mfy - \mfx_{k}\rVert_{2}^{2} \leq (B+R)^{2}.
	\end{align*}
	
	Let $ a = 0, b = (B+R)^{2} $ in Lemma \ref{guarantee: risk-bounds}. Then, combining the results in Lemmas \ref{guarantee: risk-bounds} and \ref{guarantee: rademacher-complexity} yields this theorem.
	\Halmos\endproof
	
	Essentially, this theorem indicates that the risk of the estimator constructed by solving \ref{saa-general-inverse-model}, which can be seen as the test error for fixed weight samples $ \{w_{k}\}_{k \in [K]} $, is no worse than the empirical risk, which can be seen as the training error, by an additional term that is of $ \mathcal{O}(1/\sqrt{N}) $.

	\section{Identifiability Analysis for IMOP} \label{section:identifiability}
	In this section, we propose the concept of identifiability in the context of decision making problems with multiple objectives, and show its strong correlation with the performance of our inverse multiobjective optimization model. 
	
	\begin{definition}[Hausdorff semi-distance]
		Let $ X $ and $ Y $ be two nonempty set. We define their \textit{Hausdorff semi-distance} by
		\begin{align*}
		d_{sH}(X,Y) = \sup_{x \in X}\inf_{y \in Y}d(x,y).
		\end{align*}
	\end{definition}
	
	Clearly, $ d_{sH}(X,Y) = 0 $ if $ X = Y $. Nevertheless, $ d_{sH}(X,Y) = 0 $ does not always lead to $ X = Y $.
	\begin{lemma}
		$ d_{sH}(X,Y) = 0 $ if and only if $ X \subseteq Y $.
	\end{lemma}
	\proof{Proof. }	
	Sufficiency: $ d_{sH}(X,Y) = 0 $ implies that $ \inf_{y \in Y}\norm{x - y} = 0, \forall x \in X $. That is, $ \exists y \in Y $, st. $ x = y $. Hence, $ X \subseteq Y $. Necessity: $ X \subseteq Y $ implies that $ \forall x \in X $, $ \exists y \in Y $, s.t. $ y = x $. Thus, $ \inf_{y \in Y}\norm{x - y} = 0 $. Therefore, $ d_{sH}(X,Y) = 0 $.
	\Halmos\endproof
	
	We are now ready to state our definition of Identifiability in the context of \ref{mop}.
	\begin{definition}[Identifiability]
		A \ref{mop} is said to be identifiable at $ \theta \in \Theta $, if for all $ \theta' \in \Theta \setminus\theta $,
		\begin{align*}
		d_{sH}(X_{E}(\theta), X_{E}(\theta')) > 0.
		\end{align*}
	\end{definition}
	
	Intuitively, a \ref{mop} is identifiable if its efficient set can not be covered by that of any other DMP with parameter in $ \Theta $. More precisely, $ X_{E}(\theta) $ is not a subset of $ X_{E}(\theta') $ for any $ \theta' \in \Theta \setminus\theta $.
	
	\subsection{Estimation Consistency of IMOP under Identifiability}
	
	Let $ \theta_{0} $ be the underlying parameter of the DMP that generates the data. If DMP is identifiable at $ \theta_{0} $, and the data is not corrupted by noise, then $ M(\theta) $ achieves its minimum uniquely at $ \theta_{0} $. We are now ready to state our result regarding the estimation consistency of $ \hat{\theta}_{K}^{N} $.
	\begin{theorem}[Consistency of $ \hat{\theta}_{K}^{N} $]\label{theorem:estimation consistency}
		Suppose Assumptions \ref{convex_setting} - \ref{set-continuous-assumption} hold. Suppose also that $ \mathbf{f}(\mfx,\theta) $ is strongly convex in $ \mfx $ for each $ \theta \in \Theta $, and that $ \forall \mfy \in \mathcal{Y}, \mfy \in X_{E}(\theta_{0}) $. That is, there is no noise in the data. If \ref{mop} is identifiable at $ \theta_{0} \in \Theta $, then $ \hat{\theta}^{N}_{K} \overset{P}{\dashrightarrow} \theta_{0} $.
	\end{theorem}
	\proof{Proof.}
	First, we show that $ \theta_{0} $ minimizes $ M(\theta) $ among $ \Theta $. This is readily true since $ M(\theta_{0}) = 0 $ by noting that there is no noise in the data. By Theorem \ref{theorem: M(theta_J^N) - M}, a direct result is $ M(\hat{\theta}_{K}^{N}) \overset{P}{\dashrightarrow} M(\theta_{0}) = 0 $. Second, we show that $ \theta_{0} $ is the unique solution that minimizes $ M(\theta) $ among $ \Theta $. $ \forall \theta' \in \Theta \setminus\theta $, $ M(\theta) = \bE_{\mfy \in X_{E}(\theta_{0})}\big(\min_{\mfx \in X_{E}(\theta)} \norm{\mfy - \mfx}^{2}\big) > 0 $ as $ d_{sH}(X_{E}(\theta), X_{E}(\theta')) > 0 $. Consequently, we have $ M(\theta) > M(\theta_{0}) = 0 $. Finally, since \ref{mop} is identifiable at $ \theta_{0} $, then $ \forall \epsilon > 0 $, $ \exists \delta > 0 $, s.t. $ M(\theta) - M(\theta_{0}) > \delta $ for every $ \theta $ with $ d(\theta,\theta_{0}) > \epsilon $. Thus, the event  $ \{d(\hat{\theta}^{N}_{K},\theta_{0}) > \epsilon\} $ is contained in the event $ \{M(\hat{\theta}^{N}_{K}) - M(\theta_{0}) > \delta\} $. Namely, $ \bP(d(\hat{\theta}^{N}_{K},\theta_{0}) > \epsilon) \leq \bP(M(\hat{\theta}^{N}_{K}) - M(\theta_{0}) > \delta) $. We complete the proof by noting that the probability of the right term converges to $ 0 $ as $ M(\hat{\theta}_{K}^{N}) \overset{P}{\dashrightarrow} M(\theta_{0}) $.
	\Halmos\endproof
	
	On top of the ability of inferring parameters in \ref{mop}, we would like to point out that our inverse model has an additional benefit of learning the distribution of decision makers' preferences.
	
	By solving \ref{saa-general-inverse-model}, note that we obtain not only an estimation of $\theta$ and $\{\mfx_{k}\}_{k\in[K]}$, but also the value of $ z_{ik} $ for each $i\in[N]$ and $k\in[K]$. We group all those noisy decisions with $z_{ik}=1$ among $\{\mfy_{i}\}_{i\in[N]}$ to the cluster $C_{k}$ for each $k\in[K]$. For the cluster $ C_{k} $, all the noisy decisions share the same preference over objective functions. More precisely, we let $ w_{k} $, the $k$th weight sample, represent the preference of the decision makers in $ C_{k} $ over multiple objective functions. Here, one latent assumption we make is that decision makers in the same cluster are homogeneous in their preferences for different objectives. Next, we propose the concept of the bijectivity of a \ref{mop} to support the performance analysis of the inferred preference.
	
	\begin{definition}[Bijectivity]
		A \ref{mop} is said to be bijective at $ \theta \in \Theta $ if $X_{E}(\theta) = \bigcup_{w\in \mathscr{W}_{p}}S(w,\theta)$, $S(w,\theta)$ is single valued for $w$ almost surely, and $\forall w_{1}, w_{2} \in \mathscr{W}_{p}$, $w_{1} \neq w_{2}$ implies $S(w_{1},\theta) \neq S(w_{2},\theta)$.
	\end{definition}
	
	With a slight abuse of notation, we let $ w_{\mfy} $ be the true weight for $ \mfy $, and $ w^{NK}_{\mfy} $ be the estimated weight for $ \mfy $ given $ \hat{\theta}_{K}^{N} $. More precisely, $ w^{NK}_{\mfy} = \argmin_{w_{k}:k \in [K]}\{l_{K}(\mfy,\hat{\theta}_{K}^{N})\} $. The following theorem shows that the inferred preference converges in probability to the true preference if the \ref{mop} we investigate enjoys the identifiability and the bijectivity defined above.
	
	\begin{theorem}[Consistency of $ w^{NK}_{\mfy} $]\label{theorem:weight convergence}
		Suppose the same conditions of Theorem \ref{theorem:estimation consistency} hold. If \ref{mop} is bijective at $ \theta_{0} $, then $ \norm{w_{\mfy} - w^{NK}_{\mfy}} \overset{P}{\dashrightarrow} 0 $ for $ \mfy \in \mathcal{Y} $ almost surely.
	\end{theorem}
	\proof{Proof.}
	First, note that
	\begin{align}\label{prooftheorem:weight convergence1}
	\begin{array}{llll}
	\norm{S(w^{NK}_{\mfy}, \theta_{0}) - S(w_{\mfy},\theta_{0})} & = \norm{S(w^{NK}_{\mfy}, \theta_{0}) - S(w^{NK}_{\mfy}, \hat{\theta}^{N}_{K}) + S(w^{NK}_{\mfy}, \hat{\theta}^{N}_{K}) - S(w_{\mfy},\theta_{0})} \vspace{1mm} \\
	& \leq \norm{S(w^{NK}_{\mfy}, \theta_{0}) - S(w^{NK}_{\mfy}, \hat{\theta}^{N}_{K})} + \norm{S(w^{NK}_{\mfy}, \hat{\theta}^{N}_{K}) - S(w_{\mfy},\theta_{0})}.
	\end{array}
	\end{align}
	
	By Theorem \ref{theorem:estimation consistency}, we have $ \hat{\theta}^{N}_{K} \overset{P}{\dashrightarrow} \theta_{0} $. Note that $ S(w,\theta) $ is continuous in $ \theta \in \Theta $. By continuous mapping theorem, the first term in the last line of \eqref{prooftheorem:weight convergence1} $ \norm{S(w^{NK}_{\mfy}, \theta_{0}) - S(w^{NK}_{\mfy}, \hat{\theta}^{N}_{K})} \overset{P}{\dashrightarrow} 0 $.
	
	By the argument in the proof of Theorem \ref{theorem:estimation consistency}, the second term in the last line of \eqref{prooftheorem:weight convergence1} $ \norm{S(w^{NK}_{\mfy}, \hat{\theta}^{N}_{K}) - S(w_{\mfy},\theta_{0})} \overset{P}{\dashrightarrow} 0 $ almost surely. Otherwise, $ M(\hat{\theta}^{N}_{K}) = \bE_{\mfy \in X_{E}(\theta_{0})}\big(\min_{\mfx \in X_{E}(\hat{\theta}^{N}_{K})} \norm{\mfy - \mfx}^{2}\big) = \bE_{\mfy \in X_{E}(\theta_{0})}\norm{S(w^{NK}_{\mfy}, \hat{\theta}^{N}_{K}) - S(w_{\mfy},\theta_{0})}^{2} > 0 $, and thus will not converge to $ M(\theta_{0}) $.
	
	Putting the above two results into \eqref{prooftheorem:weight convergence1} yields $ \norm{S(w^{NK}_{\mfy}, \theta_{0}) - S(w_{\mfy},\theta_{0})} \overset{P}{\dashrightarrow} 0 $ almost surely.
	
	Next, note that $ S(w,\theta_{0}) $ is continuous in $ w $, and that $ MOP(\theta_{0}) $ is bijective. Then, we have that $ S(\cdot,\theta_{0}): \mathscr{W}_{p} \rightarrow X_{E}(\theta_{0}) $ is a one-to-one correspondence. Thus, $ S(\cdot,\theta_{0}) $ is a homeomorphism by the inverse mapping theorem \citep{sutherland2009introduction}, meaning that the inverse map $ S^{-1}(\cdot,\theta_{0}):  X_{E}(\theta_{0}) \rightarrow \mathscr{W}_{p} $ is also continuous. Therefore, $ \norm{S(w^{NK}_{\mfy}, \theta_{0}) - S(w_{\mfy},\theta_{0})} \overset{P}{\dashrightarrow} 0 $ implies that $ \norm{w_{\mfy} - w^{NK}_{\mfy}} \overset{P}{\dashrightarrow} 0 $ by the continuous mapping theorem.
	\Halmos\endproof

	\subsection{Non-identifiability of a Decision Making Problem}
	A \ref{mop} might be non-identifiable in various ways. One trivial non-identifiability occurs due to scaling or permuting the component functions in $ \mathbf{f}(\mfx,\theta) $ or $ \mathbf{g}(\mfx, \theta) $. Nevertheless, this is not a serious problem in practice because some components of $ \mathbf{f}(\mfx,\theta) $ or $ \mathbf{g}(\mfx, \theta) $ might be known a priori, which helps avoid the occurrence of non-identifiability. Otherwise, this type of non-identifiability could be prevented by normalizing some components of the parameter before solving \ref{saa-general-inverse-model}.
	
	A more subtle non-identifiability issue occurs as shown by the following two examples.
	\vspace{10pt}
	
	\begin{minipage}{0.5\textwidth}
		\begin{example}\label{ex:counter-ex1}
			\begin{align*}
			\begin{array}{llll}
			\min & \left(
			\begin{array}{llll}
			x_{1}^{2} + 2x_{2}^{2} + 6x_{1} + 2x_{2} \\ 2x_{1}^{2} + x_{2}^{2} - 12x_{1} - 10x_{2}
			\end{array}
			\right) \vspace{1mm}\\
			\text{s.t.} & 3x_{1} - x_{2} \leq  6, \vspace{1mm}\\
			& x_{2} \leq 3, \vspace{1mm}\\
			& x_{1},x_{2} \geq 0.
			\end{array}
			\end{align*}
		\end{example}
	\end{minipage}
	\begin{minipage}{0.5\textwidth}
		\begin{example}\label{ex:counter-ex2}
			\begin{align*}
			\begin{array}{llll}
			\min & \left(
			\begin{array}{llll}
			7x_{1}^{2} + 11x_{2}^{2} + 19x_{1} \\ 12x_{1}^{2} + 6x_{2}^{2} - 72x_{1} - 60x_{2}
			\end{array}
			\right) \vspace{1mm}\\
			\text{s.t.} & 3x_{1} - x_{2} \leq  6, \vspace{1mm}\\
			& x_{2} \leq 3, \vspace{1mm}\\
			& x_{1},x_{2} \geq 0.
			\end{array}
			\end{align*}
		\end{example}
	\end{minipage}

	\begin{proposition}\label{non-uniqueness}
		EXAMPLE \ref{ex:counter-ex1} and EXAMPLE \ref{ex:counter-ex2} have the same efficient set.	
	\end{proposition}
	\proof{Proof. }	
	Since both examples are strongly convex MOPs, any efficient solution of them can be obtained by solving \ref{weighting problem} according to Proposition \ref{weight_convex}. Also, every optimal solution of the weighting problem is an efficient solution by part (b) of Proposition \ref{prop:unique-weakly}.
	
	Let $ w \in [0,1] $ be the weight of the first function. The optimal solutions for \ref{weighting problem} in Example \ref{ex:counter-ex1} can be characterized parametrically by $ w $ as
	\begin{align}\label{ex1:solution}
	x_{1}^{1}(w) =
	\begin{cases}
	\frac{6 - 9w}{2 - w}, & \text{if}\;\; 0 \leq w \leq 2/3, \\
	0, & \text{if}\;\; 2/3 < w \leq 1,
	\end{cases} \hspace{0.1in}
	x_{2}^{1}(w) =
	\begin{cases}
	3, & \text{if}\;\; 0 \leq w \leq 2/9, \\
	\frac{5 - 6w}{1 + w}, & \text{if}\;\; 2/9 < w \leq 5/6, \\
	0, & \text{otherwise}.
	\end{cases}
	\end{align}
	
	Similarly, the optimal solutions for the \ref{weighting problem} in Example \ref{ex:counter-ex2} can be characterized parametrically as
	\begin{align}\label{ex2:solution}
	x_{1}^{2}(w) =
	\begin{cases}
	\frac{36 - 45w}{12 - 5w}, & \text{if}\;\; 0 \leq w \leq 4/5, \\
	0, & \text{otherwise},
	\end{cases} \hspace{0.1in}
	x_{2}^{2}(w) =
	\begin{cases}
	3, & \text{if}\;\; 0 \leq w \leq 4/15, \\
	\frac{30 - 30w}{6 + 5w}, & \text{otherwise}.
	\end{cases}
	\end{align}
	
	We can show that $ x_{1}^{1}(w) = x_{1}^{2}(\frac{6}{5}w) $ and $ x_{2}^{1}(w) = x_{2}^{2}(\frac{6}{5}w) $ for $ 0 \leq w \leq \frac{5}{6} $. In addition, $ x_{1}^{1}(w) = x_{2}^{1}(w) = 0 $ for $ \frac{5}{6} \leq w \leq 1 $. Therefore, these parametric  points in \eqref{ex1:solution} and \eqref{ex2:solution} correspond to the same curve. Hence, EXAMPLE \ref{ex:counter-ex1} and EXAMPLE \ref{ex:counter-ex2} have the same efficient set.	
	\Halmos\endproof
	
\begin{figure}[ht]
		\centering
		\includegraphics[width=0.5\linewidth]{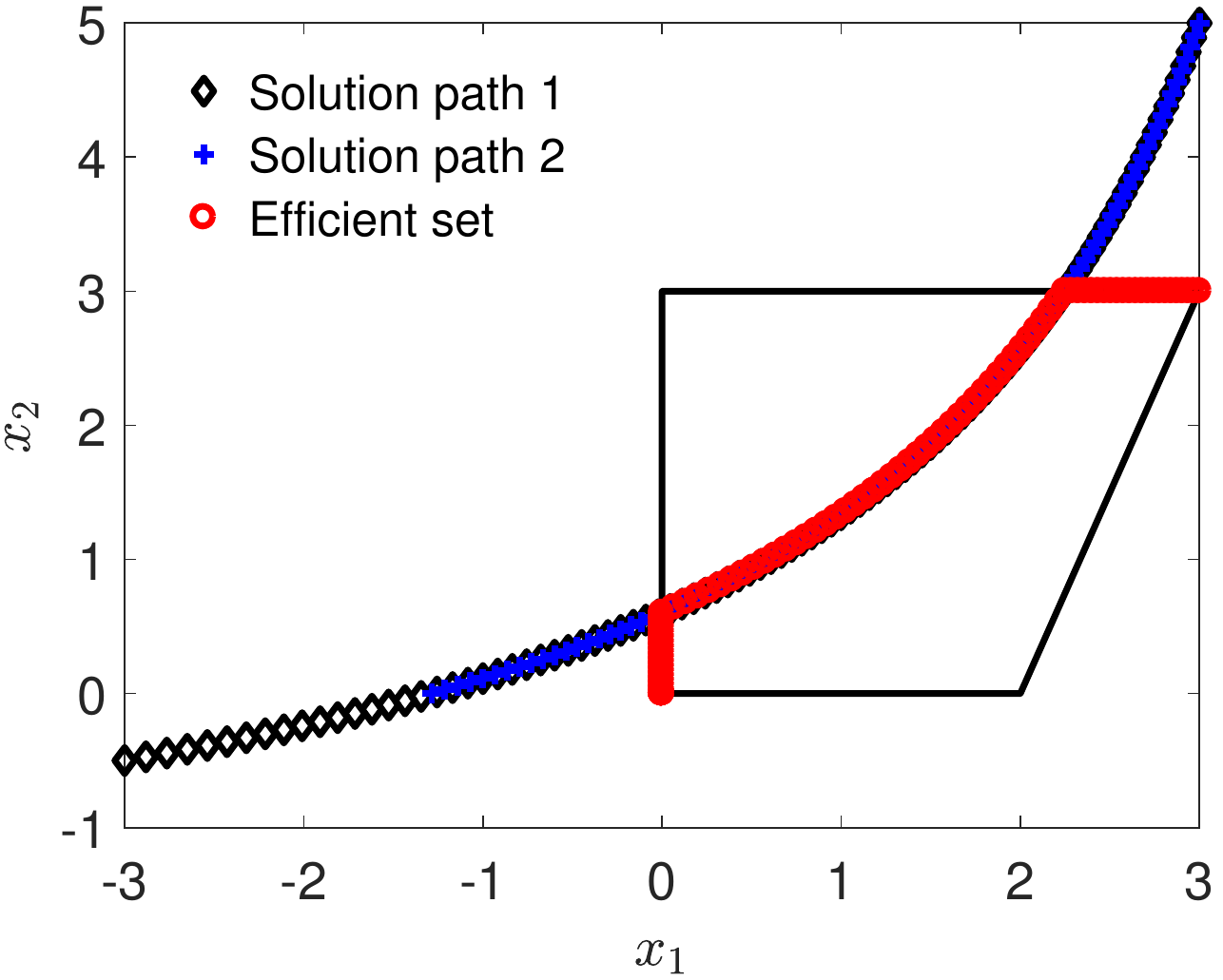}
		\caption{The black diamond dots represent the solution path of Example \ref{ex:counter-ex1}. The blue "+" dots show the solution path of Example \ref{ex:counter-ex2}. The red circle dots indicate the efficient set for both examples.}
		\label{fig:sameefficientset}
	\end{figure}

	We plot the two efficient sets in Figure \ref{fig:sameefficientset}. One can see that the two examples share the same efficient set. Suppose no restrictions on the variables $ x_{1} $ and $ x_{2} $, we obtain a set of points that consists of the optimal solution of \ref{weighting problem} for each $ w \in [0,1] $. We call it the solution path for \ref{mop}. To further illustrate why these two examples share the same efficient set, we plot the solution paths for both of them in Figure \ref{fig:sameefficientset}. It shows that solution path 2 is covered by solution path 1. Note that both solution paths have points lying outside of the feasible region. These points are rendered to become the same efficient solutions on the boundary of the feasible region, which explains why two MOPs with different solution paths have the same efficient set.
	
	%
	
	\subsection{Test Non-identifiability of a Decision Making Problem}
	
	As shown in previous section, non-identifiability of a \ref{mop} occurs in various ways, which would bring serious problems to the inference of parameters. Therefore, it is necessary to provide a systematic procedure to test whether a \ref{mop} is identifiable or not. To achieve this, we first introduce the test problem in the following.
	\leqnomode\begin{align*}
	\label{test-problem}
	\tag*{TEST-PROBLEM}
	\begin{array}{llll}
	\max\limits_{\theta \in \Theta} & \norm[1]{\theta - \hat{\theta}^{N}_{K}} \vspace{1mm}\\
	\;\text{s.t.} & \mfx_{i} \in \bigcup\limits_{k \in [K']} S(w_{k},\theta) & \forall i \in [N'], \vspace{1mm}
	\end{array}
	\end{align*}\reqnomode
	where $ \hat{\theta}^{N}_{K} $ is an optimal solution of \ref{saa-general-inverse-model}, and $ \{\mfx_{i}\}_{i \in [N]'} $ are the efficient points on $ X_{E}(\hat{\theta}^{N}_{K}) $ which could be obtained a priori by solving \ref{weighting problem} with a set of weights $ \{w_{i}\}_{i \in [N']} $. 
	
	Indeed, \ref{test-problem} seeks to find the furthest $ \theta $ to $ \hat{\theta}^{N}_{K} $ that still keeps $ X_{E}(\hat{\theta}^{N}_{K}) $ efficient. Thus, the test statistic could be the optimal value $ z_{test} $ of \ref{test-problem}, where $ z_{test}>0 $ suggests that there might exist multiple parameters keeping $ X_{E}(\hat{\theta}^{N}_{K}) $ efficient, and that \ref{mop} is non-identifiable.
	
	We need three sets of weight samples to solve \ref{test-problem}. The first set of weight samples $ \{w_{k}\}_{k \in [K]} $ is used in \ref{saa-general-inverse-model}. Once obtaining $ \hat{\theta}^{N}_{K} $, we use the second set of weight samples $ \{w_{i}\}_{i \in [N']} $ to generate the efficient points on $ X_{E}(\hat{\theta}^{N}_{K}) $. The third set of weight samples $ \{w_{k}\}_{k \in [K']} $ is used to find the furthest $ \theta $ to $ \hat{\theta}^{N}_{K} $ that keeps $ \{\mfx_{i}\}_{i \in [N']} $ efficient. These three sets of weights do not necessarily be the same. Since the weighting problem \ref{weighting problem} is a convex program and thus is the easiest one among the three problems, $ \{w_{i}\}_{i \in [N']} $ should be the largest set. In addition, \ref{saa-general-inverse-model} is the most difficult one to solve, and thus $ \{w_{k}\}_{k \in [K]} $ should be the smallest set.
	
	Suppose $ \mathbf{f}(\mfx, \theta) $ and $ \mathbf{g}(\mfx, \theta) $ are smooth in $ \mfx $, we can reformulate the \ref{test-problem} by replacing the optimal set $ S(w_{k}, \theta) $ with strong duality or its KKT conditions and using binary variables to indicate the inclusion relationship between $ \mfx_{i} $ and $ S(w_{k},\theta) $. The reformulation is given in APPENDIX \ref{app:test-problem}. The test process is formally presented in Algorithm \ref{alg:test-process}.
	\begin{algorithm}[ht]
		\caption{Test Non-identifiability of a Decision Making Problem}\label{alg:test-process}
		\begin{algorithmic}[1]
			\State  Choose weight samples $ \{w_{k}\}_{k \in [K]} $. Solve \ref{saa-general-inverse-model}. Denote $ \hat{\theta}^{N}_{K} $ the optimal solution.
			\State  Choose a new set of weight samples $ \{w_{i}\}_{i \in [N']} $. Generate $ |N'| $ efficient points on $ X_{E}(\hat{\theta}^{N}_{K}) $ by solving \ref{weighting problem}. Namely, $ \mfx_{i} \in S(w_{i},\hat{\theta}^{N}_{K}) $ for each $ i \in [N'] $.
			\State  Choose another set of weight samples $ \{w_{k}\}_{k \in [K']} $. Solve \ref{test-problem}. Let the test statistic be the optimal value $ z_{test} $.
			\State  If $ z_{test} \neq 0 $ , we believe \ref{mop} is non-identifiable based on the data.
		\end{algorithmic}
	\end{algorithm}
	
	\subsection{Eliminating Non-identifiability of a Decision Making Problem}\label{sec:dispose non-uniqueness}
	Suppose Algorithm \ref{alg:test-process} assures us that \ref{mop} is non-identifiable, there are at least three underlying reasons to explain this phenomenon. These reasons include the lack of data, information about decision maker's preference, or knowledge about the parameter. Accordingly, there exist at least three ways to tackle the non-identifiability issue.
	
	The most natural way that might help avoid the occurrence of non-identifiability is to collect more data when it is practical and economically available. Another way is to seek for additional information about the decision making process. For example, if we know all or part of the decision makers' preferences, i.e., we know the weight-decision correspondences, we could use \eqref{saa-general-inverse-model-modified} instead of \ref{saa-general-inverse-model}. Another method is to obtain more knowledge about the parameters. For example, if we seek to learn the coefficients of the term $ x_{2} $ in EXAMPLE \ref{ex:counter-ex1} given that all the other coefficients are known, solving \ref{saa-general-inverse-model} would find the true parameter. In the worst case, none of these approaches works individually, and we need to combine all of them to tackle the non-identifiability issue as is demonstrated in Section \ref{experiments}.

	\section{Solutions Approaches to \ref{saa-general-inverse-model}}\label{solution-approach}
	
	The most natural way to solve \ref{saa-general-inverse-model} is to transform it into a single level optimization problem by replacing the constraints $ \mfx_{k} \in S(w_{k},\theta) $ with optimality conditions \citep{dempe2015bilevel}. In general, there are at least three ways to achieve this. One way is to replace $ \mfx_{k} \in S(w_{k},\theta) $ by the variational inequalities, the second way is to employ the strong duality theorem of convex optimization, and the third way is to replace $ \mfx_{k} \in S(w_{k},\theta) $ by the KKT conditions. Note that the first and second ways will introduce product terms of the upper level decision variables (i.e., $ \theta $) and lower level decision variables (i.e., $ \mfx_{k} $), making the reformulated problems extremely difficult to solve. Nevertheless, the third approach would avoid such a situation since the complementary constraints in KKT conditions can be linearized. Hence, we will present our solution approaches based on the reformulations using KKT conditions.
	
	The single level reformulation of \ref{saa-general-inverse-model} is a mixed integer nonlinear program (MINLP), which is known to be extremely difficult to solve. To tackle this challenge, we develop a fast heuristic algorithm based on alternating direction method of multipliers (ADMM) and a clustering-based algorithm that guarantees to converge to a (local) optimal solution.

	\subsection{ADMM for IMOP}
	The ADMM was originally proposed in \cite{glowinski1975approximation} and \cite{gabay1976dual}, and recently revisited by \cite{boyd2011distributed}. In practice, ADMM often exhibits a substantially faster convergence rate than traditional methods in solving convex optimization problems. Characterizing the convergence rate of ADMM for convex optimization problems is still a popular research topic \citep{shi2014linear,deng2016global,Hong2017}.  Although ADMM might not converge even for convex problems with more than two blocks of variables \citep{chen2016direct}, many recent papers have numerically demonstrated the fast and appealing convergence behavior of ADMM on nonconvex problems \citep{diamond2016general,magnusson2016convergence,7926116}. Hence, we apply ADMM as a heuristic to solve the nonconvex problem \ref{saa-general-inverse-model}. 
	
	\ref{saa-general-inverse-model} is closely related to the global consensus problem discussed heavily in \citet{boyd2011distributed}, but with the important difference that \ref{saa-general-inverse-model} is a nonconvex problem. In order to use ADMM, we first partition $\{\mfy_{i}\}_{i \in [N]}$ equally into $T$ groups, and denote $ \{\mfy_{i}\}_{i\in[N_{t}]} $ the observations in $ t $-th group. Then, we introduce a set of new variables $\{\theta^{t}\}_{t \in T}$, typically called local variables, and transform \ref{saa-general-inverse-model} equivalently to the following problem:
	\begin{align}
	\label{imop:admm}
	\begin{array}{llll}
	\min\limits_{\theta \in \Theta, \theta^{t} \in \Theta } & \sum\limits_{t \in T}\sum\limits_{i \in [N_{t}]}l_{K}(\mfy_{i},\theta^{t}) \vspace{1mm} \\
	\text{s.t.} & \theta^{t} = \theta, & \forall t \in [T].
	\end{array}
	\end{align}
	
	ADMM for problem \eqref{imop:admm} can be derived directly from the augmented Lagrangian
	\begin{align*}
	L_{\rho}(\theta,\{\theta^{t}\}_{t\in[T]},\{\mfv^{t}\}_{t\in[T]}) = \sum\limits_{t \in [T]}\bigg( \sum\limits_{i\in [N_{t}]}l_{K}(\mfy_{i},\theta^{t}) + <\mfv^{t},\theta^{t}-\theta> + (\rho/2)\norm{\theta^{t} - \theta}^{2}\bigg),
	\end{align*}
	where $ \rho > 0 $ is an algorithm parameter, $ \mfv^{t} $ is the dual variable for the constraint $ \theta^{t} = \theta $.
	
	Let $ \overline{\theta}^{k} = \frac{1}{|T|}\sum_{t \in T}\theta^{t,k} $. As suggested in \citet{boyd2011distributed}, the primal and dual residuals are
	\begin{align*}
	r_{pri}^{k} = \begin{pmatrix}
	\theta^{1,k} - \overline{\theta}^{k},\ldots,
	\theta^{|T|,k} - \overline{\theta}^{k}
	\end{pmatrix}, & \hspace{5mm}
	r_{dual}^{k} = -\rho\begin{pmatrix}
	\overline{\theta}^{k} - \overline{\theta}^{k-1},\ldots,
	\overline{\theta}^{k} - \overline{\theta}^{k-1}
	\end{pmatrix},
	\end{align*}
	
	so their squared norms are
	\begin{align*}
	\norm{r_{pri}^{k}}^{2}  = \sum_{t \in T}\norm{\theta^{t,k} - \overline{\theta}^{k}}^{2},\;\; \norm{r_{dual}^{k}}^{2} = |T|\rho^{2}\norm{\overline{\theta}^{k} - \overline{\theta}^{k-1}}^{2}.
	\end{align*}
	$ \norm{r_{pri}^{k}}^{2} $ is $ |T| $ times the variance of $ \{\theta^{t,k}\}_{t \in T} $, which can be interpreted as a natural measure of (lack of) consensus. Similarly, $ \norm{r_{dual}^{k}}^{2} $ is a measure of the step length. These suggest that a reasonable stopping criterion is that the primal and dual residuals must be small.
	
	The resulting ADMM algorithm in scaled form is formally presented in the following.
	\begin{algorithm}[ht]
		\caption{ADMM for \ref{saa-general-inverse-model}}\label{alg:admm-imop}
		\begin{algorithmic}[1]
			\Require Noisy decisions $ \{\mfy_{i}\}_{i \in [N]} $, weight samples $ \{w_{k}\}_{k \in [K]} $.
			\State Set $ k = 0 $ and initialize $ \theta^{0} $ and $ \mfv^{t,0} $ for each $ t \in T $.
			\While {stopping criterion is not satisfied}
			\For{$t \in [T]$}
			\State $ \theta^{t,k+1} \leftarrow \argmin_{\theta^{t}}\left\{
			\sum_{i\in N_{t}}l_{K}(\mfy_{i},\theta^{t}) + (\rho/2)\norm{\theta^{t} - \theta^{k} + \mfv^{t,k}}^{2} \right\} $.
			\EndFor
			\State $ \theta^{k+1} \leftarrow \frac{1}{|T|}\sum\limits_{t \in T}\bigg(\theta^{t,k+1} + \mfv^{t,k}\bigg) $.
			\For{$t \in [T]$}
			\State $ \mfv^{t,k+1} \leftarrow \mfv^{t,k} + \theta^{t,k+1} - \theta^{k+1} $.
			\EndFor
			\State $ k \leftarrow k + 1 $.
			\EndWhile
		\end{algorithmic}
	\end{algorithm}
	\begin{remark}
		$ (i) $ With a slight abuse of notation, we use $ \theta^{k} $ to denote the estimation of $ \theta $ in the $ k $-th iteration, and $ \theta^{t} $ to denote the local variable for the observations in $ t $-th group. $ (ii) $ The stopping criterion could be that $ \norm{r_{pri}^{k}} < \epsilon^{pri} $ and $ \norm{r_{dual}^{k}} < \epsilon^{dual} $, or the maximum iteration number is reached. $ (iii) $ Note that $ L_{\rho}(\theta,\{\theta^{t}\}_{t\in[T]},\{\mfv^{t}\}_{t\in[T]}) $ is separable in $ \theta^{t} $. Hence, the $ \theta^{t} $-update step splits into $ |T| $ independent problems that can be implemented in parallel. We show in experiments parallel computing would drastically improve the computational efficiency. For the same reason, the dual variables $ \mfv^{t} $-update step can be carried out in parallel for each $ t \in [T] $.
	\end{remark}
	\begin{remark}
		For the initialization of $ \theta^{0} $ in Algorithm \ref{alg:admm-imop}, we can incorporate the idea in  \citet{keshavarz2011imputing} that imputes a convex objective function by minimizing the residuals of KKT conditions incurred by noisy data. This leads to the following initialization problem:
		\begin{align}
		\label{dis-qp}
		\begin{array}{llll}
		\min\limits_{\theta\in\Theta} &  \sum\limits_{i \in [N]}\big(r_{comp}^{i} + r_{stat}^{i}\big) \vspace{1mm}\\
		\text{s.t.} & \mfu_{i} \geq \zero_{m}, \vspace{1mm}\\
		& \lvert\mfu_{i}^{T}\mathbf{g}(\mfy_{i},\theta) \rvert \leq r_{comp}^{i}, & \forall i \in [N], \vspace{1mm}\\
		& \bigvee\limits_{k \in [K]}\left[\begin{array}{l}
		\norm[2]{\nabla w_{k}^{T}\mathbf{f}(\mfy_{i},\theta) + \mfu_{i}^{T}\nabla\mathbf{g}(\mfy_{i},\theta)} \leq r_{stat}^{i} \vspace{1mm}
		\end{array}\right], & \forall i \in [N], \\
		& \mfu_{i} \in \bR^{m}_{+},  \;\;  r_{comp}^{i} \in \bR_{+}, \;\; r_{stat}^{i} \in \bR_{+}, & \forall i \in [N],
		\end{array}
		\end{align}
		where $ r_{comp2}^{i} $ and $ r_{stat}^{i}  $ are residuals corresponding to the complementary slackness and stationarity in KKT conditions for the $ i $-th noisy decision $ \mfy_{i} $. The disjunction constraints are imposed to assign one of the weight samples to $ \mfy_{i} $. Similarly, we can integrate the approach of minimizing the slackness needed to render observations to (approximately) satisfy variational inequalities \citep{bertsimas2015data} into our model, to provide an initialization of $ \theta^{0} $.
	\end{remark}

	\subsection{Solving IMOP through a Clustering-based Approach}\label{section:clustering}
	We provide in this section deep insights on the connections between \ref{saa-general-inverse-model} and the K-means clustering problem. Leveraging these insights, we develop an efficient clustering-based algorithm to solve \ref{saa-general-inverse-model}.
	
	K-means clustering aims to partition the observations into $ K $ clusters (or groups) such that the average squared distance between each observation and its closest cluster centroid is minimized. Given observations $\{\mfy_{i}\}_{i \in [N]}$, a mathematical formulation of K-means clustering is presented in the following \citep{bagirov2008modified,aloise2009branch}.
	\begin{align*}
	\label{formulation:clustering}
	\tag*{K-means clustering}
	\begin{array}{llll}
	\min\limits_{\mfx_{k},z_{ik}} & \frac{1}{N}\sum\limits_{i \in [N]} \norm{\mfy_{i} - \sum\limits_{k \in [K]}z_{ik}\mfx_{k}}^{2}\vspace{1mm}\\
	\;\text{s.t.} & \sum\limits_{k \in [K]}z_{ik}  = 1, & \forall i \in [N],  \vspace{1mm}  \\
	& \mfx_{k} \in \bR^{n},\;\;z_{ik} \in \{0,1\}, & \forall i \in [N],\; k \in [K],
	\end{array}
	\end{align*}
	where $ K $ is the number of clusters, and $\{\mfx_{k}\}_{k \in [K]}$ are the centroids of the clusters.
	
	Clearly, in both \ref{saa-general-inverse-model} and \ref{formulation:clustering}, one needs to assign $\{\mfy_{i}\}_{i \in [N]}$ to certain clusters in such a way that the average squared distance between $ \mfy_{i} $ and its closest $ \mfx_{k} $ is minimized. The difference is whether $\mfx_{k}$ has restriction or not. In \ref{saa-general-inverse-model}, each $\mfx_{k}$ is restricted to belong to $S(w_{k},\theta)$, while there is no restriction for $\mfx_{k}$ in \ref{formulation:clustering}. As such, each $\mfx_{k}$ in \ref{formulation:clustering} is the centroid of the observations in the $k$th cluster. Nevertheless, we will show in the following that the centroid of cluster  $ k $ is closely related to $\mfx_{k}$ in \ref{saa-general-inverse-model} for each $ k \in [K] $. More precisely, we are able to obtain $\mfx_{k}$ given only the centroid and the number of observations in each cluster.
	
	For each $k \in [K]$, we denote $ C_{k} $ the set of noisy decisions with $ z_{ik} = 1 $ after solving \ref{saa-general-inverse-model} to optimal. That is, observations in $ C_{k} $ are closest to $ \mfx_{k} $. Consequently, we partition $\{\mfy_{i}\}_{i \in [N]}$ into $K$ clusters $\{C_{k}\}_{k \in[K]}$. Let $\overline{\mfy}_{k} = \frac{1}{|C_{k}|}\sum_{\mfy_{i} \in C_{k}}\mfy_{i}$ be the centroid of cluster $ C_{k} $, and denote $Var(C_{k})$ the variance of $ C_{k} $. Through an algebraic calculation, we get
	\begin{align}\label{clustering:transformation}
	M^{N}_{K}(\theta) = \frac{1}{N}\sum_{i \in [N]}\lVert \mfy_{i} - \sum_{k \in [K]}z_{ik}\mfx_{k}\rVert_{2}^{2} = \frac{1}{N}\sum_{k \in [K]}|C_{k}|\bigg(\lVert \overline{\mfy}_{k} - \mfx_{k}\rVert_{2}^{2} + Var(C_{k})\bigg).
	\end{align}
	
	Note that $\{Var(C_{k})\}_{k \in [K]}$ is a set of fixed values when clusters $\{C_{k}\}_{k \in[K]}$ are given. If we know the clusters $\{C_{k}\}_{k \in[K]}$ beforehand, we see in \eqref{clustering:transformation} that $ K $ centroids $\{\overline{\mfy}_{k}\}_{k \in [K]}$ and $ \{|C_{k}|\}_{k\in[K]} $ are enough to solve \ref{saa-general-inverse-model}. This is the key insight we leverage to solve \ref{saa-general-inverse-model}. However, similar to K-means clustering, $\{C_{k}\}_{k \in[K]}$ are not known a priori. In K-means clustering algorithm \citep{kmeans1982}, this problem is solved by initializing the clusters, and then iteratively updating the clusters and centroids until convergence. Similarly, we propose a procedure that alternately clusters the noisy decisions (assignment step) and find $ \theta $ and $ \{\mfx_{k}\}_{k\in[K]} $ (update step) until convergence. Given $ \theta $ and $ \{\mfx_{k}\}_{k\in[K]} $, the assignment step can be done easily as we discussed previously.  Moreover, the update step can be established by solving the problem as follows.
	\begin{align*}
	\label{update-imop-kmeans}
	\tag*{Kmeans-IMOP}
	\begin{array}{llll}
	\min\limits_{\theta,\mfx_{k'}} & \frac{1}{N}\sum\limits_{k \in [K]}|C_{k}|\lVert \overline{\mfy}_{k} - \sum_{k' \in [K]}z_{kk'}\mfx_{k'}\rVert_{2}^{2} \vspace{1mm}\\
	\;\text{s.t.} & \mfx_{k'} \in S(w_{k'},\theta), & \forall k' \in [K], \vspace{1mm}\\
	& \sum\limits_{k' \in [K]}z_{kk'}  = 1, & \forall k \in [K], \vspace{1mm} \\
	& z_{kk'} \in \{0,1\}, & \forall k \in [K],\; k' \in [K].
	\end{array}
	\end{align*}
	
	The algorithm is formally presented in the following.
	\begin{algorithm}[ht]
		\caption{Solving \ref{saa-general-inverse-model} through a Clustering-based Approach}\label{alg:clustering-imop}
		\begin{algorithmic}[1]
			\Require Noisy decisions $\{\mfy_{i}\}_{i \in [N]}$, weight samples $\{w_{k}\}_{k \in [K]}$.
			\State \textbf{Initialization}: Partition $\{\mfy_{i}\}_{i \in [N]}$ into $K$ clusters using K-means clustering. Calculate $\{\overline{\mfy}_{k}\}_{k \in [K]}$. Solve \ref{update-imop-kmeans} and get an initial estimation of $\theta$ and $\{\mfx_{k}\}_{k \in [K]}$.
			\While {stopping criterion is not satisfied}
			\State \textbf{Assignment step}: Assign each $\mfy_{i} $ to the closest $ \mfx_{k}$ to form new clusters. Calculate their centroids $\{\overline{\mfy}_{k}\}_{k \in [K]}$. 
			\State \textbf{Update step}: Update $\theta$ and $\{\mfx_{k}\}_{k \in [K]}$ by solving \ref{update-imop-kmeans}.
			\EndWhile
			\Ensure An estimate of the parameter of \ref{mop}. Denote it by $\hat{\theta}_{C}$.
		\end{algorithmic}
	\end{algorithm}
	\begin{remark}
		$(i)$ In practice, we would apply one of the following as the stopping criterion: cluster assignments do not change; or, the maximum number of iterations is reached. $(ii)$ In \textbf{Initialization} step, we take K-means++ algorithm \citep{arthur2007k} as the default clustering method, run it multiple times and select the centroids of the best clustering results to further solve \ref{update-imop-kmeans}. $(iii)$ In the \textbf{Assignment step}, note that we only handle   non-empty clusters and break ties consistently, e.g., by assigning an observation $ \mfy_{i} $ to the cluster with the lowest index if there are several equidistant $ \mfx_{k} $. Otherwise, the algorithm can cycle forever in a loop of clusters that have the same cost. $(iv)$ In the \textbf{Update step}, \ref{update-imop-kmeans} can be solved either by directly computing the KKT based single level reformulation or by applying the ADMM approach to \ref{saa-general-inverse-model}.
	\end{remark}

	Since \ref{saa-general-inverse-model} is non-convex, there may exist multiple local optimal solutions. Nevertheless, we will establish that Algorithm \ref{alg:clustering-imop} indeed converges to a (local) optimal solution in finite steps. The key step of the proof is the following lemma.
	
	\begin{lemma}
		Both the \textbf{Assignment step} and the \textbf{Update step} in Algorithm \ref{alg:clustering-imop} decrease $M^{N}_{K}(\theta)$.
	\end{lemma}
	\proof{Proof. }
	First, $M^{N}_{K}(\theta)$ decreases in the \textbf{Assignment step} since each $\mfy_{i}$ is assigned to the closest $\mfx_{k}$. So the distance $\mfy_{i}$ contributes to $M^{N}_{K}(\theta)$ decreases. Second, $\overline{M}^{N}_{K}(\theta)$ decreases in the \textbf{Update step} because the new $ \theta $ and $ \{\mfx_{k}\}_{k \in [K]} $ are the ones for which $M^{N}_{K}(\theta)$ attains its minimum.
	\Halmos\endproof
	
	\begin{theorem}
		Suppose there is an oracle to solve \ref{update-imop-kmeans}. Algorithm \ref{alg:clustering-imop} converges to a (local) optimal solution of \ref{saa-general-inverse-model} in a finite number of iterations.
	\end{theorem}
	\proof{Proof. }
	Since there is at most $ K^N $ ways to partition $\{\mfy_{i}\}_{i \in [N]}$ into $K$ clusters, the monotonically decreasing Algorithm \ref{alg:clustering-imop} will eventually arrive at a (local) optimal solution in finite steps.
	\Halmos\endproof
	
	\begin{remark}
		$(i)$ In practice, Algorithm \ref{alg:clustering-imop} converges pretty fast, typically within several iterations. The main reason is that the \textbf{Initialization} step often provides a good estimation of the true parameter, since the $ K $ centroids returned by K-means clustering represent the observations well in general, especially when $K$ is large. $(ii)$ Algorithm \ref{alg:clustering-imop} is extremely efficient in computation especially when $N \gg K$. The reason is that in each iteration only $K$ representative points (i.e., the centroids of clusters) are used to update $\theta$, instead of the whole batch of observations.
	\end{remark}

	\section{Computational Experiments} \label{experiments}
	
	In this section, we illustrate the performances of the proposed algorithms on a multiobjective linear program (MLP), two multiobjective quadratic programs (MQP) and a general multiobjective nonlinear program. Our experiments have been run on Bridges system at the Pittsburgh Supercomputing Center (PSC) \citep{6866038,Nystrom}.  The mixed integer second order conic problems (MISOCP) are solved with Gurobi. The mixed-integer noncovex programming problmes are solved with FilMINT \citep{abhishek2010filmint}. All the algorithms are programmed with Julia \citep{bezanson2017julia} unless otherwise specified. All the single level reformulations of the IMOP are given in Appendix. Throughout this section we use SRe to refer that we solve the single level reformulation directly without using the ADMM or Clustering-based approach.
	
	\subsection{Learning the Objective Functions of an MLP}
	Consider the following Tri-objective linear programming problem
	\begin{align*}
	\begin{array}{llll}
	\min & \{ \begin{matrix} -x_{1}, -x_{2}, -x_{3}\end{matrix} \} \vspace{1mm} \\
	\;s.t.   & x_{1} + x_{2} + x_{3} \leq  5, \vspace{1mm} \\
	& x_{1} + x_{2} + 3x_{3} \leq 9, \vspace{1mm} \\
	& x_{1},x_{2}, x_{3}  \geq 0.
	\end{array}
	\end{align*}
	
\begin{figure}[ht]
		\centering
		\includegraphics[width=0.5\linewidth]{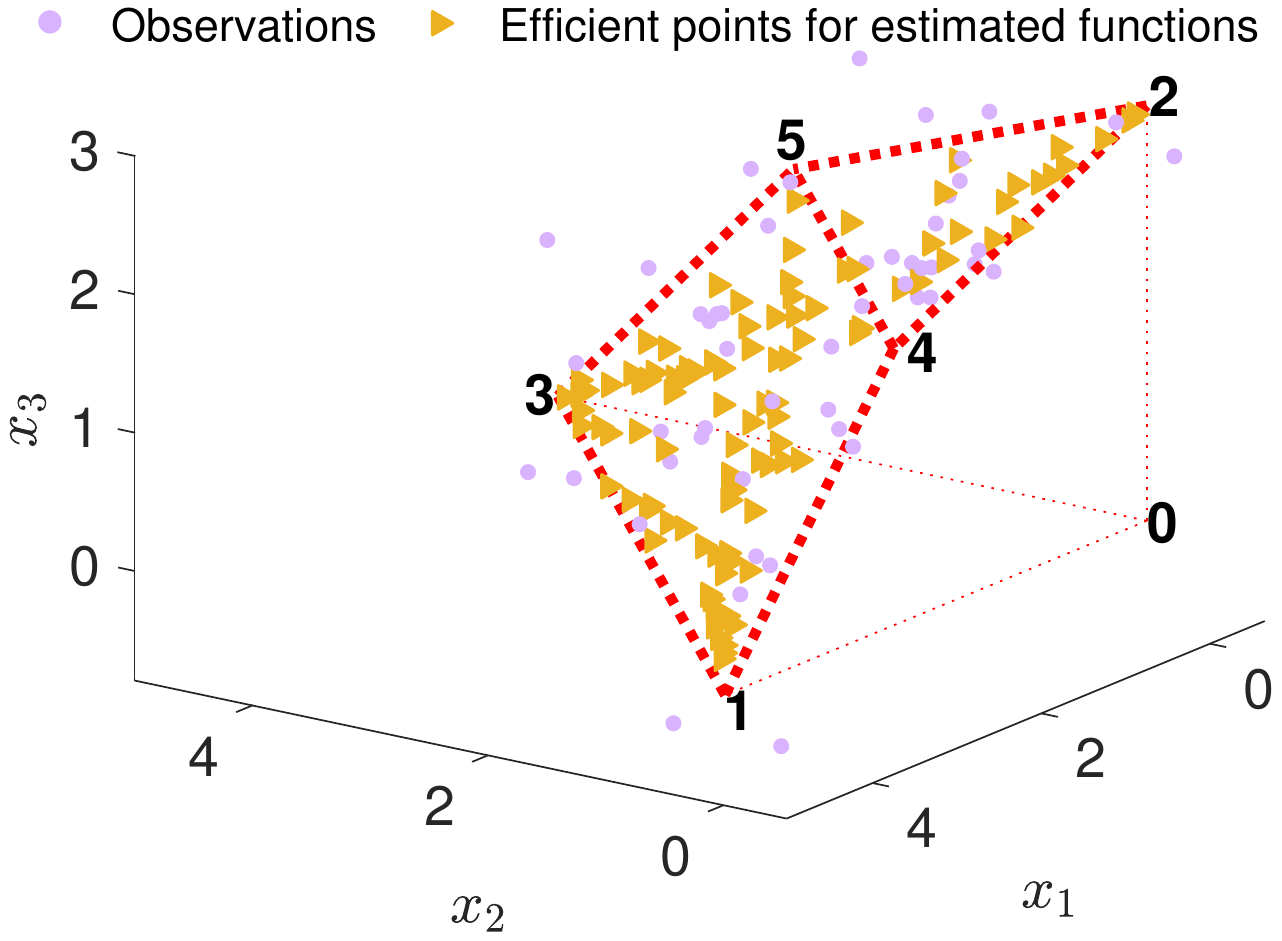}
		\caption{Learning the objective functions of an MLP. The Arabic numbers indicate the six vertices of the feasible region. Efficient edges are represented by bold dashed lines. Light blue dots indicate the $ 50 $ observations that are randomly generated. Orange triangles indicate the 100 efficient points generated by Genetic algorithm using the estimated functions.}
		\label{fig:c_estimated}
	\end{figure}

	In this example, there are two efficient faces, one is the triangle defined by vertices $ (2,4,5) $, the other one is the tetragon defined by vertices $ (1,3,5,4) $ as shown by Figure \ref{fig:c_estimated}.
	
	We seek to learn the objective functions, i.e., $\{ \mfc_{1},\mfc_{2}, \mfc_{3}\} $, given efficient solutions corrupted by noises. We generate the data as follows. First, $ N $ efficient points $ \{\mfx_{i}\}_{i \in [N]} $ are uniformly sampled on efficient faces $ (2,4,5) $ and $ (1,3,5,4) $. Next, the noisy decision $ \mfy_{i} $ is obtained by adding noise to $ \mfx_{i} $ for each $ i \in [N] $. More precisely, $ \mfy_{i} = \mfx_{i} + \epsilon_{i} $, where each element of $ \epsilon_{i} $ has a normal distribution with mean $ 0 $ and standard deviation $ 0.5 $ for all $ i \in [N] $. We assume that the parameters to be learned are negative. In addition, we add the normalization constraints $ \one^{T}\mfc_{1} = -1, \one^{T}\mfc_{2} = -1 $ and $ \one^{T}\mfc_{3} = -1 $ to prevent the arise of trivial solutions, such as $ \mfc_{1} = \mfc_{2} = \mfc_{3} = [0,0,0]^{T} $. Then, we uniformly choose the weights $ \{w_{k}\}_{k \in [K]} $ such that $ w_{k} \in \mathscr{W}_{3} $ for each $ k \in [K] $.
	
	We use the SRe approach to solve for the objective functions. MIP gap and time limit for the solver are set to be $ 10^{-3} $ and $ 100 $s, respectively. Figure \ref{fig:c_estimated} shows the randomly generated observations and the estimated efficent surfaces for $ N = 20 $ and $ K = 50 $. The estimating results are $ \hat{\mfc}_{1} = [-0.2280, -0.1594, -0.6126]^{T} $, $ \hat{\mfc}_{2} = [-0.5, -0.5, 0.0]^{T} $ and  $ \hat{\mfc}_{3} = [0.0, -0.3264, -0.6736]^{T} $. Then we generate the efficient set for the estimated parameters using the Genetic algorithm (GA). As shown in Figure \ref{fig:c_estimated}, $ 100 $ efficient points generated by GA spread on the faces $ (2,4,5) $ and $ (1,3,5,4) $, indicating that they are the efficient faces for the estimated parameters, which coincide with the efficient faces for the true parameters. Hence, we successfully learn the objective functions that reconcile the true efficient set.
	
	Given the estimation $ \hat{\mfc}_{1} = [-0.3333, -0.3333, -0.3333]^{T} $, $ \hat{\mfc}_{2} = [-0.3450, -0.3450, -0.3099]^{T} $ and  $ \hat{\mfc}_{3} = [-0.1227, -0.1227, -0.7546]^{T} $, we apply Algorithm \ref{alg:test-process} to test whether this example is identifiable or not. Step 1 is omitted since it has been completed in the previous experiment. In Step 2, we randomly sample $ |N'| = 200 $ points from the efficient set. In Step 3, we uniformly generate $ |K'| = 200 $ weights. In Step 4, we replace the optimal set $ S(w_{k}, \theta) $ by KKT conditions and solve the \ref{test-problem} , and it achieves the maximum value when $ \mfc_{1} =  [-0.0222, -0.0222, -0.9556]^{T}, \mfc_{2} =  [0.0, -1.0, 0.0]^{T}, \mfc_{3} =  [-1, 0, 0]^{T}$. The test statistic $ z_{test} = 4.3089 $, which is greater than $ 0 $. Thus, we claim that this example is non-identifiable.
	
	Recall that we propose three ways to tackle the non-identifiability issue in Section \ref{sec:dispose non-uniqueness}. Obviously, the first method of collecting more data fails because multiple parameters lead to the same efficient set in this example. For the third method, we restrict the parameter space by fixing $ \mfc_{1} = [-1, 0, 0]^{T} $ and $ \mfc_{2} = [0, -1, 0]^{T} $. Then, we use the SRe approach to solve for $ \mfc_{3} $ in the same setting as before, and we get $ \hat{\mfc}_{3} = [-0.0758, -0.0758, -0.8484]^{T} $. Thus, the third method can not eliminate the non-identifiability issue either as there are multiple parameters that could explain the data. Lastly, we combine the first, second and third methods. Namely, we use $ 200 $ noisy decisions randomly generated as previous, set $ \mfc_{1} = [-1, 0, 0]^{T} $, $ \mfc_{2} = [0, -1, 0]^{T} $, and assume that the weight for each of the randomly generated observations is known. Then, we use the SRe approach to solve for $ \mfc_{3} $ in the same setting as before. We successfully recover the true parameter this time.

	\subsection{Learning the Preferences and Constraints of an MQP}
	
	We consider the following multiobjective quadratic programming problem.
	\begin{align*}
	\begin{array}{llll}
	\min\limits_{\mfx \in \bR_{+}^{2}} & \left( \begin{matrix} f_{1}(\mfx) = \frac{1}{2}\mfx^{T}Q_{1}\mfx + \mfc_{1}^{T}\mfx  \\ f_{2}(\mfx) = \frac{1}{2}\mfx^{T}Q_{2}\mfx + \mfc_{2}^{T}\mfx\end{matrix} \right) \vspace{1mm} \\
	\;s.t.   & A\mfx \geq \mfb,
	\end{array}
	\end{align*}
	where parameters of the objective functions and the constraints are
	\begin{align*}
	Q_{1} = \begin{bmatrix}
	1 & 0 \\
	0 & 2
	\end{bmatrix}, \mfc_{1} = \begin{bmatrix}
	3 \\
	1
	\end{bmatrix},
	Q_{2} = \begin{bmatrix}
	2 & 0 \\
	0 & 1
	\end{bmatrix}, \mfc_{2} = \begin{bmatrix}
	-6 \\
	-5
	\end{bmatrix},
	A = \begin{bmatrix}
	-3 & 1 \\
	0 & -1
	\end{bmatrix}, \mfb = \begin{bmatrix}
	-6 \\
	-3
	\end{bmatrix}.
	\end{align*}
	

	\subsubsection{Learning the Right-hand Side of Constraints} In the first set of experiments, suppose the right-hand side $\mfb$ is unknown, and the learner seeks to learn $\mfb$ given the noisy decisions she observes. Assume that $\mfb$ is within the range $[-8,-1]^{2}$. We generate the data as follows. We first compute efficient solutions $ \{\mfx_{i}\}_{i \in [N]} $ by solving \ref{weighting problem} with weight samples $ \{w_{i}\}_{i \in [N]} $ that are uniformly chosen from $ \mathscr{W}_{2} $. Next, the noisy decision $ \mfy_{i} $ is obtained by adding noise to $ \mfx_{i} $ for each $ i \in [N] $. More precisely, $ \mfy_{i} = \mfx_{i} + \epsilon_{i} $, where each element of $ \epsilon_{i} $ has a truncated normal distribution supported on $ [-1,1] $ with mean $ 0 $ and standard deviation $ 0.1 $ for all $ i \in [N] $.
	
	Both the SRe approach and the ADMM approach (Algorithm \ref{alg:admm-imop}) are applied to solve for $ \mfb $  with different $ N $ and $ K $. The basic parameters for the implementation of the ADMM approach are given in the following. The observations are equally partitioned into $ T = N/2 $ groups. We pick the penalty parameter $ \rho = 0.5 $ as the best out of a few trials. We use the initialization $ \mfb^{0} = \mfv^{t,0} = \zero_{2} $ for the iterations. The tolerances of the primal and dual residuals are set to be $ \epsilon^{pri} = \epsilon^{dual} = 10^{-3} $. We find that Algorithm \ref{alg:admm-imop} converges in $ 100 $ iterations in general, thus the termination criterion is set to be either the norms of the primal and dual residuals are smaller than $ 10^{-3} $ or the iteration number $ k $ reaches $ 100 $.
	
\begin{figure}[ht]
		\centering
		\subfloat[]{%
			\includegraphics[width=0.45\linewidth]{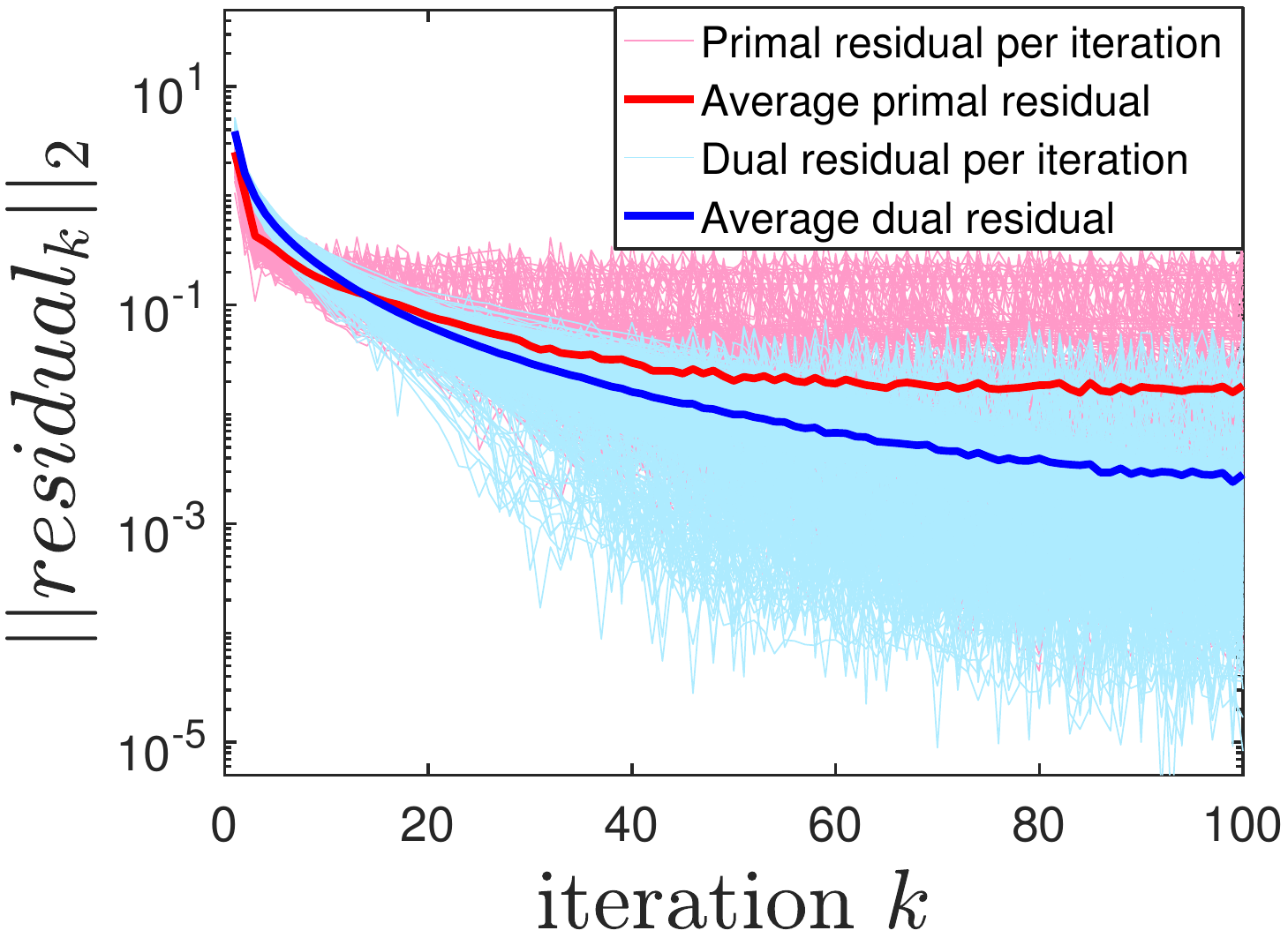}
			\label{fig:qp_b_residual}}\;\;\;
		\subfloat[]{%
			\includegraphics[width=0.45\linewidth]{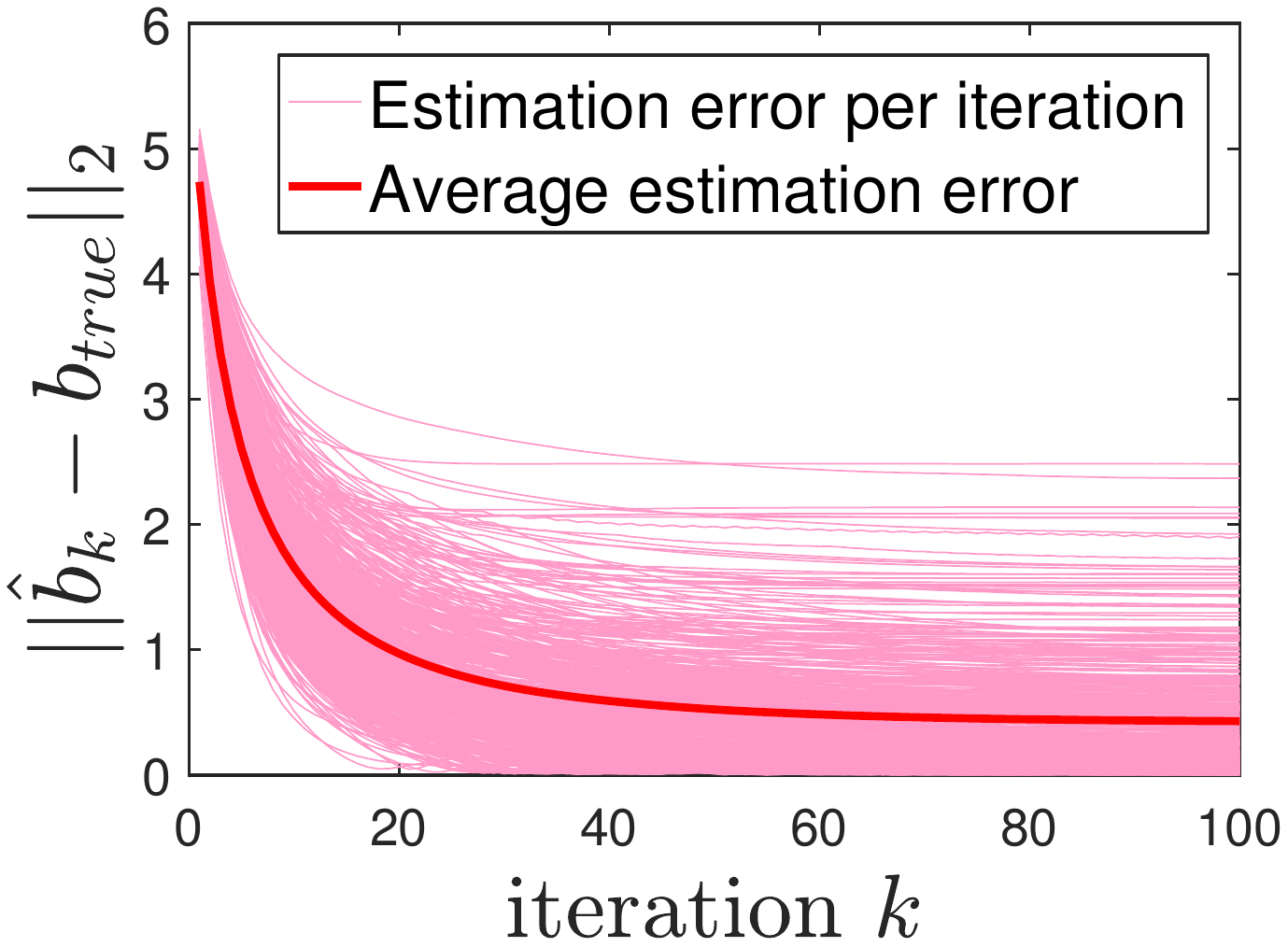}
			\label{fig:qp_b_estimation_error}}
		\caption{Learning the Right-hand Side of an MQP. We run $ 500 $ repetitions of the experiments with $ N = 20,$ $K = 21 $. (a) Norms of primal residuals and dual residuals versus iteration number. (b) Norms of estimation error versus iteration number. }
		\label{fig:qp_b}
	\end{figure}

	In Table \ref{table:qp-rhs} we summarize the computational results averaged over $ 10 $ repetitions of the experiments for each $ N $ and $ K $ using Algorithm \ref{alg:admm-imop}. Note that $ \mfb_{true} = [-3, -6]^{T} $. The smaller the estimator error is, the closer is $ \hat{\mfb} $ to $ \mfb_{true} $. Note that \ref{saa-general-inverse-model} is prediction consistent by Theorem \ref{theorem: M(theta_J^N) - M} for this example. The results in Table \ref{table:qp-rhs} show the estimation consistency of the \ref{saa-general-inverse-model} as the estimation error decreases to zero with the increase of the data size $ N $ and weight sample size $ K $, although it does not satisfy the conditions for Theorem \ref{theorem:estimation consistency}. Note that estimation consistency implies risk consistency. Thus, this result illustrates Theorem \ref{theorem: M(theta_J^N) - M}. Also, we see that the estimation error becomes more stable when using more weight samples, i.e., $ K $ becomes larger. In Tables \ref{table:qp-rhs-time2}- \ref{table:qp-rhs-time3}, we summarize the computational time that averages over $ 10 $ repetitions of the experiments for each algorithm, $ N $ and $ K $. Here p-ADMM means that we implement the $ \theta^{t} $-update step of ADMM in parallel with $ 28 $ cores. $ * $ means that we can not get reasonable estimation of the parameter within three hours. As shown in these tables, both ADMM and p-ADMM approaches drastically improve the computational efficacy over the SRe approach when $ N $ and $ K $ are large. On average, p-ADMM is two times faster than ADMM. Moreover, the SRe approach could handle only small size problems with roughly $ N \leq 20 $ and $ K \leq 11 $. To further illustrate the performance of the ADMM algorithm, we plot the primal and dual residuals versus the iteration number in each of the $ 100 $ repetitions for $ N = 20, K = 21 $, and the estimation error versus the iteration number in Figures \ref{fig:qp_b_residual} and \ref{fig:qp_b_estimation_error}, respectively. The two figures show that the ADMM approach converges within $ 100 $ iterations under the above setting.
	\begin{table}[ht]
		\centering
		\caption{Estimation Error $ \norm[2]{\hat{\mfb} - \mfb_{true}} $ for Different $ N $ and $ K $}
		\label{table:qp-rhs}
		\begin{tabular}{ccccccc}
			\hline
			& $N=5$  & $N=10$   & $N=20$   & $N=50$   & $N=100$  & $N=150$  \\ \hline
			$K=6$  & 1.496 & 1.063  & 0.861  & 0.601 & 0.531 & 0.506 \\
			$K=11$ & 1.410 & 0.956& 0.524  & 0.378 & 0.217 & 0.199 \\
			$K=21$ & 1.382  & 0.925 & 0.498 & 0.313 & 0.138 & 0.117 \\
			$K=41$ & 1.380   & 0.924 & 0.484 & 0.295 & 0.127 & 0.111 \\ \hline
		\end{tabular}
	\end{table}
	\begin{table}[ht]
		\centering
		\caption{Average Running Time over $ 10 $ Repetitions for Each of the Three Approaches (In Seconds)}
		\label{table:qp-rhs-time2}
		\begin{tabular}{cccccccccc}
			\hline
			& \multicolumn{3}{c}{$N=5$}   & \multicolumn{3}{c}{$N=10$}    & \multicolumn{3}{c}{$N=20$}    \\
			& SRe   & ADMM    & p-ADMM  & SRe   & ADMM     & p-ADMM  & SRe   & ADMM     & p-ADMM  \\ \hline
			$K=6$  & 0.31  & 14.92 & 11.72 & 0.78  & 23.13  & 15.10 & 4.07 & 43.95  & 20.73  \\
			$K=11$ & 0.42  & 20.93 & 12.83 & 3.10 & 33.88  & 19.43 & 705.36 & 66.91  & 28.95 \\
			$K=21$ & 3.83  & 33.23 & 17.74 & 391.18 & 61.99  & 36.79 & *       & 122.98 & 55.93 \\
			$K=41$ & 38.42 & 59.67 & 31.69 & *       & 156.78 & 107.48 & *       & 343.72 & 205.98 \\ \hline
		\end{tabular}
	\end{table}
	\begin{table}[ht]
		\centering
		\caption{Average Running Time over $ 10 $ Repetitions for Each of the Three Approaches (In Seconds)}
		\label{table:qp-rhs-time3}
		\begin{tabular}{cccccccccc}
			\hline
			& \multicolumn{3}{c}{$N=50$}   & \multicolumn{3}{c}{$N=100$}  & \multicolumn{3}{c}{$N=150$} \\
			& SRe   & ADMM     & p-ADMM  & SRe   & ADMM     & p-ADMM  & SRe  & ADMM     & p-ADMM  \\ \hline
			$K=6$  & 119.58 & 110.42  & 44.69 & 5423.19 & 222.19  & 87.45 & *      & 335.90 & 131.73 \\
			$K=11$ & *       & 166.39  & 69.25 & *       & 336.82  & 138.80 & *      & 508.30 & 208.95 \\
			$K=21$ & *       & 306.91 & 141.22 & *       & 613.08 & 278.28 & *      & 923.58 & 418.27 \\
			$K=41$ & *       & 819.94 & 501.40 & *       & 1705.70 & 1058.20  & *      & 2536.29 & 1572.20  \\ \hline
		\end{tabular}
	\end{table}
	
	\subsubsection{Learning the Objective Functions}\label{section:qp_c} In the second set of experiments, suppose $ \mfc_{1}$ and  $\mfc_{2}$ are unknown, and the learner seeks to learn them given the noisy decisions. Assume that $\mfc_{1}$ and $\mfc_{2}$ are within range $[-10,10]^{2}$. We generate the data in a way similar to the first set of experiments. The only difference is that each element of the noise has a uniform distribution supporting on $ [-0.25,0.25] $ with mean $ 0 $ for all $ i \in [N] $.
	
	We would like to use Algorithm \ref{alg:clustering-imop} to solve large-scale \ref{saa-general-inverse-model}. We note that the SRe approach can not handle cases when $ N \geq 10 $ and $ K \geq 11 $ in the \textbf{Update step}. Hence, the ADMM approach (Algorithm \ref{alg:admm-imop}) is applied to solve \ref{update-imop-kmeans}. The stopping criterion for Algorithm \ref{alg:clustering-imop} is that the maximum iteration number reaches five. In the \textbf{Initialization step}, we run K-means++ algorithm  $ 50 $ times to find the best clustering results. When solving \ref{update-imop-kmeans} using ADMM, we partition the observations in such a way that each group has only one observation. We pick the penalty parameter $ \rho = 0.5 $ as the best out of a few trials. We use the initialization $\mfc_{1}^{0} = \mfc_{2}^{0} = \mfv_{1}^{t,0} = \mfv_{2}^{t,0} = \zero_{2} $ for the iterations. The tolerances of the primal and dual residuals are set to be $ \epsilon^{pri} = \epsilon^{dual} = 10^{-3} $. The termination criterion is that either the norms of the primal and dual residuals are smaller than $ 10^{-3} $ or the iteration number $ k $ reaches $ 50 $.
	
\begin{figure}[ht]
		\centering
		\subfloat[]{%
			\includegraphics[width=0.33\linewidth]{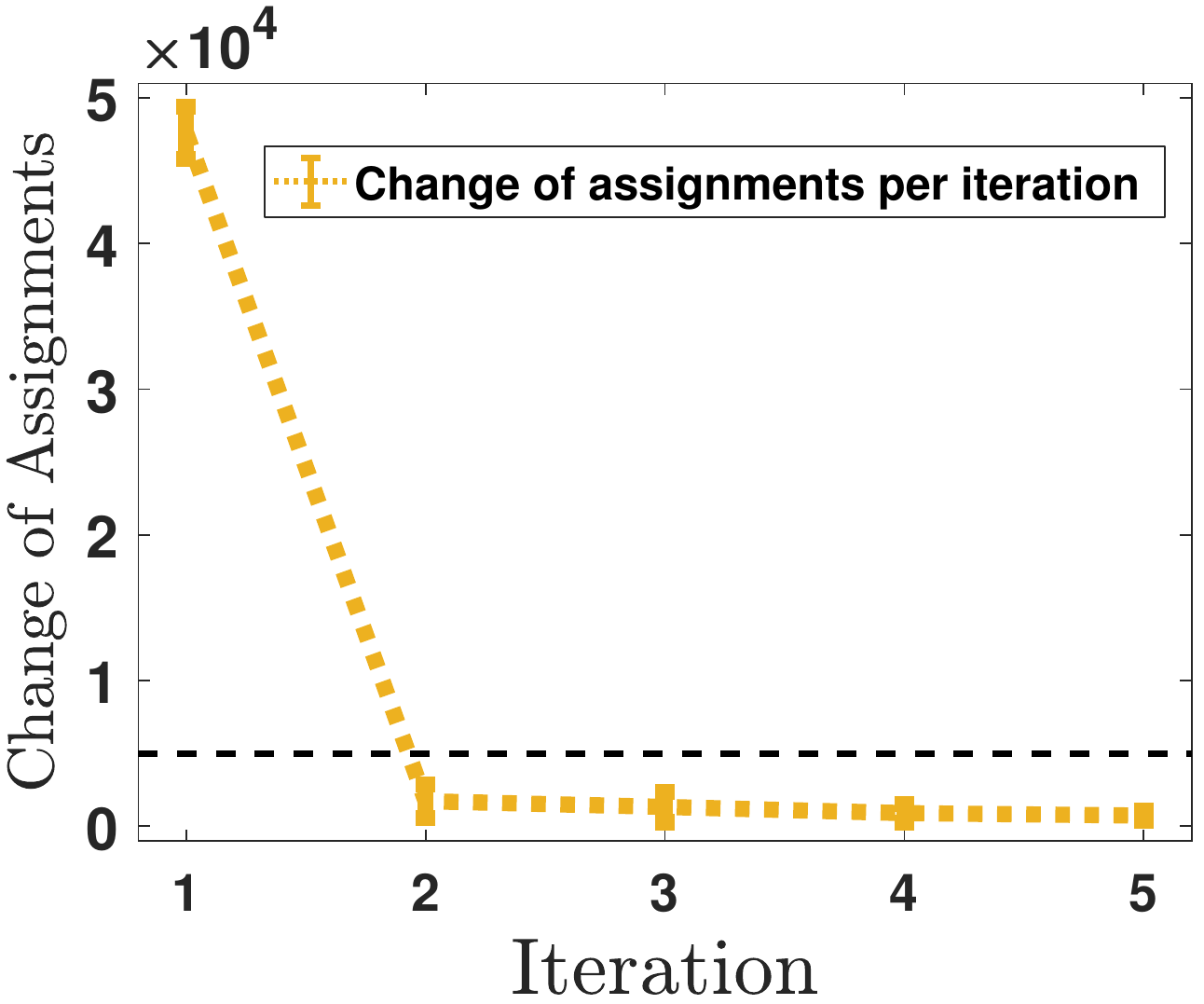}
			\label{fig:qp_c_assignment}}
		\subfloat[]{%
			\includegraphics[width=0.33\linewidth]{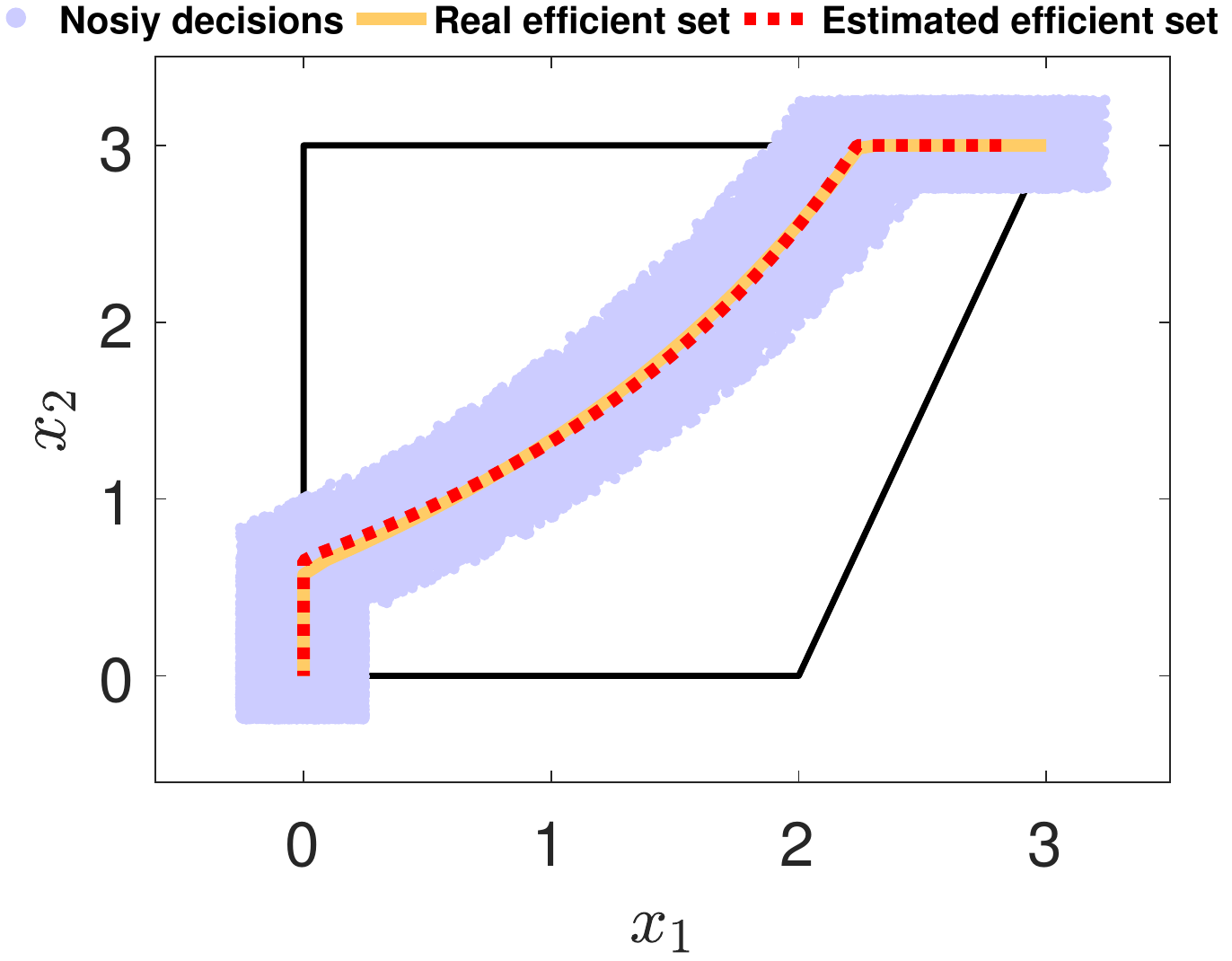}
			\label{fig:qp_c_estimate}}
		\subfloat[]{%
			\includegraphics[width=0.33\linewidth]{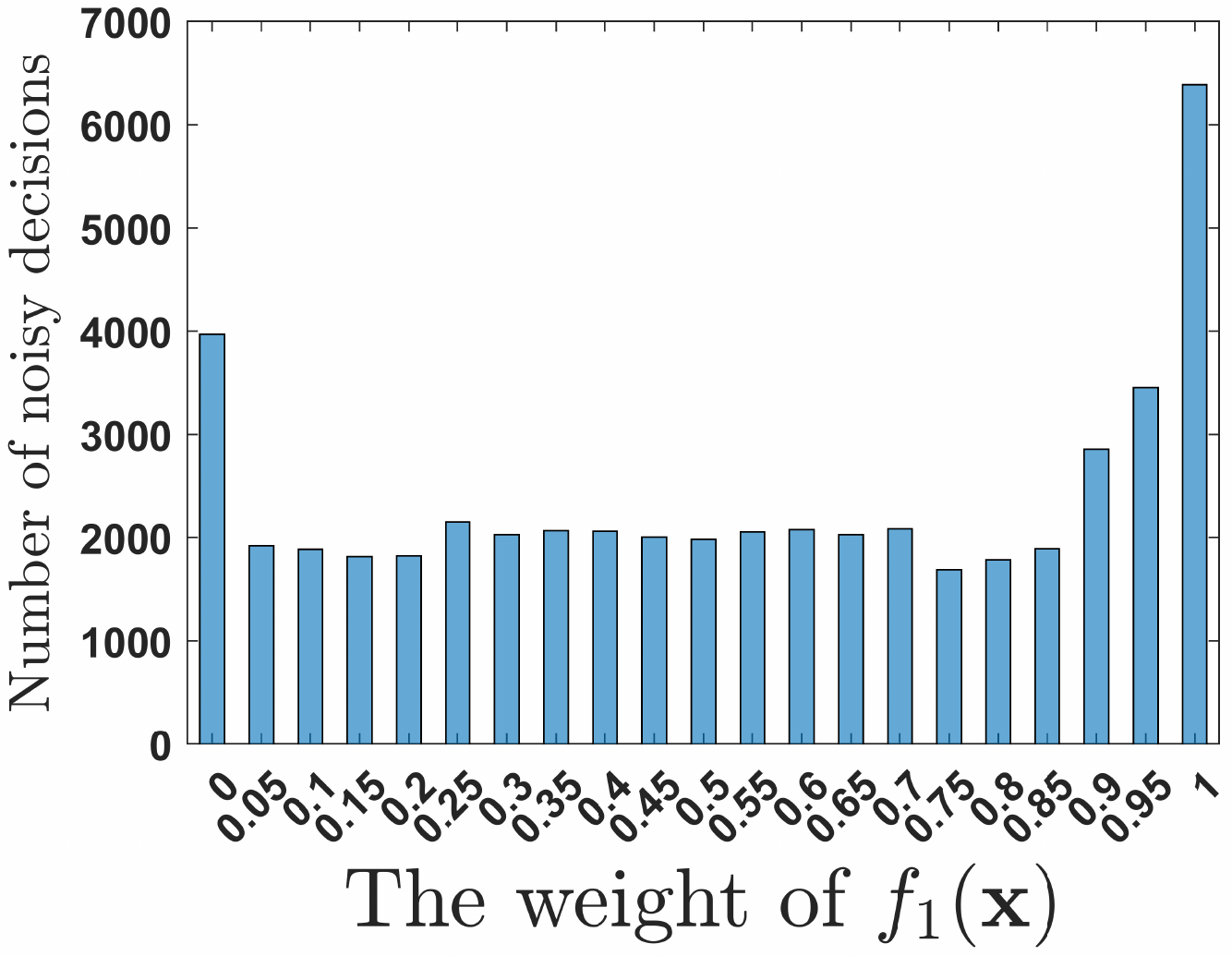}
			\label{fig:qp_c_weight}}
		\caption{Learning the objective functions of an MQP with $ N = 5\times 10^4 $ and $ K = 21 $. (a) The dotted brown line is the error bar plot of the change of the assignments in five iterations over 10 repetitions. (b) We pick the first repetition of the experiments. Blue dots indicate the noisy decisions. The estimated efficient set is indicated by green line. The real efficient set is shown by the red line. (c) Each bar represents the number of noisy decisions that have the corresponding weights for $ f_{1}(\mfx) $}
		\label{fig:qp_c}
	\end{figure}

	In Table \ref{table:qp-prediction}, we report the prediction errors averaged over $ 10 $ repetitions of the experiments for different $ N $ and $ K $. Here, we use an independent validation set that consists of $ 10^5 $ noisy decisions generated in the same way as the training data to compute the prediction error. We also calculate the prediction error using the true parameter and $ M(\theta_{true}) = 0.022742 $. More precisely, we evenly generate $ K = 10^4 $ weight samples and calculate the associated efficient solutions on the true efficient set. These efficient solutions are then used to find the prediction error of the true parameter. We observe that the prediction error has the trend to decrease to $ M(\theta_{true}) $ with the increase of the data size $ N $ and weight sample size $ K $. This makes lots of sense because \ref{saa-general-inverse-model} is prediction consistent by Theorem \ref{theorem: M(theta_J^N) - M} for this example. To further illustrate the performance of the algorithm, we plot the change of assignments versus iteration in the \textbf{Assignment step} over $ 10 $ repetitions of the experiments with $ N = 5 \times 10^4, K = 21 $ in Figure \ref{fig:qp_c_assignment}. One can see the assignments become stable in 5 iterations, indicating the fast convergence of our algorithm. Also, we plot the estimated efficient set with $ N = 5 \times 10^4, K = 21 $ in the first repetition in Figure \ref{fig:qp_c_estimate}. The estimated parameters are $ \hat{\mfc}_{1} = [1.83311,0.00047]^{T} $ and $\hat{\mfc}_{2} = [-5.63701,-4.72363]^{T} $. They are not equal to the true parameters as this MQP is non-identifiable. However, our method still recovers the unknown parameters quite well as the estimated efficient set almost coincides with the true one.
	
	We also plot our prediction of the distribution for the preferences of $f_{1}(\mfx)$ and $f_{2}(\mfx) $ among the $ 5 \times 10^4 $ noisy decisions. Since there are only two objective functions, it is sufficient to draw the distribution of the weight for $f_{1}(\mfx)$ (given that weights of $f_{1}(\mfx)$ and $f_{2}(\mfx)$ summing up to 1). As shown in Figure \ref{fig:qp_c_weight}, except in the two endpoint areas, the number of noisy decisions assigned to each weight follows roughly uniformly distribution, which matches our uniformly sampled weights. Indeed, comparing Figures \ref{fig:qp_c_estimate} and \ref{fig:qp_c_weight}, we would like to point out that a \textit{boundary effect} probably occurs in these two endpoint areas. Although different weights are imposed on component functions, the noiseless optimal solutions, as well as observed decisions, do likely to merge together due to the limited feasible space in those areas. We believe that it reflects an essential challenge in learning multiple objective functions in practice and definitely deserves a further study.
	
	\begin{table}[ht]
		\centering
		\caption{Prediction Error $ M(\hat{\theta}_{K}^{N}) $ for Different $ N $ and $ K $ }
		\label{table:qp-prediction}
		\begin{tabular}{@{}ccccccccc@{}}
			\toprule
			& $N=50$ & $N=100$ & $N=250$ & $N=500$ & $N=1000$ & $N=5000$ & $N=10000$ & $N=50000$ \\ \midrule
			$K=6$  & 0.050  & 0.043   & 0.039  & 0.040   & 0.039    & 0.040    & 0.038     & 0.038     \\
			$K=11$ & 0.030  & 0.028   & 0.028  & 0.027   & 0.027    & 0.026    & 0.026     & 0.025     \\
			$K=21$ & 0.026  & 0.025   & 0.024   & 0.024   & 0.024    & 0.024    & 0.024     & 0.024     \\
			$K=41$ & 0.025  & 0.024   & 0.024   & 0.023   & 0.023    & 0.023    & 0.023     & 0.023     \\ \bottomrule
		\end{tabular}
	\end{table}
	
	\subsection{Learning the Expected Returns in Portfolio Optimization}
	In this example, we consider various noisy decisions arising from different investors in a stock market. More precisely, we consider a portfolio selection problem, where investors need to determine the fraction of their wealth to invest in each security in order to maximize the total return and minimize the total risk. The portfolio selection process typically involves the cooperation between an investor and a portfolio analyst, where the analyst provides an efficient frontier on a certain set of securities to the investor and then the investor selects a portfolio according to her preference to the returns and risks. The classical Markovitz mean-variance portfolio selection \citep{markowitz1952portfolio} in the following is often used by analysts.
	\begin{align*}
	\begin{array}{llll}
	\min\limits_{\mfx} & \left( \begin{matrix}f_{1}(\mfx) &= -\mathbf{r}^{T}\mfx \\ f_{2}(\mfx) &= \mfx^{T}Q\mfx\end{matrix} \right) \vspace{1mm} \\
	\;s.t.   & 0 \leq x_{i} \leq b_{i}, &\forall i \in [n], \\
	& \sum\limits_{i=1}^{n}x_{i} = 1,
	\end{array}
	\end{align*}
	where $  \mathbf{r} \in \bR^{n}_{+} $  is a vector of individual security expected returns, $ Q \in \bR^{n \times n} $ is the covariance matrix of securities returns, $ \mfx $ is a portfolio specifying the proportions of capital to be invested in the different securities, and $ b_{i} $ is an upper bound put on the proportion of security $ i \in [n] $.
	
	In portfolio optimization, the forecast of security expected returns $ \mfr $ is essential within the portfolio selection process. Note that different analysts might use different $ \mfr $, which are due to different information sources and insights, to make recommendations. Consider a scenario that A observes that customers of B often make more revenues. Then, A might want to use our model to infer the $ \mfr $ that B really uses.

\begin{figure}[ht]
		\centering
		\subfloat[]{%
			\includegraphics[width=0.4\textwidth]{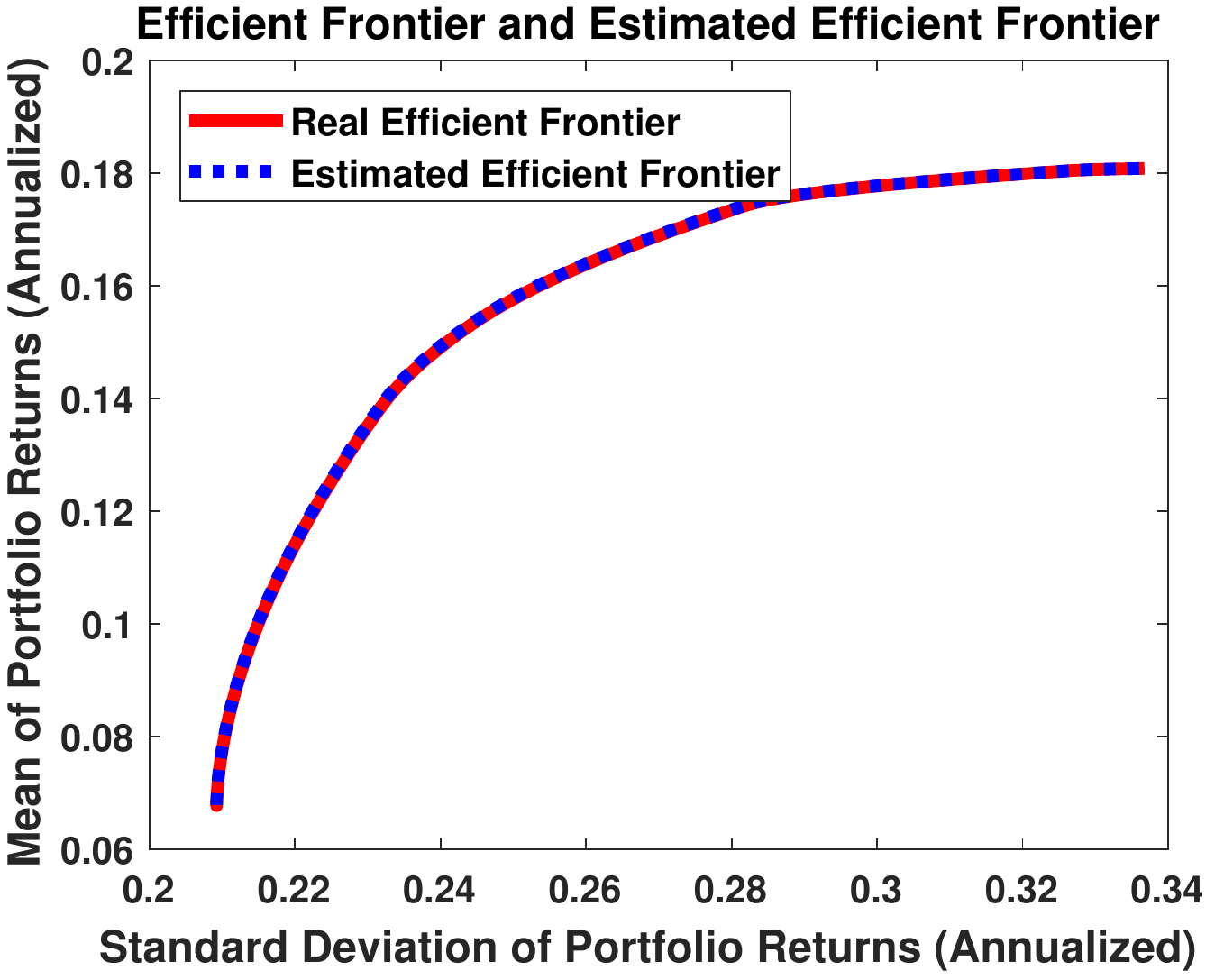}
			\label{fig:effcientfrontierport}}\;\;\;\;
		\subfloat[]{%
			\includegraphics[width=0.4\textwidth]{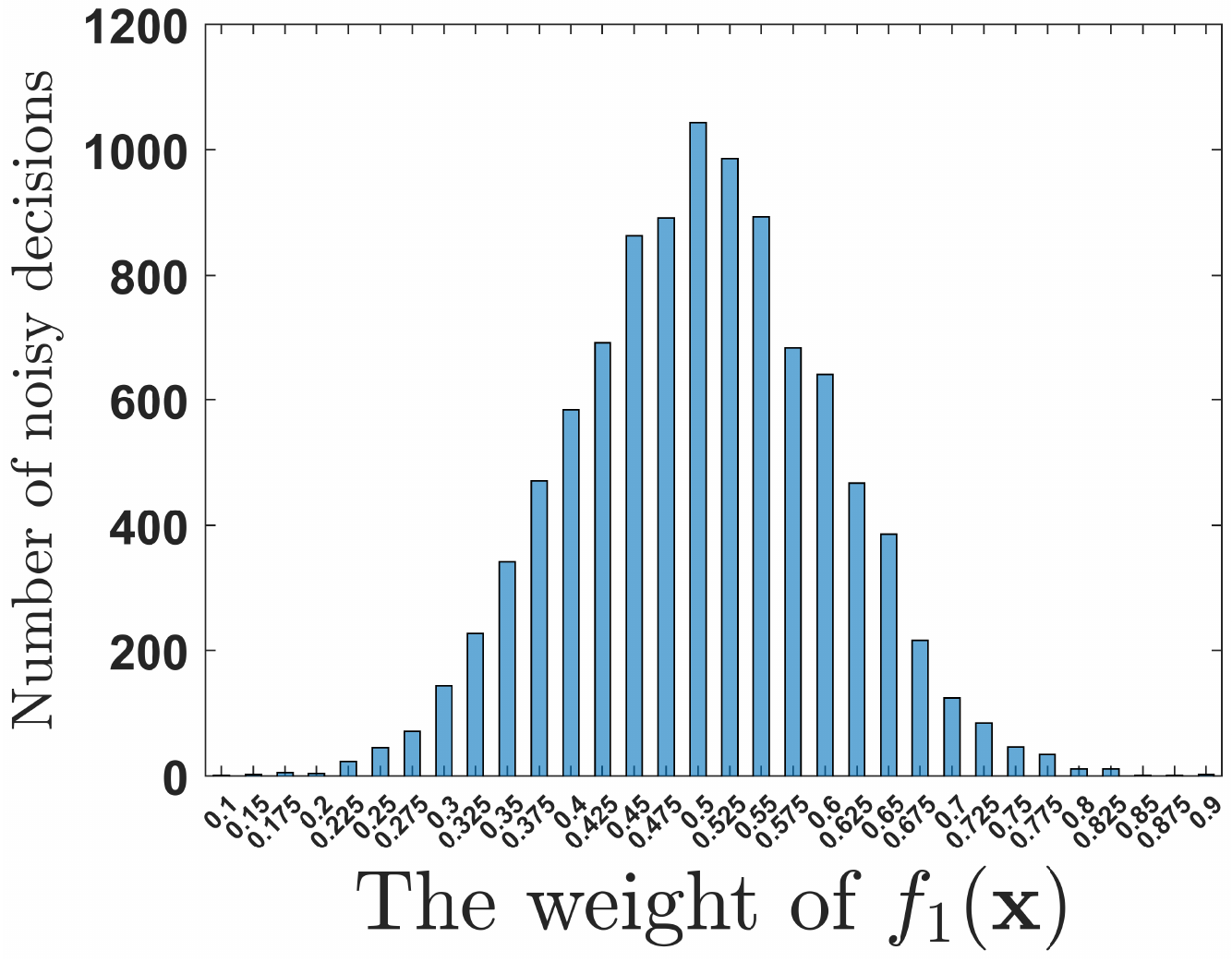}
			\label{fig:port_weight}}
		\caption{Learning the expected return of a Portfolio optimization problem with $ N = 10000 $ noisy portfolios and $J = 41$ weight samples. (a) The red line indicates the real efficient frontier. The blue dots indicates the estimated efficient frontier using the estimated expected return. (b) Each bar represents the number of the noisy portfolios that have the corresponding weights for $f_{1}(\mfx)$.}
		\label{fig:port}
	\end{figure}

	We use the Portfolio data \textit{BlueChipStockMoments} derived from real data in the Matlab Financial Toolbox. The true expected returns and true return covariances matrix for the first $ 8 $ securities are given in Appendix. W.L.O.G, we suppose that the expected returns for the last three securities are known. The data is generated as follows. We set the upper bounds for the proportion of the $ 8 $ securities to $ b_{i} = 1.0 , \forall i \in [8] $. We first generate optimal portfolios on the efficient frontier in Figure \ref{fig:effcientfrontierport} by solving \ref{weighting problem} with weight samples $ \{w_{i}\}_{i \in [N]} $ chosen from $ \mathscr{W}_{2} $. The first element of $ w_{i} $, ranging from $ 0 $ to $ 1 $, follows a truncated normal distribution derived from a normal distribution with mean $ 0.5 $ and standard deviation $ 0.1 $. In what follows, we will not distinguish truncated normal distribution from normal distribution because their difference is negligible. Subsequently, each component of these portfolios is rounded to the nearest thousandth, which can be seen as measurement error.
	
	Algorithm \ref{alg:clustering-imop} is applied in this experiment. For a reason similar to the previous experiment, we use the ADMM approach (Algorithm \ref{alg:admm-imop}) to solve \ref{update-imop-kmeans}. The stopping criterion for Algorithm \ref{alg:clustering-imop} is that the maximum iteration number reaches five. In the \textbf{Initialization step}, we run K-means++ algorithm  $ 50 $ times to find the best clustering result. When solving \ref{update-imop-kmeans} using ADMM, we partition the observations in such a way that each group has only one observation. We pick the penalty parameter $ \rho = 1 $ as the best out of a few trials. We initialize $\mfr^{0} = \mfv^{t,0} = \zero_{8} $ for the iterations. The tolerances of the primal and dual residuals are set to be $ \epsilon^{pri} = \epsilon^{dual} = 10^{-4} $. The termination criterion is that either the norms of the primal and dual residuals are smaller than $ 10^{-4} $ or the iteration number $ k $ reaches $ 10 $.
	
	In Table \ref{table:estimation_portfolio}, we list the estimation error averaged over $ 10 $ repetitions of the experiments for each $ N $ and $ K $ using Algorithm \ref{alg:admm-imop}. The estimation error has the trend to becomes smaller when $ N $ and $ K $ increase, indicating the estimation consistency and thus risk consistency of the method we propose. We also plot our estimation on the distribution of the weight of $f_{1}(\mfx)$ among the noisy decisions. As shown in Figure \ref{fig:port_weight}, the number of noisy decisions assigned to each weight follows a normal distribution with mean $ 0.5012 $ and standard deviation $ 0.1013 $. The $ 0.95 $ confidence intervals for the mean and standard deviation are $ [0.4992, 0.5032] $ and $ [0.0999, 0.1027] $, respectively. It is reasonable as we generate the portfolios by solving \ref{weighting problem} with normally sampled weights and the feasible set of $\mfx$ is of a much weaker boundary effect, comparing to that in Section \ref{section:qp_c}.

	\begin{table}[ht]
		\centering
		\caption{Estimation Error $ \norm[2]{\hat{\mfr} - \mfr_{true}} $ for Different $ N $ and $ K $}
		\label{table:estimation_portfolio}
		\begin{tabular}{@{}ccccccc@{}}
			\toprule
			& $N=100$ & $N=1000$ & $N=2500$ & $N=5000$ & $N=7500$ & $N=10000$ \\ \midrule
			$K=11$ & 0.0337  & 0.0513   & 0.0406   & 0.0264   & 0.0227   & 0.0194    \\
			$K=21$ & 0.0164  & 0.0154   & 0.0077   & 0.0055   & 0.0042   & 0.0043    \\
			$K=41$ & 0.0220  & 0.0054   & 0.0030   & 0.0022   & 0.0018   & 0.0016    \\
			$K=81$ & 0.0215  & 0.0028   & 0.0017   & 0.0008   & 0.0008   & 0.0008    \\ \bottomrule
		\end{tabular}
	\end{table}
	
	\subsection{Learning the O-D Matrix}
	Let $ G = (N,A) $ be a directed transportation network defined by a set $ N $ of nodes and a set $ A $ of directed links. Each link $ a \in A $  has an associated flow-dependent travel time $ t_{a}(v_{a}) $ that denotes the average travel time on each link. The travel time function $ t_{a}(v_{a}) $ is assumed to be differentiable, convex, and monotonically increasing with the amount of flow $ v_{a} $. Each link $ a \in A $  also has an associated flow-dependent traffic emissions $ e_{a}(v_{a}) $ that denotes the average traffic emissions on each link. Let $ W $ denote the set of O-D pairs, $ R_{w} $ denote the set of all routes between the O-D pair $ w \in W $, $ d_{w} $ represents the travel demand of O-D pair $ w $, and $ f^{w}_{r} $ denote the traffic flow on the route $ r $ connecting the O-D pair $ w $. $ \delta^{w}_{ar} = 1 $ if route $ r \in R^{w} $ uses link $ a $, and $ 0 $ otherwise.
	
	We consider the following Bi-criteria traffic network system optimization problem of minimizing congestion and traffic emissions simultaneously \citep{yin2006internalizing}:
	\begin{align*}
	\begin{array}{llll}
	\min & \left( \begin{matrix} \sum\limits_{a \in A} t_{a}(v_{a})v_{a} \vspace{2mm}\\ \sum\limits_{a \in A}e_{a}(v_{a})v_{a}\end{matrix} \right) \vspace{2mm} \\
	\;s.t.  & d_{w} = \sum\limits_{r \in R_{w}} f^{w}_{r}, & \forall w \in W, \vspace{1mm} \\
	& v_{a} = \sum\limits_{w \in W}\sum\limits_{r \in R_{w}} f^{w}_{r}\delta^{w}_{ar}, & \forall a \in A, \vspace{1mm} \\
	& v_{a}, f^{w}_{r} \geq 0,  & \forall r \in R_{w}, w \in W.
	\end{array}
	\end{align*}
	Note that the problem becomes a minimization of a weighted combination of congestion and traffic emissions if the external costs of congestion and emissions can be obtained. These costs change from time to time, which will lead to different link flows. We seek to learn the O-D matrix given the link flows under different values of time and monetary valuation of traffic emissions. In addition, the presence of measurement errors in the observed link flows are explicitly considered.
	
\begin{figure}[ht]
		\centering
		\includegraphics[width=0.33\linewidth]{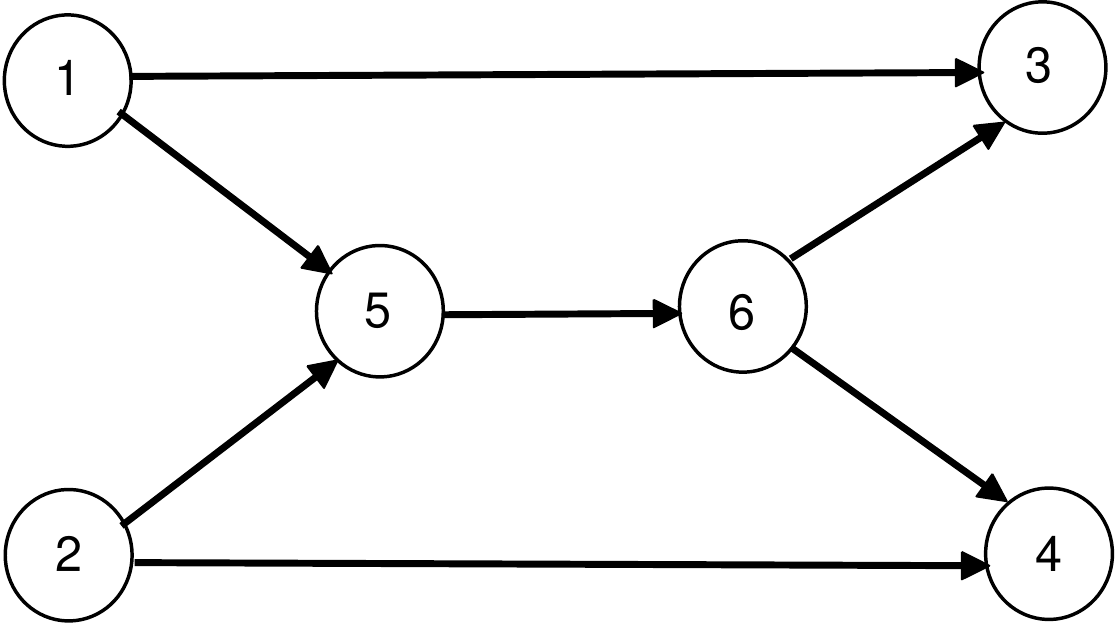}
		\caption{A Six-node Network}
		\label{fig:sixnodenetwork}
	\end{figure}

	Fig \ref{fig:sixnodenetwork} shows a road network with six nodes and seven links used in \citet{yan1996optimal,yin2006internalizing}. The network has two O-D pairs $ (1,3) $ and $ (2,4) $, where $ (1,3) $ has the demand of $ 2500 $ vehicles per hour and $ (2,4) $ has the demand of $ 3500 $ vehicles per hour. We use the US Bureau of Public Road link travel time function to determine the travel time on each link. The function is of the form
	$ t_{a}(v_{a}) = t^{0}_{a}(1 + 0.15\cdot (v_{a}/C_{a})^{4}) $,
	where $ t^{0}_{a} $ and $ C_{a} $ are parameters representing the free-flow travel time (in minutes) and capacity (vehicles per hour) of link $ a \in A $.
	
	We follow the work \citep{nagurney2000congested} and assume the total emissions generated by the vehicles on link $ a $ is
	$ e_{a}(v_{a}) = h_a v_{a} $,
	where $ h_{a} $ denotes the emission factor associated with link $ a $. The key part in the estimation of vehicle emissions is that the volume of emissions equals to the product of emission factors times the link flow. The values of the parameters are listed in Table \ref{table:six_node_network_para}.
	
	\begin{table}[ht]
		\centering
		\caption{Data for the Six-node Network}
		\label{table:six_node_network_para}
		\begin{tabular}{@{}cllllccccccccccccccccccccccccccccccc@{}}
			\toprule
			Link $ a $       &  &  &  &  & (1,3) &  &  &  &  & (2,4) &  &  &  &  & (1,5) &  &  &  &  & (5,6) &  &  &  &  & (2,5) &  &  &  &  & (6,3) &  &  &  &  & (6,4) \\ \midrule
			$t^{0}_{a}$ &  &  &  &  & 8.0   &  &  &  &  & 9.0   &  &  &  &  & 2.0   &  &  &  &  & 6.0   &  &  &  &  & 3.0   &  &  &  &  & 3.0   &  &  &  &  & 4.0   \\
			$C_{a}$     &  &  &  &  & 2000  &  &  &  &  & 2000  &  &  &  &  & 2000  &  &  &  &  & 4000  &  &  &  &  & 2000  &  &  &  &  & 2500  &  &  &  &  & 2500  \\
			$h_{a}$ &  &  &  &  & 8.0   &  &  &  &  & 9.0   &  &  &  &  & 2.0   &  &  &  &  & 6.0   &  &  &  &  & 3.0   &  &  &  &  & 3.0   &  &  &  &  & 4.0   \\ \bottomrule
		\end{tabular}
	\end{table}
	
	We generate the data as follows. We start by computing the efficient solutions $ \{\mfy_{i}\}_{i \in [N]} $ using the weighted sum approach. The weights $ \{w_{i}\}_{i \in [N]} $ are uniformly sampled such that $ w_{i} \in \mathscr{W}_{2} $ for each $ i \in [N] $, where $ N = 10 $.  Since we do not want to over emphasize either the congestion or the traffic emission in the bi-criteria traffic network system, we concentrate the weights and set $ w_{i} \in [0.3,0.7]^{2} $ for each $ i \in [N] $. Subsequently, each component of the efficient solutions is rounded to the nearest ten, which can be treated as measurement error. We assume the demand of O-D pairs $ (1,3) $ and $ (2,4) $ are bigger than $ 1000 $ and smaller than $ 10000 $ vehicles per hour. Then, we evenly sample the weights $ \{w_{k}\}_{k \in [K]} $ such that $ w_{k} \in \mathscr{W}_{2} $ for each $ k \in [K] $.
	
	We implement the SRe approach using the solver FilMINT. The solutions returned by FilMINT are not guaranteed to be optimal since the inference of the O-D matrix requires solving a mixed integer nonconvex program. FilMINT can handle instances with $ K \leq 100 $ quite efficiently. In Table \ref{table:estimated_error_od} we summarize the computational results for different $ K $. The table lists for each $ K $ the estimations for the demands of O-D pairs $ (1,3) $ and $ (2,4) $, and also the estimation error, which is given by $ \norm{\text{estimation - true O-D}}/\norm{\text{true O-D} } $. The table shows that the estimation error becomes smaller and smaller when $ K $ increases, which indicates that our method still works in the general convex MOP.
	
	\begin{table}[ht]
		\centering
		\caption{Estimation Results for Different $ K $}
		\label{table:estimated_error_od}
		\begin{tabular}{@{}cccccc@{}}
			\toprule
			$ K $          & 6       & 11      & 21      & 41      & 81      \\ \midrule
			O-D $(1,3)$      & 2056.79 & 2218.64 & 2218.64 & 2218.64 & 2288.95 \\
			O-D $(2,4)$      & 2185.46 & 3259.60  & 3259.60  & 3259.60  & 3576.67 \\
			Estimation error & 0.3225  & 0.0860  & 0.0860  & 0.0860  & 0.0522  \\ \bottomrule
		\end{tabular}
	\end{table}

	\section{Conclusions}
	
	We study in this paper the problem of learning the objective functions and constraints of a multiobjective decision making problem, based on observations of efficient solutions which might carry noise. Specifically, we formulate such a learning task as an inverse multiobjective optimization problem, and provide a deep analysis to establish the statistical significance of the inference results from the presented model. Moreover, we discuss the strong correlation between the identifiability of the decision making problem and the performance of our inverse optimization model. We then develop two numerical algorithms to handle the computational challenge from the large number of observations. We confirm by extensive numerical experiments that the proposed algorithms can learn the parameters with great accuracy while drastically improve the computational efficacy.
	
	%
	%
	%
	
	

	
	
	
	
	\begin{APPENDICES}{}
		\section{Omitted Proofs}
		
		\subsection{Proof of Lemma \ref{feasible-set-continuous}}
		\proof{Proof. }
		Since $ \mathbf{g}(\mfx,\theta) $ is continuous and thus l.s.c. on $ \bR^{n} \times \Theta $ by ASSUMPTION \ref{set-continuous-assumption}, $ X(\theta) $ is u.s.c. for each $ \theta \in \Theta $ by Theorem 10 in \citet{hogan1973point}. From ASSUMPTION \ref{convex_setting}, we know that $ \mathbf{g}(\mfx,\theta) $ is convex in $ \mfx $ for each $ \theta \in \Theta $. From ASSUMPTION \ref{set-continuous-assumption}, $ X(\theta) $ has a nonempty relatively interior. Namely, there exists a $ \bar{\mfx} \in \bR^{n} $ such that $ \mathbf{g}(\bar{\mfx},\theta) < \zero $. Then, $ X(\theta) $ is l.s.c. for each $ \theta \in \Theta $ by Theorem 12 in \citet{hogan1973point}. Hence, $ X(\theta) $ is continuous on $ \Theta $.
		\Halmos\endproof
		
		\subsection{Proof of Lemma \ref{lemma:efficient-set-continuous}}
		\proof{Proof. }
		First, we will show that $ X_{E}(\theta) $ is u.s.c. on $ \Theta $. Since $ \mathbf{f}(\mfx,\theta) $ is strictly convex in $ \mfx $ for each $ \theta \in \Theta $, the efficient set $ X_{E}(\theta) $ coincides with the weakly efficient set $ X_{wE}(\theta) $. In addition, we know that $ X(\theta) $ is continuous on $ \Theta $ by Lemma \ref{feasible-set-continuous}. Also, note the pointed convex cone we use throughout this paper has the same meaning as the domination structure $ D $ in \citet{tanino1980stability}, and we set $ D = \bR^{p}_{+} $. To this end, we can readily verify that the sufficient conditions for upper semicontinuity in Theorem 7.1 of \citet{tanino1980stability} are satisfied. Thus, $ X_{E}(\theta) $ is u.s.c..
		
		Next, we will show that $ X_{E}(\theta) $ is l.s.c. on $ \Theta $. Theorem 7.2 of \citet{tanino1980stability} provides the sufficient conditions for the lower semicontinuity of $ X_{E}(\theta) $. All of these conditions are naturally satisfied under Assumptions \ref{convex_setting} - \ref{set-continuous-assumption} except the one that requires $ \mathbf{f}(\mfx,\theta) $ to be one-to-one, i.e., injective in $ \mfx $. Next, we will show that the one-to-one condition can be safely replaced by the strict quasi-convexity of $ \mathbf{f}(\mfx,\theta) $ in $ \mfx $.
		
		Theorem 7.2 of \citet{tanino1980stability} is a direct result of part (ii) in Lemma 7.2 of \citet{tanino1980stability}. To complete our proof, we only need to sightly modify the last part of the proof in Lemma 7.2. In what follows we will use notations in that paper.
		
		Since strict convexity implies strict quasi-convexity, $ f $ is strictly quasi-convex. Suppose that $ f(\bar{x},\hat{u}) = f(\hat{x},\hat{u}) $ does not imply $ \bar{x} = \hat{x} $. Let $ z = \frac{\bar{x} + \hat{x}}{2} $. By the strict quasi-convexity of $ f $, we have
		\begin{align*}
		f(z,\hat{u}) = f(\frac{\bar{x} + \hat{x}}{2},\hat{u}) < \max\{f(\bar{x},\hat{u}),f(\hat{x},\hat{u})\} = f(\hat{x},\hat{u}).
		\end{align*}
		
		This contradicts the fact that $ \hat{x} \in M(\hat{u}) $, where $ M(\hat{u})  $ is the efficient set given $ \hat{u} $. Hence, $ \bar{x} $ must be equal to $ \hat{x} $. The remain part of the proof is the same as that of Lemma 7.2.
		\Halmos\endproof

		\subsection{Proof of Lemma \ref{lemma:M_J monotone}}
		\proof{Proof. }
		\textbf{(a)} Let $ K_{2} \geq K_{1} $. Under our setting,  $ K_{2} \geq K_{1} $ implies $ \{w_{k}\}_{k\in[K_{1}]} \subseteq \{w_{k}\}_{k\in[K_{2}]} $. By the definition of $ l_{K}(\mfy,\theta) $, we have $ l_{K_{1}}(\mfy,\theta) \geq l_{K_{2}}(\mfy,\theta) $ for all $ \mfy \in \mathcal{Y} $, and thus $ M_{K_{1}}(\theta) \geq M_{K_{2}}(\theta) $ for all $ \theta \in \Theta $. Therefore, $ \{M_{K}(\theta)\} $ is monotone decreasing in $ K $.
		
		Recall the definition of $ \hat{\theta}_{K} $ in Table \ref{table:four models}, we know $ \hat{\theta}_{K_{2}} $ minimizes $ M_{K_{2}}(\theta) $. Therefore, $ M_{K_{2}}(\hat{\theta}_{K_{1}}) \geq M_{K_{2}}(\hat{\theta}_{K_{2}}) $. In addition, $ M_{K_{1}}(\hat{\theta}_{K_{1}}) \geq M_{K_{2}}(\hat{\theta}_{K_{1}}) $ by the first part of \textbf{(a)}. Consequently,
		\[ M_{K_{1}}(\hat{\theta}_{K_{1}}) \geq M_{K_{2}}(\hat{\theta}_{K_{1}}) \geq M_{K_{2}}(\hat{\theta}_{K_{2}}). \]
		
		Therefore, $ M_{K_{1}}(\hat{\theta}_{K_{1}}) \geq M_{K_{2}}(\hat{\theta}_{K_{2}}) $ for $ K_{2} \geq K_{1} $.
		
		Similarly, we can readily show that $ M_{K}(\hat{\theta}_{K}) \geq M(\theta^{*}) $ by noting that
		\[ M_{K}(\hat{\theta}_{K}) \geq M(\hat{\theta}_{K}) \geq M(\theta^{*}). \]
		The first inequality is a direct result of the first part of \textbf{(a)}; the second inequality follows from the fact that $ \theta^{*} $ minimizes $ M(\theta) $ by definition.
		
		\textbf{(b)} Let $ K_{2} \geq K_{1} $. By the definition of $ l_{K}(\mfy,\theta) $, we have $ l_{K_{1}}(\mfy_{i},\theta) \geq l_{K_{2}}(\mfy_{i},\theta) $ for all $ i \in [N] $, and thus $ M_{K_{1}}^{N}(\theta) \geq M_{K_{2}}^{N}(\theta) $ for all $ \theta \in \Theta $. Therefore, $ \{M_{K}^{N}(\theta)\} $ is monotone decreasing in $ K $.
		
		Recall the definition of $ \hat{\theta}_{K}^{N} $ in Table \ref{table:four models}, we know $ \hat{\theta}_{K_{2}}^{N} $ minimizes $ M_{K_{2}}^{N}(\theta) $. Therefore, $ M_{K_{2}}^{N}(\hat{\theta}_{K_{1}}^{N}) \geq M_{K_{2}}^{N}(\hat{\theta}_{K_{2}}^{N}) $. In addition, $ M_{K_{1}}^{N}(\hat{\theta}_{K_{1}}^{N}) \geq M_{K_{2}}^{N}(\hat{\theta}_{K_{1}}^{N}) $ by the first part of \textbf{(b)}. Consequently,
		\begin{align*}
		M_{K_{1}}^{N}(\hat{\theta}^{N}_{K_{1}}) \geq M_{K_{2}}^{N}(\hat{\theta}^{N}_{K_{1}}) \geq M_{K_{2}}^{N}(\hat{\theta}^{N}_{K_{2}}).
		\end{align*}
		
		Hence, $  M_{K_{1}}^{N}(\hat{\theta}^{N}_{K_{1}}) \geq M_{K_{2}}^{N}(\hat{\theta}^{N}_{K_{2}}) $ for $ K_{2} \geq K_{1} $.
		
		Finally, we can show $ M_{K}^{N}(\hat{\theta}^{N}_{K}) \geq M^{N}(\hat{\theta}^{N}) $ by noting that $ M_{K}^{N}(\hat{\theta}^{N}_{K}) \geq M^{N}(\hat{\theta}^{N}_{K}) \geq M^{N}(\hat{\theta}^{N}) $.
		\Halmos\endproof
		
		\subsection{Proof of Lemma \ref{lemma:lpshitz of S(w,theta)}}
		\proof{Proof. }
		$ \forall w \in \mathscr{W}_{p} $, one can readily check that $ w^{T}\mff(\cdot,\theta) $ is strongly convex for each $ \theta $ and thus
		\begin{align*}
		w^{T}\mff(\mfy,\theta) \geq w^{T}\mff(\mfx,\theta) + \nabla w^{T}\mff(\mfx,\theta)^{T}(\mfy - \mfx) + \frac{\lambda}{2}\norm{\mfy - \mfx}^{2}.
		\end{align*}
		
		Thus, the second-order growth condition holds for $ w^{T}\mff(\cdot,\theta) $ for all $ \theta \in \Theta $. That is,
		\begin{align}
		w^{T}\mff(\mfx,\theta) \geq w^{T}\mff(S(w,\theta),\theta) + \frac{\lambda}{2}\norm{(S(w,\theta) - \mfx}^{2}.
		\end{align}
		
		In addition, $ \forall w, w_{0} \in \mathscr{W}_{p} $, we have
		\begin{align}
		\begin{array}{llll}
		|w^{T}\mff(\mfx,\theta) - w_{0}^{T}\mff(\mfx,\theta)| & = |(w^{T} - w_{0}^{T})^{T}\mff(\mfx,\theta)| \vspace{1mm} \\
		& \leq \norm{w^{T} - w_{0}^{T}} \norm{\mff(\mfx,\theta)} && \text{(Cauchy-Schwarz inequality)} \vspace{1mm} \\
		& \leq L\norm{w^{T} - w_{0}^{T}}.
		\end{array}
		\end{align}
		
		Besides, note that the feasible set $ X(\theta) $ is irrelevant to $ w $. Then, applying Proposition 6.1 \citep{bonnans1998optimization} yields $ \forall \theta \in \Theta $,
		\begin{align*}
		\lVert S(w,\theta) - S(w_{0},\theta) \rVert_{2} \leq \frac{2L}{\lambda}\norm{w - w_{0}}.
		\end{align*}
		\Halmos\endproof

		\subsection{Proof of Lemma \ref{guarantee: risk-bounds}}
		\proof{Proof. }
		Let $ \mathcal{G} $ be a class of functions $ g $ mapping from $ Z $ to $ \bR $, where
		\begin{align}\label{transform f}
		g(Z) = \frac{f(Z) - a}{b - a}.
		\end{align}
		Note that $ g(Z) \in [0,1]  $. By Theorem 3.1 in \citet{mohri2012foundations}, we have
		\begin{align}\label{empirical risk bound in g}
		\bE[g(Z)] \leq \frac{1}{N}\sum_{i \in [N]}g(Z_{i}) + 2Rad_{N}(\mathcal{G} ) + \sqrt{\frac{log(1/\delta)}{2N}}.
		\end{align}
		Using part 3 in Theorem 12 of \citet{bartlett2002rademacher}, and the translation invariant property, i.e., $ Rad_{N}(\mathcal{F} - a ) = Rad_{N}(\mathcal{F} ) $, we have
		\begin{align}\label{rademacher relationship}
		Rad_{N}(\mathcal{G}) = Rad_{N}\left(\frac{\mathcal{F} - a}{b - a}\right) = \frac{Rad_{N}(\mathcal{F})}{b - a}.
		\end{align}
		Plugging \eqref{transform f} and \eqref{rademacher relationship} in \eqref{empirical risk bound in g} yields the main result.
		\Halmos\endproof
		
		\subsection{Proof of Lemma \ref{guarantee: rademacher-complexity}}
		\proof{Proof. }
		By the definition of Rademacher complexity, we have
		\begin{align*}
		\begin{array}{llll}
		Rad_{N}\big(\mathcal{F}\big) & = \frac{1}{N}\bE\bigg[\sup\limits_{f \in \mathcal{F}}\sum\limits_{i \in [N]}\sigma_{i}f(\mfy_{i},\theta)\bigg] \vspace{1mm} \\
		& = \frac{1}{N}\bE\bigg[\sup\limits_{\theta \in \Theta}\sum\limits_{i \in [N]}\sigma_{i}\min\limits_{k \in [K]}\lVert \mfy_{i} - \mfx_{k}\rVert_{2}^{2}\bigg] \vspace{1mm} \\
		& = \frac{1}{N}\bE\bigg[\sup\limits_{\theta \in \Theta}\sum\limits_{i \in [N]}\sigma_{i}\min\limits_{k \in [K]}\big(\norm{\mfy_{i}}^{2} - 2\langle \mfy_{i}, \mfx_{k}\rangle + \norm{\mfx_{k}}^{2}\big)\bigg] \vspace{1mm} \\
		& = \frac{1}{N}\bE\bigg[\sup\limits_{\theta \in \Theta}\sum\limits_{i \in [N]}\sigma_{i}\min\limits_{k \in [K]}\big( - 2\langle \mfy_{i}, \mfx_{k}\rangle + \norm{\mfx_{k}}^{2}\big)\bigg].\vspace{1mm} \\
		\end{array}
		\end{align*}
		
		Note the fact $ \bP(\norm{\mfx} \leq B) = 1 $ by Assumption \ref{set-continuous-assumption}. Through a similar argument in statement (ii) of Lemma 4.3 in \citet{biau2008performance}, we get
		\begin{align}\label{rademacher-proof0}
		\frac{1}{N}\bE\bigg[\sup\limits_{\theta \in \Theta}\sum\limits_{i \in [N]}\sigma_{i}\min\limits_{k \in [K]}\big( - 2\langle \mfy_{i}, \mfx_{k}\rangle + \norm{\mfx_{k}}^{2}\big)\bigg] \leq 2K\bigg(\frac{1}{N}\bE\bigg[\sup\limits_{\norm{\mfx} \leq B}\sum\limits_{i \in [N]}\sigma_{i}\langle \mfy_{i}, \mfx\rangle\bigg] + \frac{B^{2}}{2\sqrt{N}}\bigg).
		\end{align}
		
		The first term on the right-hand side of \eqref{rademacher-proof0} can be upper bounded in the following way:
		\begin{align}\label{rademacher-proof}
		\begin{array}{llll}
		\frac{1}{N}\bE\bigg[\sup\limits_{\norm{\mfx} \leq B}\sum\limits_{i \in [N]}\sigma_{i}\langle \mfy_{i}, \mfx\rangle\bigg] & = \frac{1}{N}\bE\bigg[\sup\limits_{\norm{\mfx} \leq B}\langle \sum\limits_{i \in [N]}\sigma_{i}\mfy_{i}, \mfx\rangle\bigg] \vspace{1mm} \\
		& \leq \frac{1}{N}\bE\sup\limits_{\norm{\mfx} \leq B}\norm{\mfx}\norm{\sum\limits_{i \in [N]}\sigma_{i}\mfy_{i}} & & \text{(Cauchy-Schwarz inequality)} \vspace{1mm} \\
		& \leq \frac{B}{N}\bE\norm{\sum\limits_{i \in [N]}\sigma_{i}\mfy_{i}} \vspace{1mm} \\
		& \leq \frac{B}{N}\sqrt{\bE\norm{\sum\limits_{i \in [N]}\sigma_{i}\mfy_{i}}^{2}} & & \text{(Jensen's inequality)}\vspace{1mm} \\
		& = \frac{B}{N}\sqrt{N\bE\norm{\mfy}^{2}} \vspace{1mm} \\
		& \leq \frac{BR}{\sqrt{N}} & & \text{($ \bP(\norm{\mfy} \leq R) = 1 $)}.
		\end{array}
		\end{align}
		Plugging the result of \eqref{rademacher-proof} in \eqref{rademacher-proof0}, we get the bound for the Rademacher complexity of $ \mathcal{F}  $.
		\Halmos\endproof

		\section{Omitted Mathematical Formulations}
		
		\subsection{Reformulation of \ref{test-problem} Using KKT Conditions}\label{app:test-problem}
		\begin{align*}
		\begin{array}{llll}
		\max\limits_{\theta \in \Theta} & \norm[1]{\theta - \hat{\theta}^{N}_{K}} \vspace{1mm}\\
		\;s.t. & \mfu_{i} \geq \zero, & \forall i \in [N'],  \vspace{1mm}\\
		& \mfu_{i}^{T}\mathbf{g}(\mfx_{i},\theta) = 0, & \forall i \in [N'],  \vspace{1mm} \\
		& \norm[2]{\nabla_{\mfx_{i}} w_{k}^{T}\mathbf{f}(\mfx_{i},\theta) + \mfu_{i}^{T}\nabla_{\mfx_{i}}\mathbf{g}(\mfx_{i},\theta)} \leq M(1-z_{ik}), & \forall i \in [N'], k \in [K'], \vspace{1mm} \\
		& \sum\limits_{k \in [K']}z_{ik} = 1, & \forall i \in [N'], \\
		& z_{ik} \in \{0,1\},\;\; \mfu_{i} \in \bR^{q}_{+}, & \forall i \in [N'], k \in [K'].
		\end{array}
		\end{align*}

		\subsection{Single Level Reformulation for Inferring Objective Functions of MLP} \label{imlp}
		\begin{align*}
		\label{imop:linear_single_level}
		\begin{array}{llll}
		\min\limits_{\mfc_{1},\cdots,\mfc_{p}} & \sum\limits_{i \in [N]}\lVert \mfy_{i} - \sum\limits_{k \in [K]}\eta_{ik}\rVert_{2} \vspace{1mm} \\
		\;s.t. & \mfc_{l} \in C_{l}, & \forall l \in [p],\vspace{1mm}\\
		&\left[\begin{array}{llll}
		& \mfA\mfx_{k} \geq \mfb,\; \mfx_{k} \geq \zero, \vspace{1mm}\\
		& \mfA^{T}\mfu_{k} \leq w_{k}^{1}\mfc_{1} + \cdots + w_{k}^{p}\mfc_{p},\; \mfu_{k} \geq 0, \vspace{1mm} \\
		& \mfx_{k} \leq M_{1}\mft_{1k}, \vspace{1mm} \\
		& w_{k}^{1}\mfc_{1} + \cdots + w_{k}^{p}\mfc_{p} - \mfA^{T}\mfu_{k}  \leq M_{1}(1 - \mft_{1k}), \vspace{1mm}\\
		& \mfu_{k} \leq M_{2}\mft_{2k}, \vspace{1mm} \\
		& \mfA\mfx_{k} - \mfb \leq M_{2}(1 - \mft_{2k}) \end{array} \right], & \forall k \in [K], \vspace{1mm} \\
		& 0 \leq \eta_{ik} \leq M_{ik}z_{ik}, & \forall i \in [N], k \in [K], \vspace{1mm} \\
		& \mfx_{k} - M_{ik}(1 - z_{ik}) \leq \eta_{ik} \leq \mfx_{k}, & \forall i \in [N], k \in [K], \vspace{1mm} \\
		& \sum\limits_{k \in [K]}z_{ik} = 1, & \forall i \in [N], \vspace{1mm} \\
		& \mfx_{k} \in \bR^{n}_{+},\;\; \mfu_{k} \in \bR^{m}_{+}, \;\; \mft_{1k} \in \{0,1\}^{n},\;\; \mft_{2k} \in \{0,1\}^{m},\;\; z_{ik} \in \{0,1\}, & \forall i \in [N], k \in [K],
		\end{array}
		\end{align*}
		where $ C_{l} $ is a convex compact set for each $ l \in [p] $. $ M_{1} $, $ M_{2} $ and $ M_{ik} $ are Big-Ms used to linearize the program. One can establish similar reformulations for inferring RHS of MLP.
		
		\subsection{Single Level Reformulation for Inferring RHS of MQP}\label{imop:qp-b}
		\begin{align*}
		\begin{array}{llll}
		\min\limits_{\mfb} & \sum\limits_{i \in [N]}\lVert \mfy_{i} - \sum\limits_{k \in [K]}\eta_{ik}\rVert_{2} \vspace{1mm} \\
		\text{s.t.} & \mfb \in B, \vspace{1mm} \\
		& \left[\begin{array}{llll}
		& \mfA\mfx_{k} \geq \mfb,\; \mfu_{k} \geq \zero, \vspace{1mm}\\
		& \mfu_{k} \leq M_{1}\mft_{k}, \vspace{1mm} \\
		& \mfA\mfx_{k} - \mfb \leq M_{1}(1 - \mft_{k}),  \vspace{1mm} \\
		& (w_{k}^{1}Q_{1} + \cdots + w_{k}^{p}Q_{p})\mfx_{i} + w_{k}^{1}\mfc_{1} + \cdots + w_{k}^{p}\mfc_{p} - \mfA^{T}\mfu_{k} = 0, \vspace{1mm}
		\end{array} \right], & \forall k \in [K], \vspace{1mm} \\
		& 0 \leq \eta_{ik} \leq M_{ik}z_{ik}, & \forall i \in [N], k \in [K], \vspace{1mm} \\
		& \mfx_{k} - M_{ik}(1 - z_{ik}) \leq \eta_{ik} \leq \mfx_{k} + M_{ik}(1 - z_{ik}), & \forall i \in [N], k \in [K], \vspace{1mm} \\
		& \sum\limits_{k \in [K]}z_{ik} = 1, & \forall i \in [N], \vspace{1mm} \\
		& \mfb \in \bR^{m},\;\; \mfx_{k} \in \bR^{n},\;\; \mfu_{k} \in \bR^{m}_{+}, \;\; \mft_{k} \in \{0,1\}^{m},\;\; z_{ik} \in \{0,1\}, & \forall i \in [N], k \in [K],
		\end{array}
		\end{align*}
		where $ B $ is a convex compact set. $ M_{1} $ and $ M_{ik} $ are Big-Ms used to linearize the program. One can establish similar reformulations for inferring objectives of MQP.

		\section{Data for the Portfolio Optimization Problem}
		\renewcommand{\arraystretch}{.6}	
		\begin{table}[ht]
			\centering
			\caption{True Expected Return}
			\label{table:true_return}
			\begin{tabular}{@{}ccccccccc@{}}
				\toprule
				Security        & 1      & 2      & 3      & 4      & 5      & 6      & 7      & 8      \\ \midrule
				Expected Return & 0.1791 & 0.1143 & 0.1357 & 0.0837 & 0.1653 & 0.1808 & 0.0352 & 0.0368 \\ \bottomrule
			\end{tabular}
		\end{table}
		\begin{table}[ht]
			\centering
			\caption{True Return Covariances Matrix}
			\label{table:true_covariance}
			\begin{tabular}{@{}ccccccccccc@{}}
				\toprule
				Security & 1      & 2      & 3      & 4      & 5      & 6      & 7      & 8      \\ \midrule
				1        & 0.1641 & 0.0299 & 0.0478 & 0.0491 & 0.058  & 0.0871 & 0.0603 & 0.0492 \\
				2        & 0.0299 & 0.0720  & 0.0511 & 0.0287 & 0.0527 & 0.0297 & 0.0291 & 0.0326 \\
				3        & 0.0478 & 0.0511 & 0.0794 & 0.0498 & 0.0664 & 0.0479 & 0.0395 & 0.0523 \\
				4        & 0.0491 & 0.0287 & 0.0498 & 0.1148 & 0.0336 & 0.0503 & 0.0326 & 0.0447 \\
				5        & 0.0580  & 0.0527 & 0.0664 & 0.0336 & 0.1073 & 0.0483 & 0.0402 & 0.0533 \\
				6        & 0.0871 & 0.0297 & 0.0479 & 0.0503 & 0.0483 & 0.1134 & 0.0591 & 0.0387 \\
				7        & 0.0603 & 0.0291 & 0.0395 & 0.0326 & 0.0402 & 0.0591 & 0.0704 & 0.0244 \\
				8        & 0.0492 & 0.0326 & 0.0523 & 0.0447 & 0.0533 & 0.0387 & 0.0244 & 0.1028 \\ \bottomrule
			\end{tabular}
		\end{table}
		
	\end{APPENDICES}
	
	\bibliographystyle{ormsv080} 
	\bibliography{reference} 
	
\end{document}